\documentclass{article}
\usepackage[utf8]{inputenc}
\usepackage[arabic,english]{babel}
\DeclareMathSymbol{\mtimes}{\mathbin}{symbols}{"02}
\usepackage{amsmath,amssymb,scalefnt}
\usepackage{makecell}
\usepackage{tikz}
\usepackage{color,soul}
\usepackage{hyperref}
\usepackage{enumerate}
\usepackage{apacite}
\usepackage{pgfplots}
\usepackage{cancel}
\usepackage{apacite}
\usepackage[most]{tcolorbox}
\usepackage{caption}
\usepackage{float}
\usepackage{subcaption}
\usepackage[left,modulo]{lineno}
\usetikzlibrary{matrix,calc}
\usetikzlibrary{arrows,positioning,fit,shapes,calc,shadows,shapes.misc}
\pgfplotsset{width=10cm,compat=1.11}
\title{External knowledge transfer deployment inside a simple double agent Viterbi algorithm}
\usepackage[symbol]{footmisc}
\def\correspondingauthor{\footnote{Corresponding author: ziedbaklouti@outlook.com }}
\def\university{\footnote{MSc computer science and mathematics , Paris University}}
\author{Zied Baklouti\correspondingauthor{} \university{}}
\date{August 2021}
\tikzset{
every transaction/.style = {fill=white!100},
transaction1/.style = {starburst, draw , aspect=2, align=center, inner sep=1pt},
transaction4/.style = {starburst, draw , aspect=2, align=center, inner sep=1pt},
transaction2/.style = {double copy shadow={shadow xshift=2ex,shadow yshift=2ex},
transaction3/.style = {shape=rounded rectangle, shape example, inner xsep=1.5cm, inner ysep=1cm},
fill=white!20,draw=black,thick},
transaction/.style = {diamond, draw , aspect=2, align=center, inner sep=1pt},
every actor role/.style = {},
actor role/.style = {rectangle, draw=black!80, ultra thick,
    minimum size = 6mm, every actor role},
composite actor role/.style = {fill=white!50, actor role},
composite/.style = {fill=white!100, actor role},
elementary actor role/.style = {fill=white!100, actor role},
initiator/.style = {-},
executor/.style = {<-, >=squarea},
system/.style = {rectangle, fill=white!100, ultra thick, draw=black!80,
            minimum height=60mm, minimum width=4cm,outer sep=0pt}}
\begin{document}
\maketitle
	\begin{abstract}
	We consider in this paper deploying external knowledge transfer inside a simple double agent Viterbi algorithm which is an algorithm firstly introduced by the author in his preprint "Hidden Markov Based Mathematical Model dedicated to Extract Ingredients from Recipe Text"{\color{blue}\cite{baklouti2019hidden}}. The key challenge of this work lies in discovering the reason why our old model does have bad performances when it is confronted with estimating ingredient state for unknown words and see if deploying external knowledge transfer directly on calculating state matrix could be the solution instead of deploying it only on back propagating step.
	\end{abstract}
	
\section{Introduction}

Extracting ingredients from a recipe text is a very common activity especially for data scientists and developers who want to study recipes or want to make statistical representations about nutritive values of cuisine recipes. Ingredients is not the only useful information we want to extract , the quantity used for each ingredient and how they are prepared are also interesting informations that we can extract by the same method presented in this work.  Hidden Markov Models are the first idea that came in my mind because there are previous successful works that used this method for information extraction ({\color{blue}\cite{freitag2000information}},{\color{blue}\cite{freitag1999information}},{\color{blue}\cite{seymore1999learning}},{\color{blue}\cite{bikel1998nymble}},{\color{blue}\cite{leek1997information}}) , and also because modeling sequences of words where we have to estimate the hidden state is typically a hidden Markov procedure.  In this work we are concentrating on the external knowledge part deployed in what we called a simple double agent Viterbi algorithm.

\section{Previous work}

We used Named Entity recognition as a subtask of information extraction to extract ingredients, HMM are widely used to perform named entity recognition  {\color{blue}\cite{morwal2012named}} but in its basic technology , HMM is not able to incorporate external knowledge, this is why we developed on this work an advanced Viterbi algorithm to consider POS tags as an external knowledge playing an active role in calculating the state matrix. The major modification made are explained in {\color{blue}\cite{baklouti2019hidden}} .
Extracting ingredients help developers creating recipe recommendation systems, but not only ingredients are targeted in a text recipe, quantities deployed inside a recipes and cooking techniques are useful information to construct a hybrid recommendation system {\color{blue}\cite{shah2016personalized}}{\color{blue}\cite{ueda2014recipe}} and incorporating POS tags as an external knowledge will help definitely creating more performant hybrid recommendation systems.

Methodologies for transfer learning varies according to the model's task, for example for image classification layers in a neural network may be involved in recognizing specific features and by knowledge transfer we don't have to  re-train those layers in medical images to relearn those specific features {\color{blue}\cite{zhu2011heterogeneous}}. Our bigger part of our methodology is explained in {\color{blue}\cite{baklouti2019hidden}} , in this paper we are going to make some modification in iteration step to take maximum advantage of already estimated knowledge , in our case already estimated knowledge are POS tags.

\section{Methodology}

When observing the equations calculating the elements of the state matrix for a second order mono-agent Viterbi algorithm and compare it with the algorithm used in an ingredient extractor {\color{blue}\cite{baklouti2019hidden}} , we found an interesting property that kept the second order mono-layered Viterbi more precise in the context of unknown words, in other words when the probability $b_{ij}(v_k)=P(V_{r}=v_{k}/\Gamma_{r}=\gamma_{j},T_{r}=\tau_{i})$ is losing its efficiency behind an unknown word , the algorithm used for an ingredient extractor couldn't rely on the other probabilities inside the equations that determines a state matrix. In order to discover the property in second order Viterbi HMM , we created a table of cases constituted with the list of tokens constituting a sentence and we fill it with the parameters inside the lexical probability and the contextual probability for the most probable cases inside a state matrix $\delta$. For example , for the sentence ' \AR{الجيدة} \AR{النوعية} \AR{من} \AR{زيتون} \AR{زيت}' , we want to create its situations table for the token \AR{الجيدة}, we search first for the most probable cases revealed by the higher values of the state matrix $\delta$, the segment of the matrix  $\delta_{6}$ reveals 3 most probables cases : first case the token \AR{الجيدة} is not an ingredient , second case the token \AR{الجيدة} is  an ingredient and it is extracted alone, the third case the token \AR{الجيدة} is a part of an ingredient composed with two words and the word \AR{الجيدة} is the second word of that ingredient. In order to calculate one element of a matrix state $\delta_{i}$ for a mono-layered Viterbi algorithm , the algorithm choose one element of the precedent segment $\delta_{i-1}$ of the state matrix and add it with the corresponding lexical probability and transitional probability. We dressed a 3 situations tables for the most probable situations revealed by the higher elements of the state matrix filled with the parameters of those specific probabilities. After dressing the situations tables , we can distinguish between two kind of probabilities:
\begin{itemize}
\item  \textbf{A sane probability}: showing an existing case, for example the probability $P({\color{blue}\Gamma_{r}=0}/{\color{red}\Gamma_{r-1}=0},{\color{green}\Gamma_{r-2}=0})$ used in {\color{blue}Table \ref{fig:tab1}} calculating $\delta_{6}({\color{blue}1},{\color{red}1})$ in {\color{blue}equation \ref{eq1}} one of the highest elements of {\color{blue}matrix \ref{mat1}} is a sane probability because we are supposing that the token in position $r-1$ is not an ingredient (${\color{red}\Gamma_{r-1}=0}$) and the estimation obtained from the backtracking step for the token in position r is true (${\color{blue}\Gamma_{r}=0}$), and the estimation of the ingredient state for the token in position $r-2$ obtained from the mathematical expression $max_{1 \leq i \leq 4}(\delta_{5}({\color{green}i},{\color{red}2}) + log(P({\color{blue}\Gamma_{r}=0}/{\color{red}\Gamma_{r-1}=1},{\color{green}\Gamma_{r-2}=\gamma_{i}}))$ in {\color{blue}equation \ref{eq1}} is true (${\color{green}\Gamma_{r-2}=0}$).
\item  \textbf{An insane or a parasite probability}: showing a non existing case, for example the probability $P({\color{blue}\Gamma_{r}=0}/{\color{red}\Gamma_{r-1}=2},{\color{green}\Gamma_{r-2}=1})$ in {\color{blue}Table \ref{fig:tab3}} used in calculating $\delta_{6}({\color{blue}1},{\color{red}3})$ in {\color{blue}equation \ref{eq3}} is a parasite probability because we are supposing that the token in position $r-1$ is the second word of an ingredient (${\color{red}\Gamma_{r-1}=2}$) and the estimation obtained from the backtracking step for the token in position r is the real value of the token in position $r$ (${\color{blue}\Gamma_{r}=0}$), and the estimation of the ingredient state for the token in position $r-2$ obtained from the mathematical expression $max_{1 \leq i \leq 4}(\delta_{5}({\color{green}i},{\color{red}3}) + log(P({\color{blue}\Gamma_{r}=0}/{\color{red}\Gamma_{r-1}=2},{\color{green}\Gamma_{r-2}=\gamma_{i}}))$ in {\color{blue}equation \ref{eq3}} is the real value of the token in position $r-2$ (${\color{green}\Gamma_{r-2}=1}$), the first word of an ingredient.
\end{itemize}
We are supposing that a "good" algorithm is an algorithm were there is the least possible parasite probabilities in its most probable state matrix calculations and we are going to try to ameliorate our ingredient extractor in that way and see if this hypothesis is a valid hypothesis by experimentation.

\newpage
{\color{blue}Estimating an ingredient state using a second order HMM model:}

sentence=.,\AR{زيت}(oil),\AR{زيتون}(olive),\AR{من}(of),\AR{النوعية}(quality),\AR{الجيدة}(good),.

ingredient state=[0,1,2,0,0,0,0]

ingredient to be extracted=\AR{زيت زيتون}

\begin{gather} \label{mat1}
	\delta_{6}(i,j)=\begin{bmatrix}
	\color{orange}-41.8209 & -46.5677 & -53.3321 & -54.4383 \\
	\color{orange}-41.0199 & -53.0687 & -47.9396 & -54.175 \\
	\color{orange}-41.6138 & -53.0402 & -53.4034 & -51.8439 \\	
	-55.8817 &-56.4674 & -56.1374 & -55.9642
	\end{bmatrix}
	\end{gather}

\begin{equation} \label{eq1}
\begin{split}
\delta_{6}({\color{red}1},{\color{blue}1}) & = max_{1 \leq i \leq 4}(\delta_{5}({\color{green}i},{\color{red}1}) + log(a_{{\color{green}i}{\color{red}1}{\color{blue}1}}))+log(b_{{\color{red}1}{\color{blue}1}}("."))\\
 & = max_{1 \leq i \leq 4}(\delta_{5}({\color{green}i},{\color{red}1}) + log(P({\color{blue}\Gamma_{r}=0}/{\color{red}\Gamma_{r-1}=0},{\color{green}\Gamma_{r-2}=\gamma_{i}})))\\
&+log(P(V_{r}="."/{\color{blue}\Gamma_{r}=0},{\color{red}\Gamma_{r-1}=0}))\\
& = \underbrace{\delta_{5}({\color{green}1},{\color{red}1})}_{-38.13} +\underbrace{ log(P({\color{blue}\Gamma_{r}=0}/{\color{red}\Gamma_{r-1}=0},{\color{green}\Gamma_{r-2}=0}))}_{-0.832} \\
& +\underbrace{log(P(V_{r}="."/{\color{blue}\Gamma_{r}=0},{\color{red}\Gamma_{r-1}=0}))}_{-2.85} \\
& =\color{orange}-41.82
\end{split}
\end{equation}

\begin{table}[h]
\begin{center}
\begin{tabular}{ c c c c c c c c c }
  & . & \AR{زيت} & \AR{زيتون} & \AR{من} & \AR{النوعية} & \AR{الجيدة}& . \\   
 case1 & - & - & - & - & ${\color{green}\Gamma_{r-2}=0}$ & ${\color{red}\Gamma_{r-1}=0}$ & ${\color{blue}\Gamma_{r}=0}$ \\
    case2 & - & - & - & - & - & ${\color{red}\Gamma_{r-1}=0}$ & ${\color{blue}\Gamma_{r}=0}$
\end{tabular}
\caption{situations table for $\delta_{6}(1,1)$ for a second order HMM estimation with all the words are known . The probabilities $P({\color{blue}\Gamma_{r}=0}/{\color{red}\Gamma_{r-1}=0},{\color{green}\Gamma_{r-2}=0})$ and $P(V_{r}="."/{\color{blue}\Gamma_{r}=0},{\color{red}\Gamma_{r-1}=0})$ are describing existing cases from the tagged sentence, they are not parasite probabilities. }
\label{fig:tab1}
\end{center}
\end{table}

\begin{equation} \label{eq2}
\begin{split}
\delta_{6}({\color{red}2},{\color{blue}1}) & = max_{1 \leq i \leq 4}(\delta_{5}({\color{green}i},{\color{red}2}) + log(a_{{\color{green}i}{\color{red}2}{\color{blue}1}}))+log(b_{{\color{red}{\color{red}2}}{\color{blue}1}}("."))\\
 & = max_{1 \leq i \leq 4}(\delta_{5}({\color{green}i},{\color{red}2}) + log(P({\color{blue}\Gamma_{r}=0}/{\color{red}\Gamma_{r-1}=1},{\color{green}\Gamma_{r-2}=\gamma_{i}}))\\
&+log(P(V_{r}="."/{\color{blue}\Gamma_{r}=0},{\color{red}\Gamma_{r-1}=1}))\\
& = \underbrace{\delta_{5}({\color{green}1},{\color{red}2})}_{-38.43} +\underbrace{ log(P({\color{blue}\Gamma_{r}=0}/{\color{red}\Gamma_{r-1}=1},{\color{green}\Gamma_{r-2}=0}))}_{-0.351} \\
& +\underbrace{log(P(V_{r}="."/{\color{blue}\Gamma_{r}=0},{\color{red}\Gamma_{r-1}=1}))}_{-2.238} \\
& =\color{orange}-41.0199
\end{split}
\end{equation}

\begin{table}[h!]
\begin{center}
\begin{tabular}{ c c c c c c c c c }
  & . & \AR{زيت} & \AR{زيتون} & \AR{من} & \AR{النوعية} & \AR{الجيدة}& . \\   
 case1 & - & - & - & - & ${\color{green}\Gamma_{r-2}=0}$ & ${\color{red}\Gamma_{r-1}=1}$ & ${\color{blue}\Gamma_{r}=0}$ \\
    case2 & - & - & - & - & - & ${\color{red}\Gamma_{r-1}=1}$ & ${\color{blue}\Gamma_{r}=0}$
    
\end{tabular}
\caption{Situations table for $\delta_{6}(1,2)$ for a second order HMM estimation. The probabilities $P({\color{blue}\Gamma_{r}=0}/{\color{red}\Gamma_{r-1}=1},{\color{green}\Gamma_{r-2}=0})$ and $P(V_{r}="."/{\color{blue}\Gamma_{r}=0},{\color{red}\Gamma_{r-1}=1})$ are describing existing cases from the tagged sentence, supposing that the word \AR{الجيدة} is an ingredient . These probabilities are not parasite probabilities.}
\label{fig:tab2}
\end{center}
\end{table}

\begin{equation} \label{eq3}
\begin{split}
\delta_{6}({\color{red}3},{\color{blue}1}) & = max_{1 \leq i \leq 4}(\delta_{5}({\color{green}i},{\color{red}3}) + log(a_{{\color{green}i}{\color{red}3}{\color{blue}1}}))+log(b_{{\color{red}{\color{red}3}}{\color{blue}1}}("."))\\
 & = max_{1 \leq i \leq 4}(\delta_{5}({\color{green}i},{\color{red}3}) + log(P({\color{blue}\Gamma_{r}=0}/{\color{red}\Gamma_{r-1}=2},{\color{green}\Gamma_{r-2}=\gamma_{i}}))\\
&+log(P(V_{r}="."/{\color{blue}\Gamma_{r}=0},{\color{red}\Gamma_{r-1}=2}))\\
& = \underbrace{\delta_{5}({\color{green}2},{\color{red}3})}_{-38.7} +\underbrace{ log(P({\color{blue}\Gamma_{r}=0}/{\color{red}\Gamma_{r-1}=2},{\color{green}\Gamma_{r-2}=1}))}_{-0.043} \\
& +\underbrace{log(P(V_{r}="."/{\color{blue}\Gamma_{r}=0},{\color{red}\Gamma_{r-1}=2}))}_{-2.866} \\
& =\color{orange}-41,61
\end{split}
\end{equation}

\begin{table}[h!]
\begin{center}
\begin{tabular}{ c c c c c c c c c }
  & . & \AR{زيت} & \AR{زيتون} & \AR{من} & \AR{النوعية} & \AR{الجيدة}& . \\   
 case1 & - & - & - & - & ${\color{green}\cancel{\Gamma_{r-2}=1}}$ & ${\color{red}\Gamma_{r-1}=2}$ & ${\color{blue}\Gamma_{r}=0}$ \\
    case2 & - & - & - & - & - & ${\color{red}\Gamma_{r-1}=2}$ & ${\color{blue}\Gamma_{r}=0}$
    
\end{tabular}
\begin{tikzpicture}[overlay]
 \draw[red, line width=1.5pt] (-2.8,0.1) ellipse (3cm and 0.3cm);
 \node (A) at (0,0) {};
\node (B) at (1, 0) {};
\node[] at (2,0) {\makecell{parasite  \\ probability}};
\draw[->,red, to path={-| (\tikztotarget)}]
  (A) edge (B) ; 
\end{tikzpicture} 
\caption{Situations table for $\delta_{6}(3,1)$ for a second order HMM estimation. The probability $P({\color{blue}\Gamma_{r}=0}/{\color{red}\Gamma_{r-1}=2},{\color{green}\Gamma_{r-2}=1})$ is a parasite probability because ${\color{green}\Gamma_{r-2}=1}$ is not the real state of the word  \AR{النوعية} and $P(V_{r}="."/{\color{blue}\Gamma_{r}=0},{\color{red}\Gamma_{r-1}=1})$ is not a parasite probability because it is describing an existing case from the tagged sentence after considering the word \AR{الجيدة} as the second word of an ingredient.}
\label{fig:tab3}
\end{center}
\end{table}

\textbf{\textcolor{red}{after adding the sentence \AR{زيت زيتون من النوعية الجيدة} to the dataset:}}

sentence=.,\AR{زيت}(oil),\AR{زيتون}(olive),\AR{من}(of),\AR{النوعية}(quality),\AR{الجيدة}(good),.

ingredient state=[0,1,2,0,0,0,0]

ingredient to be extracted=\AR{زيت زيتون}

\begin{gather} \label{mat2}
	\delta_{6}(i,j)=\begin{bmatrix}
	\color{orange}-41.0035 & -45.7531
 & -53.3305
 & -53.6236
 \\
	\color{orange}-40.6117
 & -52.6603
 & -47.5296
 & -53.7664
 \\
	\color{orange}-41.6097
 & -53.0374
 & -53.4007
 & -51.8409
 \\	
	-55.7362
 &-56.3216
 & -55.9917
 & -55.8183
	\end{bmatrix}
	\end{gather}

\begin{equation} \label{eq4}
\begin{split}
\delta_{6}({\color{red}1},{\color{blue}1}) & = max_{1 \leq i \leq 4}(\delta_{5}({\color{green}i},{\color{red}1}) + log(a_{{\color{green}i}{\color{red}1}{\color{blue}1}}))+log(b_{{\color{red}1}{\color{blue}1}}("."))\\
 & = max_{1 \leq i \leq 4}(\delta_{5}({\color{green}i},{\color{red}1}) + log(P({\color{blue}\Gamma_{r}=0}/{\color{red}\Gamma_{r-1}=0},{\color{green}\Gamma_{r-2}=\gamma_{i}})))\\
&+log(P(V_{r}="."/{\color{blue}\Gamma_{r}=0},{\color{red}\Gamma_{r-1}=0}))\\
& = \underbrace{\delta_{5}({\color{green}1},{\color{red}1})}_{-37.314} +\underbrace{ log(P({\color{blue}\Gamma_{r}=0}/{\color{red}\Gamma_{r-1}=0},{\color{green}\Gamma_{r-2}=0}))}_{-0.831} \\
& +\underbrace{log(P(V_{r}="."/{\color{blue}\Gamma_{r}=0},{\color{red}\Gamma_{r-1}=0}))}_{-2.8567} \\
& =\color{orange}-41.003
\end{split}
\end{equation}

\begin{table}[h]
\begin{center}
\begin{tabular}{ c c c c c c c c c }
  & . & \AR{زيت} & \AR{زيتون} & \AR{من} & \AR{النوعية} & \AR{الجيدة}& . \\   
 case1 & - & - & - & - & ${\color{green}\Gamma_{r-2}=0}$ & ${\color{red}\Gamma_{r-1}=0}$ & ${\color{blue}\Gamma_{r}=0}$ \\
    case2 & - & - & - & - & - & ${\color{red}\Gamma_{r-1}=0}$ & ${\color{blue}\Gamma_{r}=0}$

\end{tabular}
\caption{situations table for $\delta_{6}({\color{red}1},{\color{blue}1})$ for a second order HMM estimation with all the words are known after adding the sentence \AR{زيت زيتون من النوعية الجيدة} to the dataset. The probabilities $P({\color{blue}\Gamma_{r}=0}/{\color{red}\Gamma_{r-1}=0},{\color{green}\Gamma_{r-2}=0})$ and $P(V_{r}="."/{\color{blue}\Gamma_{r}=0},{\color{red}\Gamma_{r-1}=0})$ are describing existing cases from the tagged sentence, they are not parasite probabilities. }
\label{fig:tab4}
\end{center}
\end{table}

\begin{equation} \label{eq5}
\begin{split}
\delta_{6}({\color{red}2},{\color{blue}1}) & = max_{1 \leq i \leq 4}(\delta_{5}({\color{green}i},{\color{red}2}) + log(a_{{\color{green}i}{\color{red}2}{\color{blue}1}}))+log(b_{{\color{red}{\color{red}2}}{\color{blue}1}}("."))\\
 & = max_{1 \leq i \leq 4}(\delta_{5}({\color{green}i},{\color{red}2}) + log(P({\color{blue}\Gamma_{r}=0}/{\color{red}\Gamma_{r-1}=1},{\color{green}\Gamma_{r-2}=\gamma_{i}}))\\
&+log(P(V_{r}="."/{\color{blue}\Gamma_{r}=0},{\color{red}\Gamma_{r-1}=1}))\\
& = \underbrace{\delta_{5}({\color{green}1},{\color{red}2})}_{-38.021} +\underbrace{ log(P({\color{blue}\Gamma_{r}=0}/{\color{red}\Gamma_{r-1}=1},{\color{green}\Gamma_{r-2}=0}))}_{-0.3517} \\
& +\underbrace{log(P(V_{r}="."/{\color{blue}\Gamma_{r}=0},{\color{red}\Gamma_{r-1}=1}))}_{-2.238} \\
& =\color{orange}-40.61
\end{split}
\end{equation}

\begin{table}[h!]
\begin{center}
\begin{tabular}{ c c c c c c c c c }
  & . & \AR{زيت} & \AR{زيتون} & \AR{من} & \AR{النوعية} & \AR{الجيدة}& . \\   
 case1 & - & - & - & - & ${\color{green}\Gamma_{r-2}=0}$ & ${\color{red}\Gamma_{r-1}=1}$ & ${\color{blue}\Gamma_{r}=0}$ \\
    case2 & - & - & - & - & - & ${\color{red}\Gamma_{r-1}=1}$ & ${\color{blue}\Gamma_{r}=0}$
    
\end{tabular}
\caption{Situations table for $\delta_{6}({\color{red}2},{\color{blue}1})$ for a second order HMM estimation with all the words are known after adding the sentence \AR{زيت زيتون من النوعية الجيدة} to the dataset. The probabilities $P({\color{blue}\Gamma_{r}=0}/{\color{red}\Gamma_{r-1}=1},{\color{green}\Gamma_{r-2}=0})$ and $P(V_{r}="."/{\color{blue}\Gamma_{r}=0},{\color{red}\Gamma_{r-1}=1})$ are describing existing cases from the tagged sentence, supposing that the word \AR{الجيدة} is an ingredient . These probabilities are not parasite probabilities.}
\label{fig:tab5}
\end{center}
\end{table}

\begin{equation} \label{eq6}
\begin{split}
\delta_{6}({\color{red}3},{\color{blue}1}) & = max_{1 \leq i \leq 4}(\delta_{5}({\color{green}i},{\color{red}3}) + log(a_{{\color{green}i}{\color{red}3}{\color{blue}1}}))+log(b_{{\color{red}{\color{red}3}}{\color{blue}1}}("."))\\
 & = max_{1 \leq i \leq 4}(\delta_{5}({\color{green}i},{\color{red}3}) + log(P({\color{blue}\Gamma_{r}=0}/{\color{red}\Gamma_{r-1}=2},{\color{green}\Gamma_{r-2}=\gamma_{i}}))\\
&+log(P(V_{r}="."/{\color{blue}\Gamma_{r}=0},{\color{red}\Gamma_{r-1}=2}))\\
& = \underbrace{\delta_{5}({\color{green}2},{\color{red}3})}_{-38.69} +\underbrace{ log(P({\color{blue}\Gamma_{r}=0}/{\color{red}\Gamma_{r-1}=2},{\color{green}\Gamma_{r-2}=1}))}_{-0.0437} \\
& +\underbrace{log(P(V_{r}="."/{\color{blue}\Gamma_{r}=0},{\color{red}\Gamma_{r-1}=2}))}_{-2.866} \\
& =\color{orange}-41.609
\end{split}
\end{equation}

\begin{table}[h!]
\begin{center}
\begin{tabular}{ c c c c c c c c c }
  & . & \AR{زيت} & \AR{زيتون} & \AR{من} & \AR{النوعية} & \AR{الجيدة}& . \\   
 case1 & - & - & - & - & ${\color{green}\cancel{\Gamma_{r-2}=1}}$ & ${\color{red}\Gamma_{r-1}=2}$ & ${\color{blue}\Gamma_{r}=0}$ \\
    case2 & - & - & - & - & - & ${\color{red}\Gamma_{r-1}=2}$ & ${\color{blue}\Gamma_{r}=0}$
    
\end{tabular}
\begin{tikzpicture}[overlay]
 \draw[red, line width=1.5pt] (-2.8,0.1) ellipse (3cm and 0.3cm);
 \node (A) at (0,0) {};
\node (B) at (1, 0) {};
\node[] at (2,0) {\makecell{parasite  \\ probability}};
\draw[->,red, to path={-| (\tikztotarget)}]
  (A) edge (B) ; 
\end{tikzpicture} 
\caption{Situations table for $\delta_{6}({\color{red}3},{\color{blue}1})$ for a second order HMM estimation with all the words are known after adding the sentence \AR{زيت زيتون من النوعية الجيدة} to the dataset. The probability $P({\color{blue}\Gamma_{r}=0}/{\color{red}\Gamma_{r-1}=2},{\color{green}\Gamma_{r-2}=1})$ is a parasite probability because ${\color{green}\Gamma_{r-2}=1}$ is not the real state of the word  \AR{النوعية} and $P(V_{r}="."/{\color{blue}\Gamma_{r}=0},{\color{red}\Gamma_{r-1}=1})$ is not a parasite probability because it is describing an existing case from the tagged sentence after considering the word \AR{الجيدة} as the second word of an ingredient.}
\label{fig:tab6}
\end{center}
\end{table}

\textbf{\textcolor{red}{after adding the sentence \AR{كعبة طماطم من الحجم الكبير} to the dataset:}}

sentence=.,\AR{زيت}(oil),\AR{زيتون}(olive),\AR{من}(of),\AR{النوعية}(quality),\AR{الجيدة}(good),.

ingredient state=[0,1,2,0,0,0,0]

ingredient to be extracted=\AR{زيت زيتون}

\begin{gather} \label{mat3}
	\delta_{6}(i,j)=\begin{bmatrix}
	\color{orange}-41.8278
 & -46.5781
 & -53.3457
 & -54.4495
 \\
	\color{orange}-41.0313
 & -53.0804
 & -47.9504
 & -54.1875
 \\
	\color{orange}-41.6249
 & -53.0533
 & -53.4171
 & -51.8577
 \\	
	-55.8946
 &-56.4807
 & -56.1513
 & -55.9783
	\end{bmatrix}
	\end{gather}

\begin{equation} \label{eq7}
\begin{split}
\delta_{6}({\color{red}1},{\color{blue}1}) & = max_{1 \leq i \leq 4}(\delta_{5}({\color{green}i},{\color{red}1}) + log(a_{{\color{green}i}{\color{red}1}{\color{blue}1}}))+log(b_{{\color{red}1}{\color{blue}1}}("."))\\
 & = max_{1 \leq i \leq 4}(\delta_{5}({\color{green}i},{\color{red}1}) + log(P({\color{blue}\Gamma_{r}=0}/{\color{red}\Gamma_{r-1}=0},{\color{green}\Gamma_{r-2}=\gamma_{i}})))\\
&+log(P(V_{r}="."/{\color{blue}\Gamma_{r}=0},{\color{red}\Gamma_{r-1}=0}))\\
& = \underbrace{\delta_{5}({\color{green}1},{\color{red}1})}_{-38.138} +\underbrace{ log(P({\color{blue}\Gamma_{r}=0}/{\color{red}\Gamma_{r-1}=0},{\color{green}\Gamma_{r-2}=0}))}_{-0.831} \\
& +\underbrace{log(P(V_{r}="."/{\color{blue}\Gamma_{r}=0},{\color{red}\Gamma_{r-1}=0}))}_{-2.857} \\
& =\color{orange}-41.827
\end{split}
\end{equation}

\begin{table}[h]
\begin{center}
\begin{tabular}{ c c c c c c c c c }
  & . & \AR{زيت} & \AR{زيتون} & \AR{من} & \AR{النوعية} & \AR{الجيدة}& . \\   
 case1 & - & - & - & - & ${\color{green}\Gamma_{r-2}=0}$ & ${\color{red}\Gamma_{r-1}=0}$ & ${\color{blue}\Gamma_{r}=0}$ \\
    case2 & - & - & - & - & - & ${\color{red}\Gamma_{r-1}=0}$ & ${\color{blue}\Gamma_{r}=0}$

\end{tabular}
\caption{situations table for $\delta_{6}({\color{red}1},{\color{blue}1})$ for a second order HMM estimation with all the words are known after adding the sentence \AR{كعبة طماطم من الحجم الكبير} to the dataset. The probabilities $P({\color{blue}\Gamma_{r}=0}/{\color{red}\Gamma_{r-1}=0},{\color{green}\Gamma_{r-2}=0})$ and $P(V_{r}="."/{\color{blue}\Gamma_{r}=0},{\color{red}\Gamma_{r-1}=0})$ are describing existing cases from the tagged sentence, they are not parasite probabilities. }
\label{fig:tab7}
\end{center}
\end{table}

\begin{equation} \label{eq8}
\begin{split}
\delta_{6}({\color{red}2},{\color{blue}1}) & = max_{1 \leq i \leq 4}(\delta_{5}({\color{green}i},{\color{red}2}) + log(a_{{\color{green}i}{\color{red}2}{\color{blue}1}}))+log(b_{{\color{red}{\color{red}2}}{\color{blue}1}}("."))\\
 & = max_{1 \leq i \leq 4}(\delta_{5}({\color{green}i},{\color{red}2}) + log(P({\color{blue}\Gamma_{r}=0}/{\color{red}\Gamma_{r-1}=1},{\color{green}\Gamma_{r-2}=\gamma_{i}}))\\
&+log(P(V_{r}="."/{\color{blue}\Gamma_{r}=0},{\color{red}\Gamma_{r-1}=1}))\\
& = \underbrace{\delta_{5}({\color{green}1},{\color{red}2})}_{-38.44} +\underbrace{ log(P({\color{blue}\Gamma_{r}=0}/{\color{red}\Gamma_{r-1}=1},{\color{green}\Gamma_{r-2}=0}))}_{-0.3517} \\
& +\underbrace{log(P(V_{r}="."/{\color{blue}\Gamma_{r}=0},{\color{red}\Gamma_{r-1}=1}))}_{-2.239} \\
& =\color{orange}-41.0312
\end{split}
\end{equation}

\begin{table}[h!]
\begin{center}
\begin{tabular}{ c c c c c c c c c }
  & . & \AR{زيت} & \AR{زيتون} & \AR{من} & \AR{النوعية} & \AR{الجيدة}& . \\   
 case1 & - & - & - & - & ${\color{green}\Gamma_{r-2}=0}$ & ${\color{red}\Gamma_{r-1}=1}$ & ${\color{blue}\Gamma_{r}=0}$ \\
    case2 & - & - & - & - & - & ${\color{red}\Gamma_{r-1}=1}$ & ${\color{blue}\Gamma_{r}=0}$

\end{tabular}
\caption{Situations table for $\delta_{6}({\color{red}2},{\color{blue}1})$ for a second order HMM estimation  with all the words are known after adding the sentence \AR{كعبة طماطم من الحجم الكبير} to the dataset. The probabilities $P({\color{blue}\Gamma_{r}=0}/{\color{red}\Gamma_{r-1}=1},{\color{green}\Gamma_{r-2}=0})$ and $P(V_{r}="."/{\color{blue}\Gamma_{r}=0},{\color{red}\Gamma_{r-1}=1})$ are describing existing cases from the tagged sentence, supposing that the word \AR{الجيدة} is an ingredient . These probabilities are not parasite probabilities.}
\label{fig:tab8}
\end{center}
\end{table}

\begin{equation} \label{eq9}
\begin{split}
\delta_{6}({\color{red}3},{\color{blue}1}) & = max_{1 \leq i \leq 4}(\delta_{5}({\color{green}i},{\color{red}3}) + log(a_{{\color{green}i}{\color{red}3}{\color{blue}1}}))+log(b_{{\color{red}{\color{red}3}}{\color{blue}1}}("."))\\
 & = max_{1 \leq i \leq 4}(\delta_{5}({\color{green}i},{\color{red}3}) + log(P({\color{blue}\Gamma_{r}=0}/{\color{red}\Gamma_{r-1}=2},{\color{green}\Gamma_{r-2}=\gamma_{i}}))\\
&+log(P(V_{r}="."/{\color{blue}\Gamma_{r}=0},{\color{red}\Gamma_{r-1}=2}))\\
& = \underbrace{\delta_{5}({\color{green}2},{\color{red}3})}_{-38.713} +\underbrace{ log(P({\color{blue}\Gamma_{r}=0}/{\color{red}\Gamma_{r-1}=2},{\color{green}\Gamma_{r-2}=1}))}_{-0.0437} \\
& +\underbrace{log(P(V_{r}="."/{\color{blue}\Gamma_{r}=0},{\color{red}\Gamma_{r-1}=2}))}_{-2.867} \\
& =\color{orange}-41.6249
\end{split}
\end{equation}

\begin{table}[h!]
\begin{center}
\begin{tabular}{ c c c c c c c c c }
  & . & \AR{زيت} & \AR{زيتون} & \AR{من} & \AR{النوعية} & \AR{الجيدة}& . \\   
 case1 & - & - & - & - & ${\color{green}\cancel{\Gamma_{r-2}=1}}$ & ${\color{red}\Gamma_{r-1}=2}$ & ${\color{blue}\Gamma_{r}=0}$ \\
    case2 & - & - & - & - & - & ${\color{red}\Gamma_{r-1}=2}$ & ${\color{blue}\Gamma_{r}=0}$
    
\end{tabular}
\begin{tikzpicture}[overlay]
 \draw[red, line width=1.5pt] (-2.8,0.1) ellipse (3cm and 0.3cm);
 \node (A) at (0,0) {};
\node (B) at (1, 0) {};
\node[] at (2,0) {\makecell{parasite  \\ probability}};
\draw[->,red, to path={-| (\tikztotarget)}]
  (A) edge (B) ; 
\end{tikzpicture} 
\caption{Situations table for $\delta_{6}({\color{red}3},{\color{blue}1})$ for a second order HMM estimation with all the words are known after adding the sentence \AR{كعبة طماطم من الحجم الكبير} to the dataset. The probability $P({\color{blue}\Gamma_{r}=0}/{\color{red}\Gamma_{r-1}=2},{\color{green}\Gamma_{r-2}=1})$ is a parasite probability because ${\color{green}\Gamma_{r-2}=1}$ is not the real state of the word  \AR{النوعية} and $P(V_{r}="."/{\color{blue}\Gamma_{r}=0},{\color{red}\Gamma_{r-1}=2})$ is not a parasite probability because it is describing an existing case from the tagged sentence after considering the word \AR{الجيدة} as the second word of an ingredient.}
\label{fig:tab9}
\end{center}
\end{table}

For a simple double agent Viterbi algorithm without external knowledge transfer, the parasite probabilities can easily infiltrate the most probable cases inside the equations of a state matrix . For example , for the sentence ' \AR{الجيدة} \AR{النوعية} \AR{من} \AR{زيتون} \AR{زيت}' , the calculation of the element of the state matrix $\psi_{6}(7,1)$ in the {\color{blue}equation \ref{form1}} to estimate the ingredient state of the token \AR{الجيدة} shows three parasite probabilities, $P({\color{blue}\Gamma_{r}=0}/\Gamma_{r-1}=0,{\color{red}T_{r-1}="\textbf{\AR{نعت}}"})$ in {\color{blue}table \ref{fig:tab10}} and  $P({\color{blue}\Gamma_{r}=1}/\Gamma_{r-1}=0,{\color{red}T_{r-1}="\textbf{\AR{نعت}}"})$ in {\color{blue}table \ref{fig:tab11}} and  $P({\color{blue}\Gamma_{r}=2}/\Gamma_{r-1}=3,{\color{red}T_{r-1}="\textbf{\AR{نعت}}"})$ in {\color{blue}table \ref{fig:tab12}} are three parasite probabilities appearing when we assume that the token \AR{الجيدة} in position $r$ is considered as respectively a non ingredient ${\color{blue}\Gamma_{r}=0}$ or as an ingredient ${\color{blue}\Gamma_{r}=1}$ or a second word of an ingredient ${\color{blue}\Gamma_{r}=2}$ because the token in position $r-1$ haven't a POS tag of type \AR{نعت}. In order to avoid the creation of parasite probability , we decided to use the information obtained in the first layer in calculating the elements of the state matrix , this operation is called external knowledge transfer.

\newpage

{\color{blue}Estimating an ingredient state using a simple double agent Viterbi algorithm without unknown words consideration without external knowledge transfer:}

sentence=.,\AR{زيت}(oil),\AR{زيتون}(olive),\AR{من},\AR{النوعية},\AR{الجيدة},.

ingredient state=[0,1,2,0,0,0,0]

POS tags=[.,C,D,G,E,F,.] coded as [4,2,3,9,6,7,4]

ingredient to be extracted=\AR{زيت زيتون}

\begin{gather}\label{mat4}
	\psi_{6}(i,j)=\begin{bmatrix}
	0 & 0 & 1 & 1 \\
	3 & 3 & 1 & 3 \\
	2 & 0 & 3 & 3 \\	
	0 &0 & 1 & 1 \\
          0 & 0 & 0 & 0 \\
         0 & 3 & 1 & 3 \\
          0 & 0 & 0 & 0 \\
	3 & 3 & 3 & 3 \\
	3 & 3 & 3 & 3 \\	
	3 &3 & 3 & 3 \\
           3 & 3& 3 & 3 \\
	1 & 3 & 3 & 3 \\
	3 & 3 & 3 & 3 \\	
	3 &3 & 3 & 3 
	\end{bmatrix}
	\end{gather}

\begin{equation} \label{form1}
\begin{split}
\psi_{6}(7,1) & = argmax_{1 \leq i \leq 4}(\delta_{5}(7,i) + log(a_{7i1}))
\end{split}
\end{equation}

\begin{equation} \label{eq10}
\begin{split}
\delta_{5}({\color{red}7},{\color{blue}1}) + log(a_{{\color{red}7}{\color{blue}1}{\color{green}1}}) & =max_{1 \leq i \leq 4}(\delta_{4}({\color{red}7},i)+log(a_{{\color{red}7}i{\color{blue}1}}))+\lambda_{max} log(b_{{\color{red}7}{\color{blue}1}}("\textbf{\AR{الجيدة}}"))+ log(a_{{\color{red}7}{\color{blue}1}1}) \\
&  = \underbrace{\delta_{4}({\color{red}7},1)}_{-134.9168}+\underbrace{log(P({\color{blue}\Gamma_{r}=0}/\Gamma_{r-1}=0,{\color{red}T_{r-1}="\textbf{\AR{نعت}}"}))}_{-2.717} \\
& +\underbrace{\lambda_{max} log(P(V_{r}="\textbf{\AR{الجيدة}}"/{\color{blue}\Gamma_{r}=0},{\color{red}T_{r}="\textbf{\AR{نعت}}"})}_{-36.629} \\
& + \underbrace{log(P({\color{green}\Gamma_{r}=0}/{\color{blue}\Gamma_{r-1}=0},{\color{red}T_{r-1}="\textbf{\AR{نعت}}"}))}_{-2,717}\\
& = -176.982
\end{split}
\end{equation}

\begin{table}[h]
\begin{center}
\begin{tabular}{ c c c c c c c c c }
  & . & \AR{زيت} & \AR{زيتون} & \AR{من} & \AR{النوعية} & \AR{الجيدة}& . \\   
 case1 & - & - & - & - & \makecell{${\color{red}\cancel{T_{r-1}="\textbf{\AR{نعت}}"}}$ \\ ${\color{black}\Gamma_{r-1}=0}$} & ${\color{blue}\Gamma_{r}=0}$ & - \\
    case2 & - & - & - & - & - & \makecell{${\color{red}T_{r}="\textbf{\AR{نعت}}"}$ \\ ${\color{blue}\Gamma_{r}=0}$} & - \\
    case3 & - & - & - & - & - & \makecell{${\color{red}T_{r-1}="\textbf{\AR{نعت}}"}$ \\ ${\color{blue}\Gamma_{r-1}=0}$} & ${\color{green}\Gamma_{r}=0}$
\end{tabular}
\begin{tikzpicture}[overlay]
 \draw[red, line width=1.5pt] (-3.8,0.7) ellipse (3cm and 0.5cm);
 
 \node (A) at (-1.3, 0.5) {};
\node (B) at (1.8, -0.2) {};

\draw[->,red, to path={-| (\tikztotarget)}]
  (A) edge (B) ;

  \node[] at (1.5,-0.5) {\makecell{parasite probability}};
\end{tikzpicture} 
\caption{situations table for $\delta_{5}({\color{red}7},{\color{blue}1}) + log(a_{{\color{red}7}{\color{blue}1}{\color{green}1}})$ for a second order HMM estimation, the probability $P({\color{blue}\Gamma_{r}=0}/\Gamma_{r-1}=0,{\color{red}T_{r-1}="\textbf{\AR{نعت}}"})$ is a parasite probability because the POS tag of the word  \AR{النوعية} is \AR{منعوت} and not \AR{نعت}}
\label{fig:tab10}
\end{center}
\end{table}

\begin{equation} \label{eq11}
\begin{split}
\delta_{5}({\color{red}7},{\color{blue}2}) + log(a_{{\color{red}7}{\color{blue}2}{\color{green}1}}) & =max_{1 \leq i \leq 4}(\delta_{4}({\color{red}7},i)+log(a_{{\color{red}7}i{\color{blue}2}}))+\lambda_{max} log(b_{{\color{red}7}{\color{blue}2}}("\textbf{\AR{الجيدة}}"))+ log(a_{{\color{red}7}{\color{blue}2}{\color{green}1}}) \\
&  = \underbrace{\delta_{4}({\color{red}7},1)}_{-134.916}+\underbrace{log(P({\color{blue}\Gamma_{r}=1}/\Gamma_{r-1}=0,{\color{red}T_{r-1}="\textbf{\AR{نعت}}"}))}_{-3.487} \\
& +\underbrace{\lambda_{max} log(P(V_{r}="\textbf{\AR{الجيدة}}"/{\color{blue}\Gamma_{r}=1},{\color{red}T_{r}="\textbf{\AR{نعت}}"})}_{-38.016} \\
& + \underbrace{log(P({\color{green}\Gamma_{r}=0}/{\color{blue}\Gamma_{r-1}=1},{\color{red}T_{r-1}="\textbf{\AR{نعت}}"}))}_{-6.583}\\
& = -183
\end{split}
\end{equation}

\begin{table}[h]
\begin{center}
\begin{tabular}{ c c c c c c c c c }
  & . & \AR{زيت} & \AR{زيتون} & \AR{من} & \AR{النوعية} & \AR{الجيدة}& . \\   
 case1 & - & - & - & - & \makecell{${\color{red}\cancel{T_{r-1}="\textbf{\AR{نعت}}"}}$ \\ ${\color{black}\Gamma_{r-1}=0}$} & ${\color{blue}\Gamma_{r}=1}$ & - \\
    case2 & - & - & - & - & - & \makecell{${\color{red}T_{r}="\textbf{\AR{نعت}}"}$ \\ ${\color{blue}\Gamma_{r}=1}$} & - \\
    case3 & - & - & - & - & - & \makecell{${\color{red}T_{r-1}="\textbf{\AR{نعت}}"}$ \\ ${\color{blue}\Gamma_{r-1}=1}$} & ${\color{green}\Gamma_{r}=0}$
\end{tabular}
\begin{tikzpicture}[overlay]
 \draw[red, line width=1.5pt] (-3.8,0.7) ellipse (3cm and 0.5cm);
 \node (A) at (-1, 0.5) {};
\node (B) at (0, 0.5) {};
\node[] at (1.2,0.5) {\makecell{parasite  \\ probability}};
\draw[->,red, to path={-| (\tikztotarget)}]
  (A) edge (B) ; 
\end{tikzpicture} 
\caption{situations table for $\delta_{5}({\color{red}7},{\color{blue}2}) + log(a_{{\color{red}7}{\color{blue}2}{\color{green}1}})$ for a second order HMM estimation, the probability $P({\color{blue}\Gamma_{r}=1}/\Gamma_{r-1}=0,{\color{red}T_{r-1}="\textbf{\AR{نعت}}"})$ is a parasite probability because the POS tag of the word \AR{النوعية} is \AR{منعوت} and not \AR{نعت}}
\label{fig:tab11}
\end{center}
\end{table}

\begin{equation} \label{eq12}
\begin{split}
\delta_{5}({\color{red}7},{\color{blue}3}) + log(a_{{\color{red}7}{\color{blue}3}{\color{green}1}}) & =max_{1 \leq i \leq 4}(\delta_{4}({\color{red}7},i)+log(a_{{\color{red}7}i{\color{blue}3}}))+\lambda_{max} log(b_{{\color{red}7}{\color{blue}3}}("\textbf{\AR{الجيدة}}"))+ log(a_{{\color{red}7}{\color{blue}3}{\color{green}1}}) \\
&  = \underbrace{\delta_{4}({\color{red}7},3)}_{-137.789}+\underbrace{log(P({\color{blue}\Gamma_{r}=2}/\Gamma_{r-1}=3,{\color{red}T_{r-1}="\textbf{\AR{نعت}}"}))}_{-4.077} \\
& +\underbrace{\lambda_{max} log(P(V_{r}="\textbf{\AR{الجيدة}}"/{\color{blue}\Gamma_{r}=2},{\color{red}T_{r}="\textbf{\AR{نعت}}"})}_{-37.541} \\
& + \underbrace{log(P({\color{green}\Gamma_{r}=0}/{\color{blue}\Gamma_{r-1}=2},{\color{red}T_{r-1}="\textbf{\AR{نعت}}"}))}_{-0.851}\\
& = -180.259
\end{split}
\end{equation}

\begin{table}[h]
\begin{center}
\begin{tabular}{ c c c c c c c c c }
  & . & \AR{زيت} & \AR{زيتون} & \AR{من} & \AR{النوعية} & \AR{الجيدة}& . \\   
 case1 & - & - & - & - & \makecell{${\color{red}\cancel{T_{r-1}="\textbf{\AR{نعت}}"}}$ \\ \cancel{${\color{black}\Gamma_{r-1}=3}$}} & ${\color{blue}\Gamma_{r}=2}$ & - \\
    case2 & - & - & - & - & - & \makecell{${\color{red}T_{r}="\textbf{\AR{نعت}}"}$ \\ ${\color{blue}\Gamma_{r}=2}$} & - \\
    case3 & - & - & - & - & - & \makecell{${\color{red}T_{r-1}="\textbf{\AR{نعت}}"}$ \\ ${\color{blue}\Gamma_{r-1}=2}$} & ${\color{green}\Gamma_{r}=0}$
\end{tabular}
\begin{tikzpicture}[overlay]
 \draw[red, line width=1.5pt] (-3.8,0.7) ellipse (3cm and 0.5cm);
 \node (A) at (-1, 0.5) {};
\node (B) at (0, 0.5) {};
\node[] at (1.2,0.5) {\makecell{parasite  \\ probability}};
\draw[->,red, to path={-| (\tikztotarget)}]
  (A) edge (B) ; 
\end{tikzpicture} 
\caption{situations table for $\delta_{5}({\color{red}7},{\color{blue}3}) + log(a_{{\color{red}7}{\color{blue}3}{\color{green}1}})$ for a second order HMM estimation, the probability $P({\color{blue}\Gamma_{r}=2}/\Gamma_{r-1}=3,{\color{red}T_{r-1}="\textbf{\AR{نعت}}"})$ is a parasite probability because the POS tag of the word \AR{النوعية} is \AR{منعوت} and not \AR{نعت}}
\label{fig:tab12}
\end{center}
\end{table}

\newpage

\textbf{\textcolor{red}{after adding the sentence \AR{زيت زيتون من النوعية الجيدة} to the dataset:}}

sentence=.,\AR{زيت}(oil),\AR{زيتون}(olive),\AR{من},\AR{النوعية},\AR{الجيدة},.

ingredient state=[0,1,2,0,0,0,0]

POS tags=[.,C,D,G,E,F,.] coded as [4,2,3,9,6,7,4]

ingredient to be extracted=\AR{زيت زيتون}

\begin{gather}\label{mat5}
	\psi_{6}(i,j)=\begin{bmatrix}
	0 & 0 & 1 & 1 \\
	3 & 3 & 1 & 3 \\
	2 & 0 & 3 & 3 \\	
	0 &0 & 1 & 1 \\
          0 & 0 & 0 & 0 \\
         0 & 3 & 1 & 3 \\
          0 & 0 & 0 & 0 \\
	3 & 3 & 3 & 3 \\
	3 & 3 & 3 & 3 \\	
	3 &3 & 3 & 3 \\
           3 & 3& 3 & 3 \\
	1 & 3 & 3 & 3 \\
	3 & 3 & 3 & 3 \\	
	3 &3 & 3 & 3 
	\end{bmatrix}
	\end{gather}

\begin{equation*} 
\begin{split}
\psi_{6}(7,1) & = argmax_{1 \leq i \leq 4}(\delta_{5}(7,i) + log(a_{7i1}))
\end{split}
\end{equation*}

\begin{equation} \label{eq13}
\begin{split}
\delta_{5}({\color{red}7},{\color{blue}1}) + log(a_{{\color{red}7}{\color{blue}1}{\color{green}1}}) & =max_{1 \leq i \leq 4}(\delta_{4}({\color{red}7},i)+log(a_{{\color{red}7}i{\color{blue}1}}))+\lambda_{max} log(b_{{\color{red}7}{\color{blue}1}}("\textbf{\AR{الجيدة}}"))+ log(a_{{\color{red}7}{\color{blue}1}1}) \\
&  = \underbrace{\delta_{4}({\color{red}7},1)}_{-134.9174}+\underbrace{log(P({\color{blue}\Gamma_{r}=0}/\Gamma_{r-1}=0,{\color{red}T_{r-1}="\textbf{\AR{نعت}}"}))}_{-2.7157} \\
& +\underbrace{\lambda_{max} log(P(V_{r}="\textbf{\AR{الجيدة}}"/{\color{blue}\Gamma_{r}=0},{\color{red}T_{r}="\textbf{\AR{نعت}}"})}_{-35.0086} \\
& + \underbrace{log(P({\color{green}\Gamma_{r}=0}/{\color{blue}\Gamma_{r-1}=0},{\color{red}T_{r-1}="\textbf{\AR{نعت}}"}))}_{-2.7157}\\
& = -175.357
\end{split}
\end{equation}

\begin{table}[h]

\begin{center}
\begin{tabular}{ c c c c c c c c c }
  & . & \AR{زيت} & \AR{زيتون} & \AR{من} & \AR{النوعية} & \AR{الجيدة}& . \\   
 case1 & - & - & - & - & \makecell{${\color{red}\cancel{T_{r-1}="\textbf{\AR{نعت}}"}}$ \\ ${\color{black}\Gamma_{r-1}=0}$} & ${\color{blue}\Gamma_{r}=0}$ & - \\
    case2 & - & - & - & - & - & \makecell{${\color{red}T_{r}="\textbf{\AR{نعت}}"}$ \\ ${\color{blue}\Gamma_{r}=0}$} & - \\
    case3 & - & - & - & - & - & \makecell{${\color{red}T_{r-1}="\textbf{\AR{نعت}}"}$ \\ ${\color{blue}\Gamma_{r-1}=0}$} & ${\color{green}\Gamma_{r}=0}$
\end{tabular}
\begin{tikzpicture}[overlay]
 \draw[red, line width=1.5pt] (-3.8,0.7) ellipse (3cm and 0.5cm);
 
 \node (A) at (-1.3, 0.5) {};
\node (B) at (1.8, -0.2) {};

\draw[->,red, to path={-| (\tikztotarget)}]
  (A) edge (B) ;

  \node[] at (1.5,-0.5) {\makecell{parasite probability}};
\end{tikzpicture} 
\caption{situations table for $\delta_{5}({\color{red}7},{\color{blue}1}) + log(a_{{\color{red}7}{\color{blue}1}{\color{green}1}})$ for a second order HMM estimation after adding the sentence \AR{زيت زيتون من النوعية الجيدة} to the dataset , the probability $P({\color{blue}\Gamma_{r}=0}/\Gamma_{r-1}=0,{\color{red}T_{r-1}="\textbf{\AR{نعت}}"})$ is a parasite probability because the POS tag of the word  \AR{النوعية} is \AR{منعوت} and not \AR{نعت}}
\label{fig:tab13}
\end{center}

\end{table}

\begin{equation} \label{eq14}
\begin{split}
\delta_{5}({\color{red}7},{\color{blue}2}) + log(a_{{\color{red}7}{\color{blue}2}{\color{green}1}}) & =max_{1 \leq i \leq 4}(\delta_{4}({\color{red}7},i)+log(a_{{\color{red}7}i{\color{blue}2}}))+\lambda_{max} log(b_{{\color{red}7}{\color{blue}2}}("\textbf{\AR{الجيدة}}"))+ log(a_{{\color{red}7}{\color{blue}2}{\color{green}1}}) \\
&  = \underbrace{\delta_{4}({\color{red}7},1)}_{-134.917}+\underbrace{log(P({\color{blue}\Gamma_{r}=1}/\Gamma_{r-1}=0,{\color{red}T_{r-1}="\textbf{\AR{نعت}}"}))}_{-3.488} \\
& +\underbrace{\lambda_{max} log(P(V_{r}="\textbf{\AR{الجيدة}}"/{\color{blue}\Gamma_{r}=1},{\color{red}T_{r}="\textbf{\AR{نعت}}"})}_{-38.0165} \\
& + \underbrace{log(P({\color{green}\Gamma_{r}=0}/{\color{blue}\Gamma_{r-1}=1},{\color{red}T_{r-1}="\textbf{\AR{نعت}}"}))}_{-6.5843}\\
& = -183.006
\end{split}
\end{equation}

\begin{table}[h]
\begin{center}
\begin{tabular}{ c c c c c c c c c }
  & . & \AR{زيت} & \AR{زيتون} & \AR{من} & \AR{النوعية} & \AR{الجيدة}& . \\   
 case1 & - & - & - & - & \makecell{${\color{red}\cancel{T_{r-1}="\textbf{\AR{نعت}}"}}$ \\ ${\color{black}\Gamma_{r-1}=0}$} & ${\color{blue}\Gamma_{r}=1}$ & - \\
    case2 & - & - & - & - & - & \makecell{${\color{red}T_{r}="\textbf{\AR{نعت}}"}$ \\ ${\color{blue}\Gamma_{r}=1}$} & - \\
    case3 & - & - & - & - & - & \makecell{${\color{red}T_{r-1}="\textbf{\AR{نعت}}"}$ \\ ${\color{blue}\Gamma_{r-1}=1}$} & ${\color{green}\Gamma_{r}=0}$
\end{tabular}
\begin{tikzpicture}[overlay]
 \draw[red, line width=1.5pt] (-3.8,0.7) ellipse (3cm and 0.5cm);
 \node (A) at (-1, 0.5) {};
\node (B) at (0, 0.5) {};
\node[] at (1.2,0.5) {\makecell{parasite  \\ probability}};
\draw[->,red, to path={-| (\tikztotarget)}]
  (A) edge (B) ; 
\end{tikzpicture} 
\caption{situations table for $\delta_{5}({\color{red}7},{\color{blue}2}) + log(a_{{\color{red}7}{\color{blue}2}{\color{green}1}})$ for a second order HMM estimation after adding the sentence \AR{زيت زيتون من النوعية الجيدة} to the dataset, the probability $P({\color{blue}\Gamma_{r}=1}/\Gamma_{r-1}=0,{\color{red}T_{r-1}="\textbf{\AR{نعت}}"})$ is a parasite probability because the POS tag of the word \AR{النوعية} is \AR{منعوت} and not \AR{نعت}}
\label{fig:tab14}
\end{center}
\end{table}

\begin{equation} \label{eq15}
\begin{split}
\delta_{5}({\color{red}7},{\color{blue}3}) + log(a_{{\color{red}7}{\color{blue}3}{\color{green}1}}) & =max_{1 \leq i \leq 4}(\delta_{4}({\color{red}7},i)+log(a_{{\color{red}7}i{\color{blue}3}}))+\lambda_{max} log(b_{{\color{red}7}{\color{blue}3}}("\textbf{\AR{الجيدة}}"))+ log(a_{{\color{red}7}{\color{blue}3}{\color{green}1}}) \\
&  = \underbrace{\delta_{4}({\color{red}7},3)}_{-137.7902}+\underbrace{log(P({\color{blue}\Gamma_{r}=2}/\Gamma_{r-1}=3,{\color{red}T_{r-1}="\textbf{\AR{نعت}}"}))}_{-4.0775} \\
& +\underbrace{\lambda_{max} log(P(V_{r}="\textbf{\AR{الجيدة}}"/{\color{blue}\Gamma_{r}=2},{\color{red}T_{r}="\textbf{\AR{نعت}}"})}_{-37.5415} \\
& + \underbrace{log(P({\color{green}\Gamma_{r}=0}/{\color{blue}\Gamma_{r-1}=2},{\color{red}T_{r-1}="\textbf{\AR{نعت}}"}))}_{-0.8527}\\
& = -180.262
\end{split}
\end{equation}

\begin{table}[h]
\begin{center}
\begin{tabular}{ c c c c c c c c c }
  & . & \AR{زيت} & \AR{زيتون} & \AR{من} & \AR{النوعية} & \AR{الجيدة}& . \\   
 case1 & - & - & - & - & \makecell{${\color{red}\cancel{T_{r-1}="\textbf{\AR{نعت}}"}}$ \\ \cancel{${\color{black}\Gamma_{r-1}=3}$}} & ${\color{blue}\Gamma_{r}=2}$ & - \\
    case2 & - & - & - & - & - & \makecell{${\color{red}T_{r}="\textbf{\AR{نعت}}"}$ \\ ${\color{blue}\Gamma_{r}=2}$} & - \\
    case3 & - & - & - & - & - & \makecell{${\color{red}T_{r-1}="\textbf{\AR{نعت}}"}$ \\ ${\color{blue}\Gamma_{r-1}=2}$} & ${\color{green}\Gamma_{r}=0}$
\end{tabular}
\begin{tikzpicture}[overlay]
 \draw[red, line width=1.5pt] (-3.8,0.7) ellipse (3cm and 0.5cm);
 \node (A) at (-1, 0.5) {};
\node (B) at (0, 0.5) {};
\node[] at (1.2,0.5) {\makecell{parasite  \\ probability}};
\draw[->,red, to path={-| (\tikztotarget)}]
  (A) edge (B) ; 
\end{tikzpicture} 
\caption{situations table for $\delta_{5}({\color{red}7},{\color{blue}3}) + log(a_{{\color{red}7}{\color{blue}3}{\color{green}1}})$ for a second order HMM estimation after adding the sentence \AR{زيت زيتون من النوعية الجيدة} to the dataset, the probability $P({\color{blue}\Gamma_{r}=2}/\Gamma_{r-1}=3,{\color{red}T_{r-1}="\textbf{\AR{نعت}}"})$ is a parasite probability because the POS tag of the word \AR{النوعية} is \AR{منعوت} and not \AR{نعت} and the ingredient state of the word \AR{النوعية} is $\Gamma_{r}=0$ and not $\Gamma_{r}=3$}
\label{fig:tab15}
\end{center}
\end{table}

\newpage

\textbf{\textcolor{red}{after adding the sentence \AR{كعبة طماطم من الحجم الكبير} to the dataset:}}

sentence=.,\AR{زيت}(oil),\AR{زيتون}(olive),\AR{من},\AR{النوعية},\AR{الجيدة},.

ingredient state=[0,1,2,0,0,0,0]

POS tags=[.,C,D,G,E,F,.] coded as [4,2,3,9,6,7,4]

ingredient to be extracted=\AR{زيت زيتون}

\begin{gather}\label{mat6}
	\psi_{6}(i,j)=\begin{bmatrix}
	0 & 0 & 1 & 1 \\
	3 & 3 & 1 & 3 \\
	2 & 0 & 3 & 3 \\	
	0 &0 & 1 & 1 \\
          0 & 0 & 0 & 0 \\
         0 & 3 & 1 & 3 \\
          0 & 0 & 0 & 0 \\
	3 & 3 & 3 & 3 \\
	3 & 3 & 3 & 3 \\	
	3 &3 & 3 & 3 \\
           3 & 3& 3 & 3 \\
	1 & 3 & 3 & 3 \\
	3 & 3 & 3 & 3 \\	
	3 &3 & 3 & 3 
	\end{bmatrix}
	\end{gather}

\begin{equation*} 
\begin{split}
\psi_{6}(7,1) & = argmax_{1 \leq i \leq 4}(\delta_{5}(7,i) + log(a_{7i1}))
\end{split}
\end{equation*}

\begin{equation} \label{eq16}
\begin{split}
\delta_{5}({\color{red}7},{\color{blue}1}) + log(a_{{\color{red}7}{\color{blue}1}{\color{green}1}}) & =max_{1 \leq i \leq 4}(\delta_{4}({\color{red}7},i)+log(a_{{\color{red}7}i{\color{blue}1}}))+\lambda_{max} log(b_{{\color{red}7}{\color{blue}1}}("\textbf{\AR{الجيدة}}"))+ log(a_{{\color{red}7}{\color{blue}1}1}) \\
&  = \underbrace{\delta_{4}({\color{red}7},1)}_{-134.9407}+\underbrace{log(P({\color{blue}\Gamma_{r}=0}/\Gamma_{r-1}=0,{\color{red}T_{r-1}="\textbf{\AR{نعت}}"}))}_{-2.7157} \\
& +\underbrace{\lambda_{max} log(P(V_{r}="\textbf{\AR{الجيدة}}"/{\color{blue}\Gamma_{r}=0},{\color{red}T_{r}="\textbf{\AR{نعت}}"})}_{-36.6363} \\
& + \underbrace{log(P({\color{green}\Gamma_{r}=0}/{\color{blue}\Gamma_{r-1}=0},{\color{red}T_{r-1}="\textbf{\AR{نعت}}"}))}_{-2.7157}\\
& = -177.008
\end{split}
\end{equation}

\begin{table}[h]
\begin{center}
\begin{tabular}{ c c c c c c c c c }
  & . & \AR{زيت} & \AR{زيتون} & \AR{من} & \AR{النوعية} & \AR{الجيدة}& . \\   
 case1 & - & - & - & - & \makecell{${\color{red}\cancel{T_{r-1}="\textbf{\AR{نعت}}"}}$ \\ ${\color{black}\Gamma_{r-1}=0}$} & ${\color{blue}\Gamma_{r}=0}$ & - \\
    case2 & - & - & - & - & - & \makecell{${\color{red}T_{r}="\textbf{\AR{نعت}}"}$ \\ ${\color{blue}\Gamma_{r}=0}$} & - \\
    case3 & - & - & - & - & - & \makecell{${\color{red}T_{r-1}="\textbf{\AR{نعت}}"}$ \\ ${\color{blue}\Gamma_{r-1}=0}$} & ${\color{green}\Gamma_{r}=0}$
\end{tabular}
\begin{tikzpicture}[overlay]
 \draw[red, line width=1.5pt] (-3.8,0.7) ellipse (3cm and 0.5cm);
 
 \node (A) at (-1.3, 0.5) {};
\node (B) at (1.8, -0.2) {};

\draw[->,red, to path={-| (\tikztotarget)}]
  (A) edge (B) ;
  \node[] at (1.5,-0.5) {\makecell{parasite probability}};
\end{tikzpicture} 
\caption{situations table for $\delta_{5}({\color{red}7},{\color{blue}1}) + log(a_{{\color{red}7}{\color{blue}1}{\color{green}1}})$ for a second order HMM estimation after adding the sentence \AR{كعبة طماطم من الحجم الكبير} to the dataset, the probability $P({\color{blue}\Gamma_{r}=0}/\Gamma_{r-1}=0,{\color{red}T_{r-1}="\textbf{\AR{نعت}}"})$ is a parasite probability because the POS tag of the word  \AR{النوعية} is \AR{منعوت} and not \AR{نعت}}
\label{fig:tab16}
\end{center}
\end{table}

\begin{equation} \label{eq17}
\begin{split}
\delta_{5}({\color{red}7},{\color{blue}2}) + log(a_{{\color{red}7}{\color{blue}2}{\color{green}1}}) & =max_{1 \leq i \leq 4}(\delta_{4}({\color{red}7},i)+log(a_{{\color{red}7}i{\color{blue}2}}))+\lambda_{max} log(b_{{\color{red}7}{\color{blue}2}}("\textbf{\AR{الجيدة}}"))+ log(a_{{\color{red}7}{\color{blue}2}{\color{green}1}}) \\
&  = \underbrace{\delta_{4}({\color{red}7},1)}_{-134.9407}+\underbrace{log(P({\color{blue}\Gamma_{r}=1}/\Gamma_{r-1}=0,{\color{red}T_{r-1}="\textbf{\AR{نعت}}"}))}_{-3.488} \\
& +\underbrace{\lambda_{max} log(P(V_{r}="\textbf{\AR{الجيدة}}"/{\color{blue}\Gamma_{r}=1},{\color{red}T_{r}="\textbf{\AR{نعت}}"})}_{-38.02485} \\
& + \underbrace{log(P({\color{green}\Gamma_{r}=0}/{\color{blue}\Gamma_{r-1}=1},{\color{red}T_{r-1}="\textbf{\AR{نعت}}"}))}_{-6.5843}\\
& = -183.0379
\end{split}
\end{equation}

\begin{table}[h]
\begin{center}
\begin{tabular}{ c c c c c c c c c }
  & . & \AR{زيت} & \AR{زيتون} & \AR{من} & \AR{النوعية} & \AR{الجيدة}& . \\   
 case1 & - & - & - & - & \makecell{${\color{red}\cancel{T_{r-1}="\textbf{\AR{نعت}}"}}$ \\ ${\color{black}\Gamma_{r-1}=0}$} & ${\color{blue}\Gamma_{r}=1}$ & - \\
    case2 & - & - & - & - & - & \makecell{${\color{red}T_{r}="\textbf{\AR{نعت}}"}$ \\ ${\color{blue}\Gamma_{r}=1}$} & - \\
    case3 & - & - & - & - & - & \makecell{${\color{red}T_{r-1}="\textbf{\AR{نعت}}"}$ \\ ${\color{blue}\Gamma_{r-1}=1}$} & ${\color{green}\Gamma_{r}=0}$
\end{tabular}
\begin{tikzpicture}[overlay]
 \draw[red, line width=1.5pt] (-3.8,0.7) ellipse (3cm and 0.5cm);
 \node (A) at (-1, 0.5) {};
\node (B) at (0, 0.5) {};
\node[] at (1.2,0.5) {\makecell{parasite  \\ probability}};
\draw[->,red, to path={-| (\tikztotarget)}]
  (A) edge (B) ; 
\end{tikzpicture} 
\caption{situations table for $\delta_{5}({\color{red}7},{\color{blue}2}) + log(a_{{\color{red}7}{\color{blue}2}{\color{green}1}})$ for a second order HMM estimation after adding the sentence \AR{كعبة طماطم من الحجم الكبير} to the dataset, the probability $P({\color{blue}\Gamma_{r}=1}/\Gamma_{r-1}=0,{\color{red}T_{r-1}="\textbf{\AR{نعت}}"})$ is a parasite probability because the POS tag of the word \AR{النوعية} is \AR{منعوت} and not \AR{نعت}}
\label{fig:tab17}
\end{center}
\end{table}

\begin{equation} \label{eq18}
\begin{split}
\delta_{5}({\color{red}7},{\color{blue}3}) + log(a_{{\color{red}7}{\color{blue}3}{\color{green}1}}) & =max_{1 \leq i \leq 4}(\delta_{4}({\color{red}7},i)+log(a_{{\color{red}7}i{\color{blue}3}}))+\lambda_{max} log(b_{{\color{red}7}{\color{blue}3}}("\textbf{\AR{الجيدة}}"))+ log(a_{{\color{red}7}{\color{blue}3}{\color{green}1}}) \\
&  = \underbrace{\delta_{4}({\color{red}7},3)}_{-137.8215}+\underbrace{log(P({\color{blue}\Gamma_{r}=2}/\Gamma_{r-1}=3,{\color{red}T_{r-1}="\textbf{\AR{نعت}}"}))}_{-4.0775} \\
& +\underbrace{\lambda_{max} log(P(V_{r}="\textbf{\AR{الجيدة}}"/{\color{blue}\Gamma_{r}=2},{\color{red}T_{r}="\textbf{\AR{نعت}}"})}_{-37.5509} \\
& + \underbrace{log(P({\color{green}\Gamma_{r}=0}/{\color{blue}\Gamma_{r-1}=2},{\color{red}T_{r-1}="\textbf{\AR{نعت}}"}))}_{-0.85277}\\
& = -180.30279
\end{split}
\end{equation}

\begin{table}[h]
\begin{center}
\begin{tabular}{ c c c c c c c c c }
  & . & \AR{زيت} & \AR{زيتون} & \AR{من} & \AR{النوعية} & \AR{الجيدة}& . \\   
 case1 & - & - & - & - & \makecell{${\color{red}\cancel{T_{r-1}="\textbf{\AR{نعت}}"}}$ \\ \cancel{${\color{black}\Gamma_{r-1}=3}$}} & ${\color{blue}\Gamma_{r}=2}$ & - \\
    case2 & - & - & - & - & - & \makecell{${\color{red}T_{r}="\textbf{\AR{نعت}}"}$ \\ ${\color{blue}\Gamma_{r}=2}$} & - \\
    case3 & - & - & - & - & - & \makecell{${\color{red}T_{r-1}="\textbf{\AR{نعت}}"}$ \\ ${\color{blue}\Gamma_{r-1}=2}$} & ${\color{green}\Gamma_{r}=0}$
\end{tabular}
\begin{tikzpicture}[overlay]
 \draw[red, line width=1.5pt] (-3.8,0.7) ellipse (3cm and 0.5cm);
 \node (A) at (-1, 0.5) {};
\node (B) at (0, 0.5) {};
\node[] at (1.2,0.5) {\makecell{parasite  \\ probability}};
\draw[->,red, to path={-| (\tikztotarget)}]
  (A) edge (B) ; 
\end{tikzpicture} 
\caption{situations table for $\delta_{5}({\color{red}7},{\color{blue}3}) + log(a_{{\color{red}7}{\color{blue}3}{\color{green}1}})$ for a second order HMM estimation after adding the sentence \AR{كعبة طماطم من الحجم الكبير} to the dataset, the probability $P({\color{blue}\Gamma_{r}=1}/\Gamma_{r-1}=0,{\color{red}T_{r-1}="\textbf{\AR{نعت}}"})$ is a parasite probability because the POS tag of the word \AR{النوعية} is \AR{منعوت} and not \AR{نعت}}
\label{fig:tab18}
\end{center}
\end{table}

The modifications made in order to ameliorate the performances of a simple double agent Viterbi algorithm are: 
\begin{itemize}
\item Inside the mathematical expression $max_{j}[\delta_{l-1}(i,j)+\log(a_{ijk})]$, $a_{ijk}$ is replaced by $a_{y(l)jk}$ where $y(l)$ is the POS tag observed in the corresponding sequence of POS tags at position l.
\item Inside the mathematical expression $max_{j}[\delta_{l-1}(i,j)+\log(a_{ijk})]$, $\delta_{l-1}(i,j)$ is replaced by $\delta_{l-1}(y(l),j)$ where $y(l)$ is the POS tag observed in the corresponding sequence of POS tags at position l.
\end{itemize}

We observe a tremendous decrease of the parasite probabilities infiltrated inside the equations of the most probable cases of a state matrix elements. For example, the parasite probability $P({\color{blue}\Gamma_{r}=0}/\Gamma_{r-1}=0,{\color{red}T_{r-1}="\textbf{\AR{نعت}}"})$ in in {\color{blue}Table \ref{fig:tab11}} is eliminated and replaced by the sane probability $P({\color{blue}\Gamma_{r}=0}/\Gamma_{r-1}=0,{\color{orange}T_{r-1}="\textbf{\AR{منعوت}}"})$ in in {\color{blue}Table \ref{fig:tab20}} by external knowledge transfer consisting of introducing the POS tag estimated in first layer at position $r-1$.

\newpage
{\color{blue}Estimating an ingredient state using a double agent Viterbi algorithm with external knowledge transfer:}

sentence=.,\AR{زيت}(oil),\AR{زيتون}(olive),\AR{من},\AR{النوعية},\AR{الجيدة},.

ingredient state=[0,1,2,0,0,0,0]

POS tags=[.,C,D,G,E,F,.] coded as [4,2,3,9,6,7,4]

ingredient to be extracted=\AR{زيت زيتون}

\begin{equation} \label{eq19}
\begin{split}
\delta_{5}({\color{red}7},{\color{blue}1}) + log(a_{{\color{red}7}{\color{blue}1}{\color{green}1}}) & =max_{1 \leq i \leq 4}(\delta_{4}({\color{orange}POS[5]},i)+log(a_{{\color{orange}POS[5]}i{\color{blue}1}}))+\lambda_{max} log(b_{{\color{red}7}{\color{blue}1}}("\textbf{\AR{الجيدة}}"))+ log(a_{{\color{red}7}{\color{blue}1}{\color{green}1}}) \\
&  = \underbrace{\delta_{4}({\color{orange}6},2)}_{-94.593}+\underbrace{log(P({\color{blue}\Gamma_{r}=0}/\Gamma_{r-1}=1,{\color{orange}T_{r-1}="\textbf{\AR{منعوت}}"}))}_{-2.189} \\
& +\underbrace{\lambda_{max} log(P(V_{r}="\textbf{\AR{الجيدة}}"/{\color{blue}\Gamma_{r}=0},{\color{red}T_{r}="\textbf{\AR{نعت}}"})}_{-36.63} \\
& + \underbrace{log(P({\color{green}\Gamma_{r}=0}/{\color{blue}\Gamma_{r-1}=0},{\color{red}T_{r-1}="\textbf{\AR{نعت}}"}))}_{-2.7178}\\
& = -136.13
\end{split}
\end{equation}

\begin{table}[h]

\begin{center}
\begin{tabular}{ c c c c c c c c c }
  & . & \AR{زيت} & \AR{زيتون} & \AR{من} & \AR{النوعية} & \AR{الجيدة}& . \\   
 case1 & - & - & - & - & \makecell{${\color{orange}T_{r-1}="\textbf{\AR{منعوت}}"}$ \\ \cancel{${\color{black}\Gamma_{r-1}=1}$}}  & ${\color{blue}\Gamma_{r}=0}$ & - \\
    case2 & - & - & - & - & - & \makecell{${\color{red}T_{r}="\textbf{\AR{نعت}}"}$ \\ ${\color{blue}\Gamma_{r}=0}$} & - \\
    case3 & - & - & - & - & - & \makecell{${\color{red}T_{r-1}="\textbf{\AR{نعت}}"}$ \\ ${\color{blue}\Gamma_{r-1}=0}$} & ${\color{green}\Gamma_{r}=0}$
\end{tabular}
\begin{tikzpicture}[overlay]
 \draw[red, line width=1.5pt] (-4.5,0.7) ellipse (3cm and 0.5cm);
 \node (A) at (-1.7, 0.5) {};
\node (B) at (1.8, -0.2) {};
\node (C) at (0.8, -1.2) {};
\node (D) at (1.8, -0.8) {};
\draw[->,red, to path={-| (\tikztotarget)}]
  (A) edge (B) ;
  \node[] at (1.5,-0.5) {\makecell{parasite \\ probability}};
\end{tikzpicture} 
\caption{situations table for $\delta_{5}({\color{red}7},{\color{blue}1}) + log(a_{{\color{red}7}{\color{blue}1}{\color{green}1}})$ for a double agent Viterbi with external knowledge transfer, $P({\color{blue}\Gamma_{r}=0}/\Gamma_{r-1}=1,{\color{orange}T_{r-1}="\textbf{\AR{منعوت}}"})$ is a parasite probability because the ingredient state of the word \AR{النوعية} is $\Gamma_{r-1}=0$ and not $\Gamma_{r-1}=1$}
\label{fig:tab19}
\end{center}
\end{table}

\begin{equation} \label{eq20}
\begin{split}
\delta_{5}({\color{red}7},{\color{blue}2}) + log(a_{{\color{red}7}{\color{blue}2}{\color{green}1}}) & =max_{1 \leq i \leq 4}(\delta_{4}({\color{orange}POS[5]},i)+log(a_{{\color{orange}POS[5]}i{\color{blue}2}}))+\lambda_{max} log(b_{{\color{red}7}{\color{blue}2}}("\textbf{\AR{الجيدة}}"))+ log(a_{{\color{red}7}{\color{blue}2}{\color{green}1}}) \\
&  = \underbrace{\delta_{4}({\color{orange}6},1)}_{-93.928}+\underbrace{log(P({\color{blue}\Gamma_{r}=1}/\Gamma_{r-1}=0,{\color{orange}T_{r-1}="\textbf{\AR{منعوت}}"}))}_{-7.559} \\
& +\underbrace{\lambda_{max} log(P(V_{r}="\textbf{\AR{الجيدة}}"/{\color{blue}\Gamma_{r}=1},{\color{red}T_{r}="\textbf{\AR{نعت}}"})}_{-38.015} \\
& + \underbrace{log(P({\color{green}\Gamma_{r}=0}/{\color{blue}\Gamma_{r-1}=1},{\color{red}T_{r-1}="\textbf{\AR{نعت}}"}))}_{-6.583}\\
& = -146.087
\end{split}
\end{equation}

\begin{table}[h]
\begin{center}
\begin{tabular}{ c c c c c c c c c }
  & . & \AR{زيت} & \AR{زيتون} & \AR{من} & \AR{النوعية} & \AR{الجيدة}& . \\   
 case1 & - & - & - & - & \makecell{${\color{orange}T_{r-1}="\textbf{\AR{منعوت}}"}$ \\ ${\color{black}\Gamma_{r-1}=0}$} & ${\color{blue}\Gamma_{r}=1}$ & - \\
    case2 & - & - & - & - & - & \makecell{${\color{red}T_{r}="\textbf{\AR{نعت}}"}$ \\ ${\color{blue}\Gamma_{r}=1}$} & - \\
    case3 & - & - & - & - & - & \makecell{${\color{red}T_{r-1}="\textbf{\AR{نعت}}"}$ \\ ${\color{blue}\Gamma_{r-1}=1}$} & ${\color{green}\Gamma_{r}=0}$
\end{tabular}
\begin{tikzpicture}[overlay]
 \draw[green, line width=1.5pt] (-4,0.7) ellipse (3.2cm and 0.5cm);
 \node (A) at (-1.1, 0.5) {};
\node (B) at (0, 0.5) {};
\node[] at (1.2,0.5) {\makecell{parasite  \\ probability \\ eliminated}};
\draw[->,green, to path={-| (\tikztotarget)}]
  (A) edge (B) ; 
\end{tikzpicture} 
\caption{situations table for $\delta_{5}({\color{red}7},{\color{blue}2}) + log(a_{{\color{red}7}{\color{blue}2}{\color{green}1}})$ for a double agent Viterbi with external knowledge transfer, $P({\color{blue}\Gamma_{r}=1}/\Gamma_{r-1}=0,{\color{orange}T_{r-1}="\textbf{\AR{منعوت}}"})$ is a sain probability because the word \AR{النوعية} is a \AR{منعوت} and it is not an ingredient.}
\label{fig:tab20}
\end{center}
\end{table}

\begin{equation} \label{eq21}
\begin{split}
\delta_{5}({\color{red}7},{\color{blue}3}) + log(a_{{\color{red}7}{\color{blue}3}{\color{green}1}}) & =max_{1 \leq i \leq 4}(\delta_{4}({\color{orange}POS[5]},i)+log(a_{{\color{orange}POS[5]}i{\color{blue}3}}))+\lambda_{max} log(b_{{\color{red}7}{\color{blue}3}}("\textbf{\AR{الجيدة}}"))+ log(a_{{\color{red}7}{\color{blue}3}{\color{green}1}}) \\
&  = \underbrace{\delta_{4}({\color{orange}6},2)}_{-94.59}+\underbrace{log(P({\color{blue}\Gamma_{r}=2}/\Gamma_{r-1}=1,{\color{orange}T_{r-1}="\textbf{\AR{منعوت}}"}))}_{-2.029} \\
& +\underbrace{\lambda_{max} log(P(V_{r}="\textbf{\AR{الجيدة}}"/{\color{blue}\Gamma_{r}=2},{\color{red}T_{r}="\textbf{\AR{نعت}}"})}_{-37.541} \\
& + \underbrace{log(P({\color{green}\Gamma_{r}=0}/{\color{blue}\Gamma_{r-1}=2},{\color{red}T_{r-1}="\textbf{\AR{نعت}}"}))}_{-0.851}\\
& = -135.015
\end{split}
\end{equation}

\begin{table}[h]
\begin{center}
\begin{tabular}{ c c c c c c c c c }
  & . & \AR{زيت} & \AR{زيتون} & \AR{من} & \AR{النوعية} & \AR{الجيدة}& . \\   
 case1 & - & - & - & - & \makecell{${\color{orange}T_{r-1}="\textbf{\AR{منعوت}}"}$ \\ \cancel{${\color{black}\Gamma_{r-1}=1}$}} & ${\color{blue}\Gamma_{r}=2}$ & - \\
    case2 & - & - & - & - & - & \makecell{${\color{red}T_{r}="\textbf{\AR{نعت}}"}$ \\ ${\color{blue}\Gamma_{r}=2}$} & - \\
    case3 & - & - & - & - & - & \makecell{${\color{red}T_{r-1}="\textbf{\AR{نعت}}"}$ \\ ${\color{blue}\Gamma_{r-1}=2}$} & ${\color{green}\Gamma_{r}=0}$
\end{tabular}
\begin{tikzpicture}[overlay]
 \draw[red, line width=1.5pt] (-4,0.7) ellipse (3.2cm and 0.5cm);
 \node (A) at (-1.1, 0.5) {};
\node (B) at (0, 0.5) {};
\node[] at (1.2,0.5) {\makecell{parasite  \\ probability \\ maintained}};
\draw[->,red, to path={-| (\tikztotarget)}]
  (A) edge (B) ; 
\end{tikzpicture} 
\caption{situations table for $\delta_{5}({\color{red}7},{\color{blue}3}) + log(a_{{\color{red}7}{\color{blue}3}{\color{green}1}}) $ for a double agent Viterbi with external knowledge transfer, $P({\color{blue}\Gamma_{r}=2}/\Gamma_{r-1}=1,{\color{orange}T_{r-1}="\textbf{\AR{منعوت}}"})$ is a parasite probability because the word \AR{النوعية} is not an ingredient.}
\label{fig:tab21}
\end{center}
\end{table}

\textbf{\textcolor{red}{after adding the sentence \AR{كعبة طماطم من الحجم الكبير} to the dataset:}}

sentence=.,\AR{زيت}(oil),\AR{زيتون}(olive),\AR{من},\AR{النوعية},\AR{الجيدة},.

ingredient state=[0,1,2,0,0,0,0]

POS tags=[.,C,D,G,E,F,.] coded as [4,2,3,9,6,7,4]

ingredient to be extracted=\AR{زيت زيتون}

\begin{equation} \label{eq22}
\begin{split}
\delta_{5}({\color{red}7},{\color{blue}1}) + log(a_{{\color{red}7}{\color{blue}1}{\color{green}1}}) & =max_{1 \leq i \leq 4}(\delta_{4}({\color{orange}POS[5]},i)+log(a_{{\color{orange}POS[5]}i{\color{blue}1}}))+\lambda_{max} log(b_{{\color{red}7}{\color{blue}1}}("\textbf{\AR{الجيدة}}"))+ log(a_{{\color{red}7}{\color{blue}1}{\color{green}1}}) \\
&  = \underbrace{\delta_{4}({\color{orange}6},2)}_{-94.59}+\underbrace{log(P({\color{blue}\Gamma_{r}=0}/\Gamma_{r-1}=1,{\color{orange}T_{r-1}="\textbf{\AR{منعوت}}"}))}_{-2.189} \\
& +\underbrace{\lambda_{max} log(P(V_{r}="\textbf{\AR{الجيدة}}"/{\color{blue}\Gamma_{r}=0},{\color{red}T_{r}="\textbf{\AR{نعت}}"})}_{-36.637} \\
& + \underbrace{log(P({\color{green}\Gamma_{r}=0}/{\color{blue}\Gamma_{r-1}=0},{\color{red}T_{r-1}="\textbf{\AR{نعت}}"}))}_{-2.715}\\
& = -136.133
\end{split}
\end{equation}

\begin{table}[h]

\begin{center}
\begin{tabular}{ c c c c c c c c c }
  & . & \AR{زيت} & \AR{زيتون} & \AR{من} & \AR{النوعية} & \AR{الجيدة}& . \\   
 case1 & - & - & - & - & \makecell{${\color{orange}T_{r-1}="\textbf{\AR{منعوت}}"}$ \\ \cancel{${\color{black}\Gamma_{r-1}=1}$}}  & ${\color{blue}\Gamma_{r}=0}$ & - \\
    case2 & - & - & - & - & - & \makecell{${\color{red}T_{r}="\textbf{\AR{نعت}}"}$ \\ ${\color{blue}\Gamma_{r}=0}$} & - \\
    case3 & - & - & - & - & - & \makecell{${\color{red}T_{r-1}="\textbf{\AR{نعت}}"}$ \\ ${\color{blue}\Gamma_{r-1}=0}$} & ${\color{green}\Gamma_{r}=0}$
\end{tabular}
\begin{tikzpicture}[overlay]
 \draw[red, line width=1.5pt] (-4.5,0.7) ellipse (3cm and 0.5cm);
 \node (A) at (-1.7, 0.5) {};
\node (B) at (1.8, -0.2) {};
\node (C) at (0.8, -1.2) {};
\node (D) at (1.8, -0.8) {};
\draw[->,red, to path={-| (\tikztotarget)}]
  (A) edge (B) ;
  \node[] at (1.5,-0.5) {\makecell{parasite \\ probability}};
\end{tikzpicture} 
\caption{situations table for $\delta_{5}({\color{red}7},{\color{blue}1}) + log(a_{{\color{red}7}{\color{blue}1}{\color{green}1}})$ for a double agent Viterbi with external knowledge transfer, after adding the sentence \AR{كعبة طماطم من الحجم الكبير} to the dataset, $P({\color{blue}\Gamma_{r}=0}/\Gamma_{r-1}=1,{\color{orange}T_{r-1}="\textbf{\AR{منعوت}}"})$ is a parasite probability because the word \AR{النوعية} is not an ingredient in reality given \AR{الجيدة} have an ingredient state $\Gamma_{r}=0$.}
\label{fig:tab22}
\end{center}
\end{table}

\begin{equation} \label{eq23}
\begin{split}
\delta_{5}({\color{red}7},{\color{blue}2}) + log(a_{{\color{red}7}{\color{blue}2}{\color{green}1}}) & =max_{1 \leq i \leq 4}(\delta_{4}({\color{orange}POS[5]},i)+log(a_{{\color{orange}POS[5]}i{\color{blue}2}}))+\lambda_{max} log(b_{{\color{red}7}{\color{blue}2}}("\textbf{\AR{الجيدة}}"))+ log(a_{{\color{red}7}{\color{blue}2}{\color{green}1}}) \\
&  = \underbrace{\delta_{4}({\color{orange}6},1)}_{-93.91}+\underbrace{log(P({\color{blue}\Gamma_{r}=1}/\Gamma_{r-1}=0,{\color{orange}T_{r-1}="\textbf{\AR{منعوت}}"}))}_{-7.56} \\
& +\underbrace{\lambda_{max} log(P(V_{r}="\textbf{\AR{الجيدة}}"/{\color{blue}\Gamma_{r}=1},{\color{red}T_{r}="\textbf{\AR{نعت}}"})}_{-38.024} \\
& + \underbrace{log(P({\color{green}\Gamma_{r}=0}/{\color{blue}\Gamma_{r-1}=1},{\color{red}T_{r-1}="\textbf{\AR{نعت}}"}))}_{-6.584}\\
& = -146.078
\end{split}
\end{equation}

\begin{table}[h]
\begin{center}
\begin{tabular}{ c c c c c c c c c }
  & . & \AR{زيت} & \AR{زيتون} & \AR{من} & \AR{النوعية} & \AR{الجيدة}& . \\   
 case1 & - & - & - & - & \makecell{${\color{orange}T_{r-1}="\textbf{\AR{منعوت}}"}$ \\ ${\color{black}\Gamma_{r-1}=0}$} & ${\color{blue}\Gamma_{r}=1}$ & - \\
    case2 & - & - & - & - & - & \makecell{${\color{red}T_{r}="\textbf{\AR{نعت}}"}$ \\ ${\color{blue}\Gamma_{r}=1}$} & - \\
    case3 & - & - & - & - & - & \makecell{${\color{red}T_{r-1}="\textbf{\AR{نعت}}"}$ \\ ${\color{blue}\Gamma_{r-1}=1}$} & ${\color{green}\Gamma_{r}=0}$
\end{tabular}
\begin{tikzpicture}[overlay]
 \draw[green, line width=1.5pt] (-4,0.8) ellipse (3cm and 0.5cm);
 \node (A) at (-1.1, 0.8) {};
\node (B) at (0, 0.8) {};
\node[] at (1.2,0.8) {\makecell{saine  \\ probability}};
\draw[->,green, to path={-| (\tikztotarget)}]
  (A) edge (B) ; 
\end{tikzpicture} 
\caption{situations table for $\delta_{5}({\color{red}7},{\color{blue}2}) + log(a_{{\color{red}7}{\color{blue}2}{\color{green}1}})$ for a double agent Viterbi with external knowledge transfer, after adding the sentence \AR{كعبة طماطم من الحجم الكبير} to the dataset, $P({\color{blue}\Gamma_{r}=0}/\Gamma_{r-1}=0,{\color{orange}T_{r-1}="\textbf{\AR{منعوت}}"})$ is not a parasite probability because the word \AR{النوعية} is not an ingredient and \AR{النوعية} POS tag is \AR{منعوت} .}
\label{fig:tab23}
\end{center}
\end{table}

\begin{equation} \label{eq24}
\begin{split}
\delta_{5}({\color{red}7},{\color{blue}3}) + log(a_{{\color{red}7}{\color{blue}3}{\color{green}1}}) & =max_{1 \leq i \leq 4}(\delta_{4}({\color{orange}POS[5]},i)+log(a_{{\color{orange}POS[5]}i{\color{blue}3}}))+\lambda_{max} log(b_{{\color{red}7}{\color{blue}3}}("\textbf{\AR{الجيدة}}"))+ log(a_{{\color{red}7}{\color{blue}3}{\color{green}1}}) \\
&  = \underbrace{\delta_{4}({\color{orange}6},2)}_{-94.59}+\underbrace{log(P({\color{blue}\Gamma_{r}=2}/\Gamma_{r-1}=1,{\color{orange}T_{r-1}="\textbf{\AR{منعوت}}"}))}_{-2.03} \\
& +\underbrace{\lambda_{max} log(P(V_{r}="\textbf{\AR{الجيدة}}"/{\color{blue}\Gamma_{r}=2},{\color{red}T_{r}="\textbf{\AR{نعت}}"})}_{-37.55} \\
& + \underbrace{log(P({\color{green}\Gamma_{r}=0}/{\color{blue}\Gamma_{r-1}=2},{\color{red}T_{r-1}="\textbf{\AR{نعت}}"}))}_{-0.852}\\
& = -135.02
\end{split}
\end{equation}

\begin{table}[h]
\begin{center}
\begin{tabular}{ c c c c c c c c c }
  & . & \AR{زيت} & \AR{زيتون} & \AR{من} & \AR{النوعية} & \AR{الجيدة}& . \\   
 case1 & - & - & - & - & \makecell{${\color{orange}T_{r-1}="\textbf{\AR{منعوت}}"}$ \\ \cancel{${\color{black}\Gamma_{r-1}=1}$}} & ${\color{blue}\Gamma_{r}=2}$ & - \\
    case2 & - & - & - & - & - & \makecell{${\color{red}T_{r}="\textbf{\AR{نعت}}"}$ \\ ${\color{blue}\Gamma_{r}=2}$} & - \\
    case3 & - & - & - & - & - & \makecell{${\color{red}T_{r-1}="\textbf{\AR{نعت}}"}$ \\ ${\color{blue}\Gamma_{r-1}=2}$} & ${\color{green}\Gamma_{r}=0}$
\end{tabular}
\begin{tikzpicture}[overlay]
 \draw[red, line width=1.5pt] (-4,0.7) ellipse (3.2cm and 0.5cm);
 \node (A) at (-1.1, 0.5) {};
\node (B) at (0, 0.5) {};
\node[] at (1.2,0.5) {\makecell{parasite  \\ probability }};
\draw[->,red, to path={-| (\tikztotarget)}]
  (A) edge (B) ; 
\end{tikzpicture} 
\caption{situations table for $\delta_{5}({\color{red}7},{\color{blue}3}) + log(a_{{\color{red}7}{\color{blue}3}{\color{green}1}})$ for a double agent Viterbi with external knowledge transfer, after adding the sentence \AR{كعبة طماطم من الحجم الكبير} to the dataset, $P({\color{blue}\Gamma_{r}=2}/\Gamma_{r-1}=1,{\color{orange}T_{r-1}="\textbf{\AR{منعوت}}"})$ is a parasite probability because the word \AR{النوعية}  have an ingredient state $\Gamma_{r-1}=0$ and not $\Gamma_{r-1}=1$.}
\label{fig:tab24}
\end{center}
\end{table}

\textbf{\textcolor{red}{after adding the sentence \AR{زيت زيتون من النوعية الجيدة} to the dataset:}}

sentence=.,\AR{زيت}(oil),\AR{زيتون}(olive),\AR{من},\AR{النوعية},\AR{الجيدة},.

ingredient state=[0,1,2,0,0,0,0]

POS tags=[.,C,D,G,E,F,.] coded as [4,2,3,9,6,7,4]

ingredient to be extracted=\AR{زيت زيتون}

\begin{equation} \label{eq25}
\begin{split}
\delta_{5}({\color{red}7},{\color{blue}1}) + log(a_{{\color{red}7}{\color{blue}1}{\color{green}1}}) & =max_{1 \leq i \leq 4}(\delta_{4}({\color{orange}POS[5]},i)+log(a_{{\color{orange}POS[5]}i{\color{blue}1}}))+\lambda_{max} log(b_{{\color{red}7}{\color{blue}1}}("\textbf{\AR{الجيدة}}"))+ log(a_{{\color{red}7}{\color{blue}1}{\color{green}1}}) \\
&  = \underbrace{\delta_{4}({\color{orange}6},1)}_{-92.101}+\underbrace{log(P({\color{blue}\Gamma_{r}=0}/\Gamma_{r-1}=0,{\color{orange}T_{r-1}="\textbf{\AR{منعوت}}"}))}_{-3.2605} \\
& +\underbrace{\lambda_{max} log(P(V_{r}="\textbf{\AR{الجيدة}}"/{\color{blue}\Gamma_{r}=0},{\color{red}T_{r}="\textbf{\AR{نعت}}"})}_{-35.009} \\
& + \underbrace{log(P({\color{green}\Gamma_{r}=0}/{\color{blue}\Gamma_{r-1}=0},{\color{red}T_{r-1}="\textbf{\AR{نعت}}"}))}_{-2.7159}\\
& = -133.08
\end{split}
\end{equation}

\begin{table}[h]

\begin{center}
\begin{tabular}{ c c c c c c c c c }
  & . & \AR{زيت} & \AR{زيتون} & \AR{من} & \AR{النوعية} & \AR{الجيدة}& . \\   
 case1 & - & - & - & - & \makecell{${\color{orange}T_{r-1}="\textbf{\AR{منعوت}}"}$ \\ ${\color{black}\Gamma_{r-1}=0}$}  & ${\color{blue}\Gamma_{r}=0}$ & - \\
    case2 & - & - & - & - & - & \makecell{${\color{red}T_{r}="\textbf{\AR{نعت}}"}$ \\ ${\color{blue}\Gamma_{r}=0}$} & - \\
    case3 & - & - & - & - & - & \makecell{${\color{red}T_{r-1}="\textbf{\AR{نعت}}"}$ \\ ${\color{blue}\Gamma_{r-1}=0}$} & ${\color{green}\Gamma_{r}=0}$
\end{tabular}
\begin{tikzpicture}[overlay]
 \draw[green, line width=1.5pt] (-4.5,0.7) ellipse (3cm and 0.5cm);
 \node (A) at (-1.7, 0.5) {};
\node (B) at (1.8, -0.2) {};
\node (C) at (0.8, -1.2) {};
\node (D) at (1.8, -0.8) {};
\draw[->,green, to path={-| (\tikztotarget)}]
(A) edge (B) ;
  \node[] at (1.5,-0.5) {\makecell{parasite \\ probability \\ eliminated}};
\end{tikzpicture} 
\caption{situations table for $\delta_{5}({\color{red}7},{\color{blue}1}) + log(a_{{\color{red}7}{\color{blue}1}{\color{green}1}})$ for a double agent Viterbi with external knowledge transfer after adding the sentence \AR{زيت زيتون من النوعية الجيدة} to the dataset, $P({\color{blue}\Gamma_{r}=0}/\Gamma_{r-1}=0,{\color{orange}T_{r-1}="\textbf{\AR{منعوت}}"})$ is not a parasite probability because the word \AR{النوعية} is not an ingredient and the word \AR{النوعية} is \AR{منعوت} given the word \AR{الجيدة} is not an ingredient.}
\label{fig:tab25}
\end{center}

\end{table}

\begin{equation} \label{eq26}
\begin{split}
\delta_{5}({\color{red}7},{\color{blue}2}) + log(a_{{\color{red}7}{\color{blue}2}{\color{green}1}}) & =max_{1 \leq i \leq 4}(\delta_{4}({\color{orange}POS[5]},i)+log(a_{{\color{orange}POS[5]}i{\color{blue}2}}))+\lambda_{max} log(b_{{\color{red}7}{\color{blue}2}}("\textbf{\AR{الجيدة}}"))+ log(a_{{\color{red}7}{\color{blue}2}{\color{green}1}}) \\
&  = \underbrace{\delta_{4}({\color{orange}6},1)}_{-92.101}+\underbrace{log(P({\color{blue}\Gamma_{r}=1}/\Gamma_{r-1}=0,{\color{orange}T_{r-1}="\textbf{\AR{منعوت}}"}))}_{-7.56} \\
& +\underbrace{\lambda_{max} log(P(V_{r}="\textbf{\AR{الجيدة}}"/{\color{blue}\Gamma_{r}=1},{\color{red}T_{r}="\textbf{\AR{نعت}}"})}_{-38.016} \\
& + \underbrace{log(P({\color{green}\Gamma_{r}=0}/{\color{blue}\Gamma_{r-1}=1},{\color{red}T_{r-1}="\textbf{\AR{نعت}}"}))}_{-6.584}\\
& = -144.26
\end{split}
\end{equation}

\begin{table}[h]
\begin{center}
\begin{tabular}{ c c c c c c c c c }
  & . & \AR{زيت} & \AR{زيتون} & \AR{من} & \AR{النوعية} & \AR{الجيدة}& . \\   
 case1 & - & - & - & - & \makecell{${\color{orange}T_{r-1}="\textbf{\AR{منعوت}}"}$ \\ ${\color{black}\Gamma_{r-1}=0}$} & ${\color{blue}\Gamma_{r}=1}$ & - \\
    case2 & - & - & - & - & - & \makecell{${\color{red}T_{r}="\textbf{\AR{نعت}}"}$ \\ ${\color{blue}\Gamma_{r}=1}$} & - \\
    case3 & - & - & - & - & - & \makecell{${\color{red}T_{r-1}="\textbf{\AR{نعت}}"}$ \\ ${\color{blue}\Gamma_{r-1}=1}$} & ${\color{green}\Gamma_{r}=0}$
\end{tabular}
\begin{tikzpicture}[overlay]
 \draw[green, line width=1.5pt] (-4,0.8) ellipse (3cm and 0.5cm);
 \node (A) at (-1.1, 0.8) {};
\node (B) at (0, 0.8) {};
\node[] at (1.2,0.5) {\makecell{parasite  \\ probability \\ eliminated}};
\draw[->,green, to path={-| (\tikztotarget)}]
  (A) edge (B) ; 
\end{tikzpicture} 
\caption{situations table for $\delta_{5}({\color{red}7},{\color{blue}2}) + log(a_{{\color{red}7}{\color{blue}2}{\color{green}1}})$ for a double agent Viterbi with external knowledge transfer after adding the sentence \AR{زيت زيتون من النوعية الجيدة} to the dataset, $P({\color{blue}\Gamma_{r}=1}/\Gamma_{r-1}=0,{\color{orange}T_{r-1}="\textbf{\AR{منعوت}}"})$ is not a parasite probability because the word \AR{النوعية} is not an ingredient and the word \AR{النوعية} is \AR{منعوت} given the word \AR{الجيدة} is not an ingredient.}
\label{fig:tab26}
\end{center}
\end{table}

\begin{equation} \label{eq27}
\begin{split}
\delta_{5}({\color{red}7},{\color{blue}3}) + log(a_{{\color{red}7}{\color{blue}3}{\color{green}1}}) & =max_{1 \leq i \leq 4}(\delta_{4}({\color{orange}POS[5]},i)+log(a_{{\color{orange}POS[5]}i{\color{blue}3}}))+\lambda_{max} log(b_{{\color{red}7}{\color{blue}3}}("\textbf{\AR{الجيدة}}"))+ log(a_{{\color{red}7}{\color{blue}3}{\color{green}1}}) \\
&  = \underbrace{\delta_{4}({\color{orange}6},2)}_{-94.401}+\underbrace{log(P({\color{blue}\Gamma_{r}=2}/\Gamma_{r-1}=1,{\color{orange}T_{r-1}="\textbf{\AR{منعوت}}"}))}_{-2.03} \\
& +\underbrace{\lambda_{max} log(P(V_{r}="\textbf{\AR{الجيدة}}"/{\color{blue}\Gamma_{r}=2},{\color{red}T_{r}="\textbf{\AR{نعت}}"})}_{-37.541} \\
& + \underbrace{log(P({\color{green}\Gamma_{r}=0}/{\color{blue}\Gamma_{r-1}=2},{\color{red}T_{r-1}="\textbf{\AR{نعت}}"}))}_{-0.852}\\
& = -134.826
\end{split}
\end{equation}

\begin{table}[h]
\begin{center}
\begin{tabular}{ c c c c c c c c c }
  & . & \AR{زيت} & \AR{زيتون} & \AR{من} & \AR{النوعية} & \AR{الجيدة}& . \\   
 case1 & - & - & - & - & \makecell{${\color{orange}T_{r-1}="\textbf{\AR{منعوت}}"}$ \\ \cancel{${\color{black}\Gamma_{r-1}=1}$}} & ${\color{blue}\Gamma_{r}=2}$ & - \\
    case2 & - & - & - & - & - & \makecell{${\color{red}T_{r}="\textbf{\AR{نعت}}"}$ \\ ${\color{blue}\Gamma_{r}=2}$} & - \\
    case3 & - & - & - & - & - & \makecell{${\color{red}T_{r-1}="\textbf{\AR{نعت}}"}$ \\ ${\color{blue}\Gamma_{r-1}=2}$} & ${\color{green}\Gamma_{r}=0}$
\end{tabular}
\begin{tikzpicture}[overlay]
 \draw[red, line width=1.5pt] (-4,0.7) ellipse (3.2cm and 0.5cm);
 \node (A) at (-1.1, 0.5) {};
\node (B) at (0, 0.5) {};
\node[] at (1.2,0.5) {\makecell{parasite  \\ probability }};
\draw[->,red, to path={-| (\tikztotarget)}]
  (A) edge (B) ; 
\end{tikzpicture} 
\caption{situations table for $\delta_{5}({\color{red}7},{\color{blue}3}) + log(a_{{\color{red}7}{\color{blue}3}{\color{green}1}})$ for a double agent Viterbi with external knowledge transfer after adding the sentence \AR{زيت زيتون من النوعية الجيدة} to the dataset, $P({\color{blue}\Gamma_{r}=2}/\Gamma_{r-1}=1,{\color{orange}T_{r-1}="\textbf{\AR{منعوت}}"})$ is a parasite probability because the word \AR{النوعية} is not an ingredient in reality.}
\label{fig:tab27}
\end{center}
\end{table}

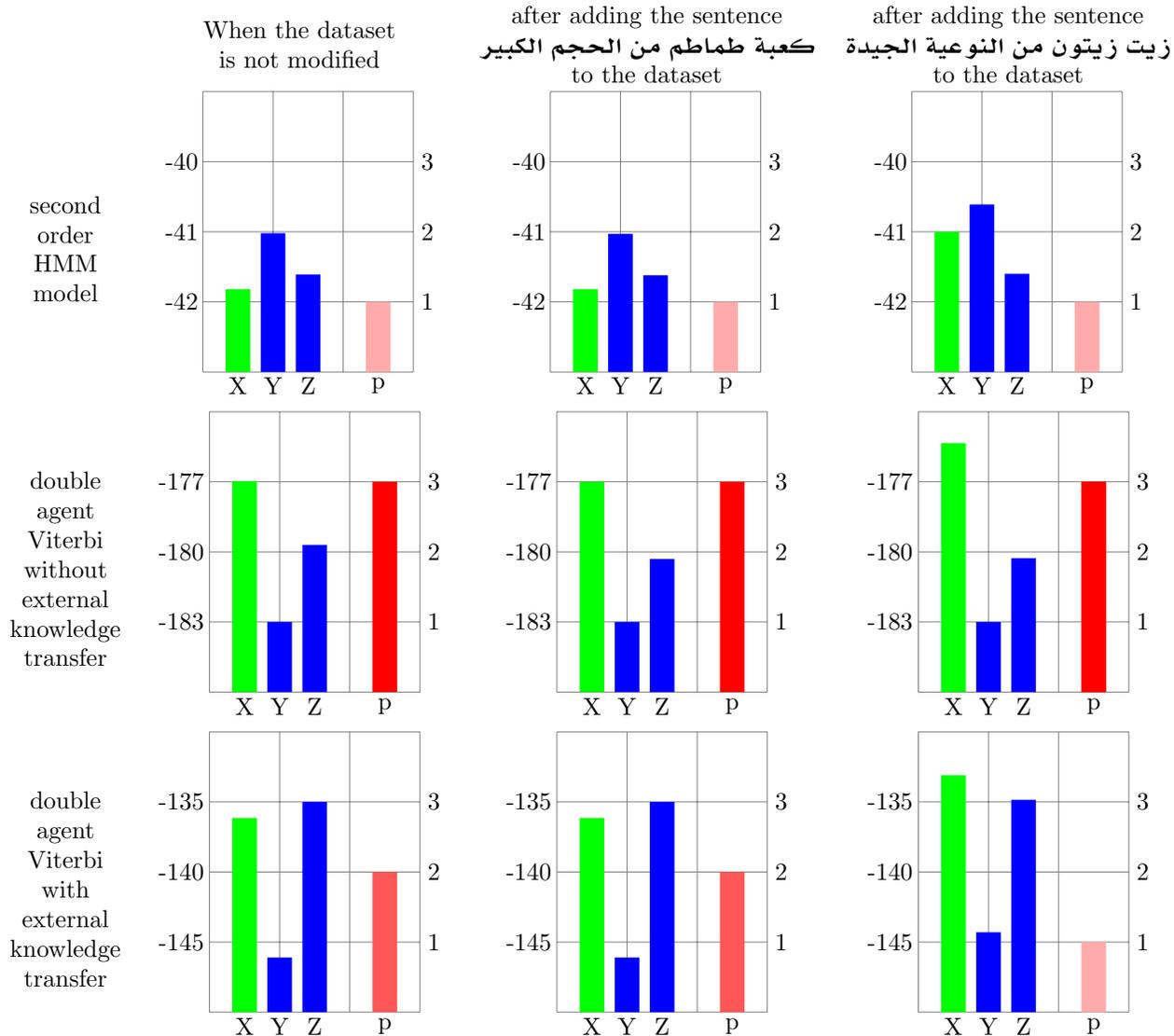
\begin{figure}
\begin{tabular}{llll}
  &  \makecell{When the dataset \\ is not modified} &\makecell{after adding the sentence \\ \AR{كعبة طماطم من الحجم الكبير} \\ to the dataset} &  \makecell{after adding the sentence \\ \AR{زيت زيتون من النوعية الجيدة} \\ to the dataset} \\
\makecell{second \\ order \\ HMM \\ model} & \makecell{
 \begin{tikzpicture}[ycomb]
\draw[] (0.5,-0.2) node {X};
\draw[] (1,-0.2) node {Y};
\draw[] (1.5,-0.2) node {Z};
\draw[] (2.5,-0.2) node {p};
\draw[] (-0.3,1) node {-42};
\draw[] (-0.3,2) node {-41};
\draw[] (-0.3,3) node {-40};
\draw[] (3.2,1) node {1};
\draw[] (3.2,2) node {2};
\draw[] (3.2,3) node {3};
\draw [help lines] grid (3,4);
\draw[color=green,line width=10pt]
plot coordinates{(0.5,1.18) };
\draw[color=blue,line width=10pt]
plot coordinates{ (1,1.98) (1.5,1.39)  };
\draw[color=red!33,line width=10pt]
plot coordinates{(2.5,1)};
\end{tikzpicture}} &
\makecell{
 \begin{tikzpicture}[ycomb]
\draw[] (0.5,-0.2) node {X};
\draw[] (1,-0.2) node {Y};
\draw[] (1.5,-0.2) node {Z};
\draw[] (2.5,-0.2) node {p};
\draw[] (-0.3,1) node {-42};
\draw[] (-0.3,2) node {-41};
\draw[] (-0.3,3) node {-40};
\draw[] (3.2,1) node {1};
\draw[] (3.2,2) node {2};
\draw[] (3.2,3) node {3};
\draw [help lines] grid (3,4);
\draw[color=green,line width=10pt]
plot coordinates{(0.5,1.18) };
\draw[color=blue,line width=10pt]
plot coordinates{ (1,1.97) (1.5,1.38)  };
\draw[color=red!33,line width=10pt]
plot coordinates{(2.5,1)};
\end{tikzpicture}} & \makecell{\begin{tikzpicture}[ycomb]
\draw[] (0.5,-0.2) node {X};
\draw[] (1,-0.2) node {Y};
\draw[] (1.5,-0.2) node {Z};
\draw[] (2.5,-0.2) node {p};
\draw[] (-0.3,1) node {-42};
\draw[] (-0.3,2) node {-41};
\draw[] (-0.3,3) node {-40};
\draw[] (3.2,1) node {1};
\draw[] (3.2,2) node {2};
\draw[] (3.2,3) node {3};
\draw [help lines] grid (3,4);
\draw[color=green,line width=10pt]
plot coordinates{(0.5,2) };
\draw[color=blue,line width=10pt]
plot coordinates{ (1,2.39) (1.5,1.4)  };
\draw[color=red!33,line width=10pt]
plot coordinates{(2.5,1)};
\end{tikzpicture}} \\
\makecell{double \\ agent \\ Viterbi \\ without \\ external \\ knowledge \\ transfer} & \makecell{
\begin{tikzpicture}[ycomb]
\draw[] (0.5,-0.2) node {X};
\draw[] (1,-0.2) node {Y};
\draw[] (1.5,-0.2) node {Z};
\draw[] (2.5,-0.2) node {p};
\draw[] (-0.4,1) node {-183};
\draw[] (-0.4,2) node {-180};
\draw[] (-0.4,3) node {-177};
\draw[] (3.2,1) node {1};
\draw[] (3.2,2) node {2};
\draw[] (3.2,3) node {3};
\draw [help lines] grid (3,4);
\draw[color=green,line width=10pt]
plot coordinates{(0.5,3.01) };
\draw[color=blue,line width=10pt]
plot coordinates{ (1,1) (1.5,2.1)  };
\draw[color=red!100,line width=10pt]
plot coordinates{(2.5,3)};
\end{tikzpicture}} & \makecell{\begin{tikzpicture}[ycomb]
\draw[] (0.5,-0.2) node {X};
\draw[] (1,-0.2) node {Y};
\draw[] (1.5,-0.2) node {Z};
\draw[] (2.5,-0.2) node {p};
\draw[] (-0.4,1) node {-183};
\draw[] (-0.4,2) node {-180};
\draw[] (-0.4,3) node {-177};
\draw[] (3.2,1) node {1};
\draw[] (3.2,2) node {2};
\draw[] (3.2,3) node {3};
\draw [help lines] grid (3,4);
\draw[color=green,line width=10pt]
plot coordinates{(0.5,3) };
\draw[color=blue,line width=10pt]
plot coordinates{ (1,1) (1.5,1.9)  };
\draw[color=red!100,line width=10pt]
plot coordinates{(2.5,3)};
\end{tikzpicture}}  & \makecell{\begin{tikzpicture}[ycomb]
\draw[] (0.5,-0.2) node {X};
\draw[] (1,-0.2) node {Y};
\draw[] (1.5,-0.2) node {Z};
\draw[] (2.5,-0.2) node {p};
\draw[] (-0.4,1) node {-183};
\draw[] (-0.4,2) node {-180};
\draw[] (-0.4,3) node {-177};
\draw[] (3.2,1) node {1};
\draw[] (3.2,2) node {2};
\draw[] (3.2,3) node {3};
\draw [help lines] grid (3,4);
\draw[color=green,line width=10pt]
plot coordinates{(0.5,3.55) };
\draw[color=blue,line width=10pt]
plot coordinates{ (1,1) (1.5,1.91)  };
\draw[color=red!100,line width=10pt]
plot coordinates{(2.5,3)};
\draw[color=red!100,line width=10pt]
plot coordinates{(2.5,3)};
\end{tikzpicture}} \\
\makecell{double \\ agent \\ Viterbi \\ with \\ external \\ knowledge \\ transfer} &
\makecell{
\begin{tikzpicture}[ycomb]
\draw[] (0.5,-0.2) node {X};
\draw[] (1,-0.2) node {Y};
\draw[] (1.5,-0.2) node {Z};
\draw[] (2.5,-0.2) node {p};
\draw[] (-0.4,1) node {-145};
\draw[] (-0.4,2) node {-140};
\draw[] (-0.4,3) node {-135};
\draw[] (3.2,1) node {1};
\draw[] (3.2,2) node {2};
\draw[] (3.2,3) node {3};
\draw [help lines] grid (3,4);
\draw[color=green,line width=10pt]
plot coordinates{(0.5,2.77) };
\draw[color=blue,line width=10pt]
plot coordinates{ (1,0.78) (1.5,3)  };
\draw[color=red!66,line width=10pt]
plot coordinates{(2.5,2)};
\end{tikzpicture}} & \makecell{\begin{tikzpicture}[ycomb]
\draw[] (0.5,-0.2) node {X};
\draw[] (1,-0.2) node {Y};
\draw[] (1.5,-0.2) node {Z};
\draw[] (2.5,-0.2) node {p};
\draw[] (-0.4,1) node {-145};
\draw[] (-0.4,2) node {-140};
\draw[] (-0.4,3) node {-135};
\draw[] (3.2,1) node {1};
\draw[] (3.2,2) node {2};
\draw[] (3.2,3) node {3};
\draw [help lines] grid (3,4);
\draw[color=green,line width=10pt]
plot coordinates{(0.5,2.77) };
\draw[color=blue,line width=10pt]
plot coordinates{ (1,0.78) (1.5,3)  };
\draw[color=red!66,line width=10pt]
plot coordinates{(2.5,2)};
\end{tikzpicture}} & \makecell{\begin{tikzpicture}[ycomb]
\draw[] (0.5,-0.2) node {X};
\draw[] (1,-0.2) node {Y};
\draw[] (1.5,-0.2) node {Z};
\draw[] (2.5,-0.2) node {p};
\draw[] (-0.4,1) node {-145};
\draw[] (-0.4,2) node {-140};
\draw[] (-0.4,3) node {-135};
\draw[] (3.2,1) node {1};
\draw[] (3.2,2) node {2};
\draw[] (3.2,3) node {3};
\draw [help lines] grid (3,4);
\draw[color=green,line width=10pt]
plot coordinates{(0.5,3.38) };
\draw[color=blue,line width=10pt]
plot coordinates{ (1,1.14) (1.5,3.03)  };
\draw[color=red!33,line width=10pt]
plot coordinates{(2.5,1)};
\end{tikzpicture}}
\end{tabular}
\caption{data sensitivity bar plot for the word \AR{الجيدة} inside the sentence \AR{زيت زيتون من النوعية الجيدة} under graduated conditions. X is the first most probable case inside a state matrix \AR{الجيدة} is not an ingredient), Y is the second most probable case inside a state matrix \AR{الجيدة} is an ingredient constituted with a single word) and Z is the third most probable case inside a state matrix (\AR{الجيدة} is the second word of an ingredient constituted with two elements) , P is the number of parasite probabilities observed when calculating X and Y and Z.}
\label{fig:fig1}
\end{figure}

\begin{table}[p]
\begin{center}
    \begin{tabular}{| l | l | l | l | l | l | l |}
    \hline
    \makecell{tag} & accuracy & f1 score & \makecell{accuracy for \\ unknown \\ words} & \makecell{accuracy for \\ known \\ words} & \makecell{number of \\ unknown \\ words} & \makecell{number of \\ known \\ words} \\ \hline
    \makecell{A \\ (\AR{إسم})}& 98.47 \% & 65.57 \% & NULL & 98.47 \% & 0 & 1112  \\ \hline
    \makecell{B \\ (\AR{رقم})} & 99.79 \% & 74.95 \% & NULL & 99.79 \% & 0 & 947  \\ \hline
    \makecell{C \\ (\AR{إسم معرف})} & 96.99 \% & 83.79 \% & NULL & 96.99 \% & 0 & 798  \\
    \hline
    \makecell{D \\ (\AR{إضافة})} & 94.69 \% & 70.83 \% & NULL & 94.69 \% & 0 & 791  \\ \hline
    \makecell{E \\ (\AR{منعوت})} & 98.8 \% & 87.95 \% & NULL & 98.8 \% & 0 & 750  \\ \hline
    \makecell{F \\ (\AR{نعت})} & 85.31 \% & 69.92 \% & NULL & 85.31 \% & 0 & 844  \\
    \hline 
    \makecell{G \\ (\AR{حرف جر})}& 99.54 \% & 49.89 \% & NULL & 99.54 \% & 0 & 219  \\ \hline
    \makecell{H \\ (\AR{إسم مجرور})} & 86.81 \% & 60.21 \% & NULL & 86.81 \% & 0 & 182  \\ \hline
    \makecell{I \\ (\AR{وحدة قيس})}& 99.36 \% & 49.89 \% & NULL & 99.36 \% & 0 & 469  \\ \hline
    \makecell{J \\ (\AR{واو العطف})} & 98.77 \% & 49.69 \% & NULL & 98.77  \% & 0 & 162  \\ \hline
    \makecell{K \\ \AR{فعل مبني })\\\AR{للمجهول)}} &  78.57 \% & 31.45 \% &  NULL & 78.57 \% & 0 & 140  \\ \hline
    \makecell{L \\ (\AR{المفعول المطلق})} & 60 \% & 25 \% & NULL & 60 \%  & 0 & 5  \\ \hline
      \makecell{M \\ \AR{أداةُ عَطْفٍ غير })\\\AR{واو العطف)}} & 92.31 \% & 48 \% & NULL  & 92.31 \% & 0 & 39 \\ \hline
    \makecell{.} & 99.9 \% & 49.97 \% & NULL & 99.9 \% & 0 & 1972 \\
    \hline 
    
    \end{tabular}
    \label{tab:tab28}
\end{center}
\caption{Performances for each tag of a first order HMM , the trained dataset is used as a test dataset , and there is zero unknown words to predict }
\end{table}

\begin{table}[p]
\begin{center}
    \begin{tabular}{| l | l | l | l | l | l | l |}
    \hline
    \makecell{tag} & accuracy & f1 score & \makecell{accuracy for \\ unknown \\ words} & \makecell{accuracy for \\ known \\ words} & \makecell{number of \\ unknown \\ words} & \makecell{number of \\ known \\ words} \\ \hline
    \makecell{A \\ (\AR{إسم})}& 97.3 \% & 64.68 \% & NULL & 97.3 \% & 0 & 1112  \\ \hline
    \makecell{B \\ (\AR{رقم})} & 99.89 \% & 83.31 \% & NULL & 99.89 \% & 0 & 947  \\ \hline
    \makecell{C \\ (\AR{إسم معرف})} & 94.86 \% & 85.36 \% & NULL & 94.86 \% & 0 & 798  \\
    \hline
    \makecell{D \\ (\AR{إضافة})} & 91.02 \% & 68.01 \% & NULL & 91.02 \% & 0 & 791  \\ \hline
    \makecell{E \\ (\AR{منعوت})} & 98.53 \% & 98.84 \% & NULL & 98.53 \% & 0 & 750  \\ \hline
    \makecell{F \\ (\AR{نعت})} & 77.61 \% & 54.6 \% & NULL & 77.61 \% & 0 & 844  \\
    \hline 
    \makecell{G \\ (\AR{حرف جر})}& 99.09 \% & 49.77 \% & NULL & 99.09 \% & 0 & 219  \\ \hline
    \makecell{H \\ (\AR{إسم مجرور})} & 76.37 \% & 52.49 \% & NULL & 76.37 \% & 0 & 182  \\ \hline
    \makecell{I \\ (\AR{وحدة قيس})}& 99.57 \% & 74.89 \% & NULL & 99.57 \% & 0 & 469  \\ \hline
    \makecell{J \\ (\AR{واو العطف})} & 98.77 \% & 49.69 \% & NULL & 98.77  \% & 0 & 162  \\ \hline
    \makecell{K \\ \AR{فعل مبني })\\\AR{للمجهول)}} &  78.57 \% & 32.2 \% &  NULL & 78.57 \% & 0 & 140  \\ \hline
    \makecell{L \\ (\AR{المفعول المطلق})} & 20 \% & 16.67 \% & NULL & 20 \%  & 0 & 5  \\ \hline
      \makecell{M \\ \AR{أداةُ عَطْفٍ غير })\\\AR{واو العطف)}} & 94.87 \% & 48.68 \% & NULL  & 94.87 \% & 0 & 39 \\ \hline
    \makecell{.} & 99.85 \% & 49.96 \% & NULL & 99.85 \% & 0 & 3945 \\
    \hline 
    
    \end{tabular}
     \label{tab:tab29}
\end{center}
\caption{Performances for each tag of a second order HMM , the trained dataset is used as a test dataset , and there is zero unknown words to predict }
\end{table}

\begin{table}
\begin{center}
    \begin{tabular}{| l | l | l | l | l | l | l |}
    \hline
    \makecell{tag} & accuracy & f1 score & \makecell{accuracy for \\ unknown \\ words} & \makecell{accuracy for \\ known \\ words} & \makecell{number of \\ unknown \\ words} & \makecell{number of \\ known \\ words} \\ \hline
    \makecell{A \\ (\AR{إسم})}& 97.84 \% & 97.06 \% & NULL & 97.84 \% & 0 & 1113  \\ \hline
    \makecell{B \\ (\AR{رقم})} & 99.79 \% & 49.95 \% & NULL & 99.79 \% & 0 & 947  \\ \hline
    \makecell{C \\ (\AR{إسم معرف})} & 96.25 \% & 97.44 \% & NULL & 96.25 \% & 0 & 799  \\
    \hline
    \makecell{D \\ (\AR{إضافة})} & 93.94 \% & 94.92 \% & NULL & 93.94 \% & 0 & 792  \\ \hline
    \makecell{E \\ (\AR{منعوت})} & 98.8 \% & 65.77 \% & NULL & 98.8 \% & 0 & 748  \\ \hline
    \makecell{F \\ (\AR{نعت})} & 88.02 \% & 58.34 \% & NULL & 88.02 \% & 0 & 843  \\
    \hline 
    \makecell{G \\ (\AR{حرف جر})}& 99.54 \% & 49.89 \% & NULL & 99.54 \% & 0 & 219  \\ \hline
    \makecell{H \\ (\AR{إسم مجرور})} & 89.56 \% & 89.36 \% & NULL & 89.56 \% & 0 & 182  \\ \hline
    \makecell{I \\ (\AR{وحدة قيس})}& 99.79 \% & 83.28 \% & NULL & 99.79 \% & 0 & 469  \\ \hline
    \makecell{J \\ (\AR{واو العطف})} & 98.77 \% & 49.69 \% & NULL & 98.77 \% & 0 & 162  \\ \hline
    \makecell{K \\ \AR{فعل مبني })\\\AR{للمجهول)}} & 99.29 \%& 49.82 \% & NULL & 99.29 \% & 0 & 140  \\ \hline
    \makecell{L \\ (\AR{المفعول المطلق})} & 100 \% & 100 \% & NULL & 100 \%  & 0 & 5 \\ \hline
      \makecell{M \\ \AR{أداةُ عَطْفٍ غير })\\\AR{واو العطف)}} & 97.44 \% & 49.35 \% & NULL  & 97.44 \% & 0 & 39 \\ \hline
    \makecell{.} & 100 \% & 100 \% & NULL & 100 \% & 0 & 3946 \\
    \hline 
    
    \end{tabular}
    \label{tab:tab30}
\end{center}
\caption{Performance comparison using a simple double agent Viterbi with external knowledge transfer with 100 \% accuracy on first layer , the trained dataset is used as a test dataset to avoid unknown words prediction }
\end{table}

\begin{figure}[p]
\begin{subfigure}{.9\textwidth}
  \centering
  \includegraphics[width=1.2\linewidth]{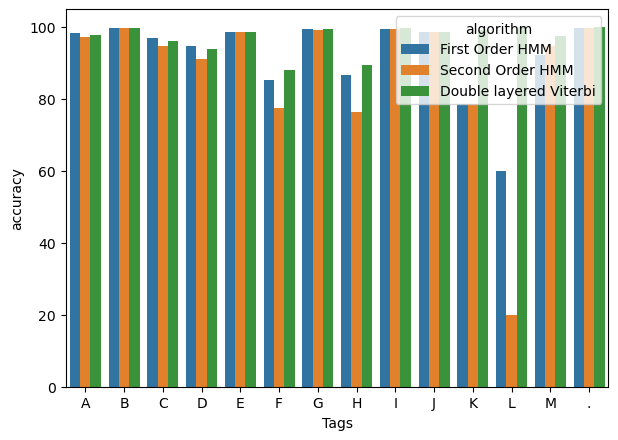}  
  \caption{accuracy for each tag without unknown words consideration}
\end{subfigure}

\begin{subfigure}{.9\textwidth}
  \centering
  \includegraphics[width=1.2\linewidth]{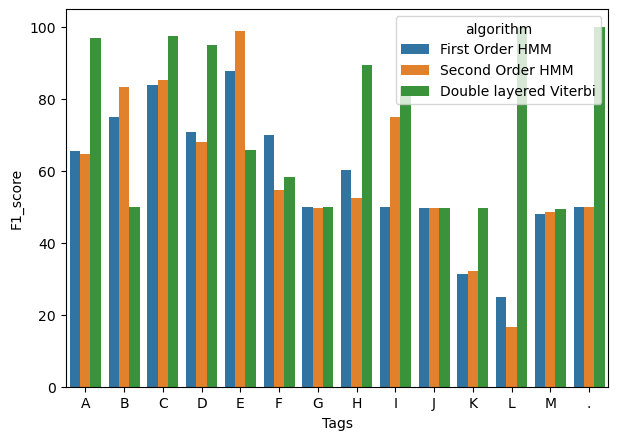}  
  \caption{f1 score for each tag without unknown words consideration}
\end{subfigure}
\caption{Performance comparison between a double agent Viterbi algorithm with external knowledge transfer and other mono layered Viterbi algorithm}
\label{fig:fig2}
\end{figure}

\begin{table}[p]
\begin{center}
		\begin{tabular}{|l|c|r|}
			\cline{2-3}
			\multicolumn{1}{c|}{} & Accuracy  & F1 score   \\
			\hline
				\makecell{First order Hidden \\ Markov Model applied to tokens \\ to predict ingredients} & 96.61 \% & 70.06 \% \\ \hline
				\hline
				\makecell{Second order Hidden \\ Markov Model applied to tokens \\ to predict ingredients} & 95.82 \% & 68.18 \% \\ \hline
			
				\hline
				\makecell{Our ingredient extractor \\ with 100 \% accuracy on the first layer \\ with $\lambda=4$} & 98.43 \% & 81.43 \%  \\ \hline 
				\hline
				\makecell{A double layered Viterbi algorithm \\ with 100 \% accuracy on the first layer \\ with $\lambda=4$} & 97.7 \% & 96.32 \%  \\ \hline 
		\end{tabular}
		\centering
\caption{ Comparison of performances between a simple double agent Viterbi algorithm with external knowledge transfer and other HMM mono-agent methods using the trained corpus as a testing dataset}
\label{tab:tab31}
\end{center}
\end{table}

\begin{table}[p]
\begin{center}
    \begin{tabular}{| l | l | l | l | l | l | l |}
    \hline
    \makecell{tag} & accuracy & f1 score & \makecell{accuracy for \\ unknown \\ words} & \makecell{accuracy for \\ known \\ words} & \makecell{number of \\ unknown \\ words} & \makecell{unknown \\ words \\ percentage} \\ \hline
   0& 94.77 \% & 67.67 \% & 67.78 \% & 96.33 \% & 78 & 9.67 \%  \\ \hline
    1 & 92.96 \% & 65.52 \% & 63.93 \% & 95.14 \% & 107 & 13.26 \%  \\ \hline
    2 &  94.03 \% & 67.36 \% & 69.9 \% & 95.68 \% & 88& 10.9 \%  \\
    \hline
    3  & 93.95 \% & 87.84 \% & 62.65 \% & 95.64 \% & 75 & 9.29 \%  \\ \hline
   4  & 92.91 \% & 87.29 \% & 66.99 \% & 94.54 \% & 92 & 11.4 \% \\ \hline
   5&  95.78 \% & 69.61 \% & 76.4 \% & 96.83 \% & 83 & 10.29 \% \\
    \hline 
    6&  94.38 \% & 67.53 \% & 54.65 \% & 96.49 \% & 80 & 9.91 \%  \\ \hline
    7& 95.08 \% & 91.17 \% & 61.22 \% & 97.17 \% & 90 & 11.15 \%  \\ \hline
    8&  96.83 \% & 94.04 \% & 79.76 \% & 97.7 \% & 77 & 9.54 \% \\ \hline
    9 & 94.92 \% & 90.7 \% & 64.95 \% & 96.72 \% & 88 & 10.9 \%  \\ \hline \hline
   Avg & 94.56 \% & 78.87 \% & 66.82 \% & 96.22 \% & 86 & 10.63 \% \\ \hline
  
    \end{tabular}
   \label{tab:tab32}
\end{center}
\caption{Performances of first order HMM used to predict ingredients state using 10 fold cross-validation with unknown word consideration, the test dataset have 20 \% sentences from the total dataset}
\end{table}

\begin{table}[p]
\begin{center}
    \begin{tabular}{| l | l | l | l | l | l | l |}
    \hline
    \makecell{tag} & accuracy & f1 score & \makecell{accuracy for \\ unknown \\ words} & \makecell{accuracy for \\ known \\ words} & \makecell{number of \\ unknown \\ words} & \makecell{unknown \\ words \\ percentage} \\ \hline
   0& 93.72 \% & 64.13 \% & 70.0 \% &94.81 \% & 78 & 9.67 \%  \\ \hline
    1 &92.48 \% & 62.96 \% & 64.75 \% &94.14 \% & 107 & 13.26 \%  \\ \hline
    2 &94.04 \% & 65.53 \% & 70.87 \% &95.29 \% & 88 & 10.9 \%  \\
    \hline
    3  &93.89 \% & 86.87 \% & 68.67 \% &94.97 \% & 75 & 9.29 \%   \\ \hline
   4  &93.42 \% & 86.61 \% & 66.02 \% &94.81 \% & 92 & 11.4 \%  \\ \hline
   5&93.83 \% & 65.85 \% & 62.92 \% &95.18 \% & 83 & 10.29 \% \\
    \hline 
    6& 93.76 \% & 65.63 \% & 60.47 \% &95.19 \% & 80 & 9.91 \%  \\ \hline
    7& 94.04 \% & 88.56 \% & 55.1 \% &95.96 \% & 90 & 11.15 \% \\ \hline
    8& 95.34 \% & 90.99 \% & 60.71 \% &96.77 \% & 77 & 9.54 \%  \\ \hline
    9 & 94.34 \% & 88.56 \% & 61.86 \% &95.91 \% & 88 & 10.9 \%  \\ \hline \hline
   Avg & 93.89 \% & 76.57 \% & 64.14 \% &95.3 \% & 86 & 10.63 \% \\ \hline
  
    \end{tabular}
    \label{tab:tab33}
\end{center}
\caption{Performances of second order HMM used to predict ingredients state using 10 fold cross-validation with unknown word consideration, the test dataset have 20 \% sentences from the total dataset }
\end{table}

\begin{table}[p]
\begin{center}
    \begin{tabular}{| l | l | l | l | l | l | l |}
    \hline
    \makecell{tag} & accuracy & f1 score & \makecell{accuracy for \\ unknown \\ words} & \makecell{accuracy for \\ known \\ words} & \makecell{number of \\ unknown \\ words} & \makecell{unknown \\ words \\ percentage} \\ \hline
   0& 95.78 \% & 67.91 \% & 61.11 \% & 97.38 \% & 78 & 9.67 \%  \\ \hline
    1 & 94.34  \% & 66.51 \% & 68.85 \% & 95.87 \% & 107 & 13.26 \%  \\ \hline
    2 & 94.98  \% & 67.32 \% & 67.96 \% & 96.44 \% & 88  & 10.9 \%  \\
    \hline
    3  & 95.83  \% & 67.71 \% & 62.65 \% & 97.25 \% & 75  & 9.29 \%  \\ \hline
   4  & 94.88  \% & 67.76 \% & 71.84 \% & 96.05 \% & 92  & 11.4 \%  \\ \hline
   5& 95.9  \% & 69.57 \% & 70.79 \% & 97.0 \% & 83  & 10.29 \%  \\
    \hline 
    6& 94.43  \% & 65.92 \% & 55.81 \% & 96.08 \% & 80  & 9.91 \%  \\ \hline
    7&94.04  \% & 65.54 \% & 51.02 \% & 96.17 \% & 90  & 11.15 \%  \\ \hline
    8& 95.63  \% & 68.82 \% & 60.71 \% & 97.06 \% & 77  & 9.54 \% \\ \hline
    9 &94.39  \% & 65.95 \% &56.7 \% & 96.21 \% &88  & 10.9\%  \\ \hline \hline
   Avg & 95.02 \% & 67.3 \% & 62.74 \% & 96.55 \%  & 86  & 10.63 \% \\ \hline
  
    \end{tabular}
    \label{tab:tab34}
\end{center}
\caption{Average performances in predicting ingredients states with $\lambda = 4$ when accuracy is 100\% on first layer with our ingredient extractor }
\end{table}

\begin{table}[p]
\begin{center}
    \begin{tabular}{| l | l | l | l | l | l | l |}
    \hline
    \makecell{tag} & accuracy & f1 score & \makecell{accuracy for \\ unknown \\ words} & \makecell{accuracy for \\ known \\ words} & \makecell{number of \\ unknown \\ words} & \makecell{unknown \\ words \\ percentage} \\ \hline
  0& 96.61 \% & 69.39 \% & 80 \% & 97.38 \% & 78 & 9.67 \%  \\ \hline
    1 & 94.89  \% & 67.6 \% & 77.05 \% & 95.96 \% & 107 & 13.26 \%  \\ \hline
    2 & 95.13  \% & 67.76 \% & 85.44 \% & 95.65 \% & 88  & 10.9 \%  \\
    \hline
    3  & 95.87  \% & 91.04 \% & 80.72 \% & 96.53 \% & 75  & 9.29 \%  \\ \hline
   4  & 95.49 \% & 90.68 \% & 78.64 \% & 96.34 \% & 92  & 11.4 \%  \\ \hline
   5& 95.81  \% & 69.1 \% & 74.16 \% & 97.0 \% & 83  & 10.29 \%  \\
    \hline 
    6& 94.43  \% & 65.92 \% & 55.81 \% & 96.08 \% & 80  & 9.91 \%  \\ \hline
    7&94.04  \% & 65.54 \% & 51.02 \% & 96.17 \% & 90  & 11.15 \%  \\ \hline
    8& 95.63  \% & 68.82 \% & 60.71 \% & 97.06 \% & 77  & 9.54 \% \\ \hline
    9 &94.39  \% & 65.95 \% &56.7 \% & 96.21 \% &88  & 10.9\%  \\ \hline \hline
   Avg & 95.24 \% & 72.19 \% & 70.13 \% & 96.42 \% & 86 & 10.63 \% \\ \hline
  
    \end{tabular}
    \label{tab:tab35}
\end{center}
\caption{Average performances in predicting ingredients states with $\lambda = 4$ when accuracy is 100\% on first layer using a double layered Viterbi Algorithm }
\end{table}

	\begin{center}

\begin{table}[p]
		\begin{tabular}{|l|c|c|c|r|}
			\cline{2-5}
			\multicolumn{1}{c|}{} & known  & unknown & overall & F1 score \\
			\hline
				\makecell{First order Hidden \\ Markov Model applied to tokens} & 96.22 \% & 66.82 \% & 94.56 \% & 78.87 \% \\ \hline
				\hline
				\makecell{Second order Hidden \\ Markov Model applied to tokens} & 95.3 \% & 64.14 \% & 93.89 \% & 76.57 \% \\ \hline
			\hline
				\makecell{Our ingredient extractor \\ with 100 \% accuracy on the first layer \\ with $\lambda=4$} & 96.55 \% & 62.74 \% & 95.02 \% & 67.3 \% \\ \hline
				\hline
				\makecell{A double layered Viterbi algorithm \\ with 100 \% accuracy on the first layer \\ with $\lambda=4$} & 96.42 \% & 70.13 \% & 95.24 \% & 72.19 \% \\ \hline
			
		\end{tabular}

		\centering	
\caption{Comparison of performances between the 4 methods using 20 \% of the total dataset as a test dataset}
\label{tab:tab36}
\end{table}
	\end{center}

\newpage

{\color{blue}Estimating an ingredient state using a simple double agent Viterbi without unknown words consideration:}

sentence=.,10,\AR{غ},\AR{خميرة},\AR{جافة},\AR{للخبز},.

ingredient state=[0,0,0,1,2,0,0]

POS tags=[.,B,I,E,F,H,.] coded as [4, 1, 10, 6, 7, 12, 4]

ingredient to be extracted=\AR{خميرة جافة}

\begin{gather}\label{mat7}
	\psi_{6}(i,j)=\begin{bmatrix}
	0 & 0 & 1 & 1 \\
	3 & 3 & 1 & 3 \\
	1 & 0 & 3 & 3 \\	
	0 &0 & 1 & 1 \\
          1 & 0 & 1 & 1 \\
         1 & 3 & 1 & 3\\
          0 & 0 & 3 & 3 \\
	1 & 3 & 1 & 3 \\
	3 & 3 & 3 & 3 \\	
	3 & 3 & 3 & 3\\
          3 & 3 & 3 & 3 \\
	0 & 3 & 3 & 3 \\
	3 & 3 & 3 & 3 \\	
	3 &3 & 3 & 3 
	\end{bmatrix}
	\end{gather}

\begin{equation*} 
\begin{split}
\psi_{6}(12,1) & = argmax_{1 \leq i \leq 4}(\delta_{5}(12,i) + log(a_{12i1}))
\end{split}
\end{equation*}

\begin{equation} \label{eq28}
\begin{split}
\delta_{5}({\color{red}12},{\color{blue}1}) + log(a_{{\color{red}12}{\color{blue}1}{\color{green}1}}) & =max_{1 \leq i \leq 4}(\delta_{4}({\color{red}12},i)+log(a_{{\color{red}12}i{\color{blue}1}}))+\lambda_{max} log(b_{{\color{red}12}{\color{blue}1}}("\textbf{\AR{للخبز}}"))+ log(a_{{\color{red}12}{\color{blue}1}{\color{green}1}}) \\
&  = \underbrace{\delta_{4}({\color{red}12},2)}_{-139.3}+\underbrace{log(P({\color{blue}\Gamma_{r}=0}/\Gamma_{r-1}=1,{\color{red}T_{r-1}="\textbf{\AR{إسم مجرور}}"}))}_{-3.118} \\
& +\lambda_{max}\underbrace{log(P(V_{r}="\textbf{\AR{للخبز}}"/{\color{blue}\Gamma_{r}=0},{\color{red}T_{r}="\textbf{\AR{إسم مجرور}}"})}_{-36.629} \\
& + \underbrace{log(P({\color{green}\Gamma_{r}=0}/{\color{blue}\Gamma_{r-1}=0},{\color{red}T_{r-1}="\textbf{\AR{إسم مجرور}}"}))}_{-4.192}\\
& = -183.24
\end{split}
\end{equation}

\begin{table}[h]
\begin{center}
\begin{tabular}{ c c c c c c c c c }
&  &  &  &  &  &  &  \\
   & . & 10 & \AR{غ} & \AR{خميرة} & \AR{جافة} & \AR{للخبز}& . \\    
 case1 & - & - & - & - & \makecell{${\color{red}\cancel{T_{r-1}="\textbf{\AR{إسم مجرور}}"}}$ \\ ${\color{black}\cancel{\Gamma_{r-1}=1}}$}&  ${\color{blue}\Gamma_{r}=0}$&  - \\
    case2 & - & - & - & - & - & \makecell{${\color{red}T_{r}="\textbf{\AR{إسم مجرور}}"}$ \\ ${\color{blue}\Gamma_{r}=0}$} & - \\
    case3 & - & - & - & - & - & \makecell{${\color{red}T_{r-1}="\textbf{\AR{إسم مجرور}}"}$ \\ ${\color{blue}\Gamma_{r-1}=0}$} & ${\color{green}\Gamma_{r}=0}$
\end{tabular}
\begin{tikzpicture}[overlay]

 \draw[red, line width=1.5pt] (2,2.2) ellipse (3cm and 0.5cm);

 \node (A) at (5, 2.2) {};
\node (B) at (8.5, 1) {};
\node (C) at (0.8, -1.2) {};
\node (D) at (1.8, -0.6) {};
\draw[->,red, to path={-| (\tikztotarget)}]
  (A) edge (B) ;
  \node[] at (8.5,1) {\makecell{parasite probability}};

\end{tikzpicture} 
\caption{situations table for $\delta_{5}({\color{red}12},{\color{blue}1}) + log(a_{{\color{red}12}{\color{blue}1}{\color{green}1}})$ for a simple double agent Viterbi without unknown words consideration, the probability $P({\color{blue}\Gamma_{r}=0}/\Gamma_{r-1}=1,{\color{red}T_{r-1}="\textbf{\AR{إسم مجرور}}"})$ is a parasite probability because the real POS tag for the word \AR{جافة} is  \AR{نعت} and not \AR{إسم مجرور} and the ingredient state of the word \AR{جافة} is $\Gamma_{r-1}=2$ and not $\Gamma_{r-1}=1$}
\label{tab:tab37}
\end{center}
\end{table}

\begin{equation} \label{eq29}
\begin{split}
\delta_{5}({\color{red}12},{\color{blue}2}) + log(a_{{\color{red}12}{\color{blue}2}{\color{green}1}}) & =max_{1 \leq i \leq 4}(\delta_{4}({\color{red}12},i)+log(a_{{\color{red}12}i{\color{blue}2}}))+\lambda_{max} log(b_{{\color{red}12}{\color{blue}2}}("\textbf{\AR{للخبز}}"))+ log(a_{{\color{red}12}{\color{blue}2}{\color{green}1}}) \\
&  = \underbrace{\delta_{4}({\color{red}12},4)}_{-138.61}+\underbrace{log(P({\color{blue}\Gamma_{r}=1}/\Gamma_{r-1}=3,{\color{red}T_{r-1}="\textbf{\AR{إسم مجرور}}"}))}_{-4.077} \\
& +\lambda_{max}\underbrace{ log(P(V_{r}="\textbf{\AR{للخبز}}"/{\color{blue}\Gamma_{r}=1},{\color{red}T_{r}="\textbf{\AR{إسم مجرور}}"})}_{-38.016} \\
& + \underbrace{log(P({\color{green}\Gamma_{r}=0}/{\color{blue}\Gamma_{r-1}=1},{\color{red}T_{r-1}="\textbf{\AR{إسم مجرور}}"}))}_{-3.118}\\
& = -183.82
\end{split}
\end{equation}

\begin{table}[h]
\begin{center}
\begin{tabular}{ c c c c c c c c c }
   & . & 10 & \AR{غ} & \AR{خميرة} & \AR{جافة} & \AR{للخبز}& . \\   
 case1 & - & - & - & - & \makecell{${\color{red}\cancel{T_{r-1}="\textbf{\AR{إسم مجرور}}"}}$ \\ ${\color{black}\cancel{\Gamma_{r-1}=3}}$} & ${\color{blue}\Gamma_{r}=1}$ & - \\
    case2 & - & - & - & - & - & \makecell{${\color{red}T_{r}="\textbf{\AR{إسم مجرور}}"}$ \\ ${\color{blue}\Gamma_{r}=1}$} & - \\
    case3 & - & - & - & - & - & \makecell{${\color{red}T_{r-1}="\textbf{\AR{إسم مجرور}}"}$ \\ ${\color{blue}\Gamma_{r-1}=1}$} & ${\color{green}\Gamma_{r}=0}$
\end{tabular}
\begin{tikzpicture}[overlay]
 \draw[red, line width=1.5pt] (2,2.2) ellipse (3cm and 0.5cm);

 \node (A) at (5, 2.2) {};
\node (B) at (8.5, 1) {};
\node (C) at (0.8, -1.2) {};
\node (D) at (1.8, -0.6) {};
\draw[->,red, to path={-| (\tikztotarget)}]
  (A) edge (B) ;
  \node[] at (8.5,1) {\makecell{parasite probability}};

\end{tikzpicture} 
\caption{situations table for $\delta_{5}({\color{red}12},{\color{blue}2}) + log(a_{{\color{red}12}{\color{blue}2}{\color{green}1}})$ for a simple double agent Viterbi without unknown words consideration, the probability $P({\color{blue}\Gamma_{r}=0}/\Gamma_{r-1}=3,{\color{red}T_{r-1}="\textbf{\AR{إسم مجرور}}"})$ is a parasite probability because the real POS tag for the word \AR{جافة} is  \AR{نعت} and not \AR{إسم مجرور} and the ingredient state of the word \AR{جافة} is $\Gamma_{r-1}=2$ and not $\Gamma_{r-1}=3$}
\label{tab:tab38}
\end{center}
\end{table}

\newpage

{\color{blue}Estimating an ingredient state using a double agent Viterbi algorithm with external knowledge transfer :}

sentence=.,10,\AR{غ},\AR{خميرة},\AR{جافة},\AR{للخبز},.

ingredient state=[0,0,0,1,2,0,0]

POS tags=[.,B,I,E,F,H,.] coded as [4, 1, 10, 6, 7, 12, 4]

ingredient to be extracted=\AR{خميرة جافة}

\begin{equation} \label{eq30}
\begin{split}
\delta_{5}({\color{red}12},{\color{blue}1}) + log(a_{{\color{red}12}{\color{blue}1}{\color{green}1}}) & =max_{1 \leq i \leq 4}(\delta_{4}({\color{orange}POS[5]},i)+log(a_{{\color{orange}POS[5]}i{\color{blue}1}}))+ log(b_{{\color{red}12}{\color{blue}1}}("\textbf{\AR{للخبز}}"))+ log(a_{{\color{red}12}{\color{blue}1}{\color{green}1}}) \\
&  = \underbrace{\delta_{4}({\color{orange}7},{\color{blue}3})}_{-93.42}+\underbrace{log(P({\color{blue}\Gamma_{r}=0}/\Gamma_{r-1}=2,{\color{orange}T_{r-1}="\textbf{\AR{نعت}}"}))}_{-0.851} \\
& +\underbrace{log(P(V_{r}="\textbf{\AR{للخبز}}"/{\color{blue}\Gamma_{r}=0},{\color{red}T_{r}="\textbf{\AR{إسم مجرور}}"})}_{-36.63} \\
& + \underbrace{log(P({\color{green}\Gamma_{r}=0}/{\color{blue}\Gamma_{r-1}=0},{\color{red}T_{r-1}="\textbf{\AR{إسم مجرور}}"}))}_{-4.192}\\
& =-135.094
\end{split}
\end{equation}

\begin{table}[h]

\begin{center}
\begin{tabular}{ c c c c c c c c c }
  & . & 10 & \AR{غ} & \AR{خميرة} & \AR{جافة} & \AR{للخبز}& . \\   
 case1 & - & - & - & - & \makecell{${\color{orange}T_{r-1}="\textbf{\AR{نعت}}"}$ \\ ${\color{black}\Gamma_{r-1}=2}$}  & ${\color{blue}\Gamma_{r}=0}$ & - \\
    case2 & - & - & - & - & - & \makecell{${\color{red}T_{r}="\textbf{\AR{إسم مجرور}}"}$ \\ ${\color{blue}\Gamma_{r}=0}$} & - \\
    case3 & - & - & - & - & - & \makecell{${\color{red}T_{r-1}="\textbf{\AR{إسم مجرور}}"}$ \\ ${\color{blue}\Gamma_{r-1}=0}$} & ${\color{green}\Gamma_{r}=0}$
\end{tabular}
\begin{tikzpicture}[overlay]
 \draw[green, line width=1.5pt] (1,2.2) ellipse (3cm and 0.5cm);
 \node (A) at (4, 2) {};
\node (B) at (7, 1) {};
\node (C) at (0.8, -1.2) {};
\node (D) at (1.8, -0.8) {};
\draw[->,green, to path={-| (\tikztotarget)}]
  (A) edge (B) ;

  \node[] at (7,0.7) {\makecell{parasite probability \\ eleminated}};
\end{tikzpicture} 
\caption{Situations table for $\delta_{5}({\color{red}12},{\color{blue}1}) + log(a_{{\color{red}12}{\color{blue}1}{\color{green}1}})$ for a simple double agent Viterbi with external knowledge transfer without unknown words consideration, the parasite probability probability $P({\color{blue}\Gamma_{r}=0}/\Gamma_{r-1}=1,{\color{red}T_{r-1}="\textbf{\AR{إسم مجرور}}"})$ is eliminated and replaced by the probability $P({\color{blue}\Gamma_{r}=0}/\Gamma_{r-1}=2,{\color{orange}T_{r-1}="\textbf{\AR{نعت}}"})$}
\label{tab:tab39}
\end{center}
\end{table}

\begin{equation} \label{eq31}
\begin{split}
\delta_{5}({\color{red}12},{\color{blue}2}) + log(a_{{\color{red}12}{\color{blue}2}{\color{green}1}}) & =max_{1 \leq i \leq 4}(\delta_{4}({\color{orange}POS[5]},i)+log(a_{{\color{orange}POS[5]}i{\color{blue}2}}))+ log(b_{{\color{red}12}{\color{blue}2}}("\textbf{\AR{للخبز}}"))+ log(a_{{\color{red}12}{\color{blue}2}{\color{green}1}}) \\
&  = \underbrace{\delta_{4}({\color{orange}7},3)}_{-93.42}+\underbrace{log(P({\color{blue}\Gamma_{r}=1}/\Gamma_{r-1}=2,{\color{orange}T_{r-1}="\textbf{\AR{نعت}}"}))}_{-6.5} \\
& +\underbrace{ log(P(V_{r}="\textbf{\AR{للخبز}}"/{\color{blue}\Gamma_{r}=1},{\color{red}T_{r}="\textbf{\AR{إسم مجرور}}"})}_{-38.015} \\
& + \underbrace{log(P({\color{green}\Gamma_{r}=0}/{\color{blue}\Gamma_{r-1}=1},{\color{red}T_{r-1}="\textbf{\AR{إسم مجرور}}"}))}_{-3.118}\\
& = -141.062
\end{split}
\end{equation}

\begin{table}[h]
\begin{center}
\begin{tabular}{ c c c c c c c c c }
 & . & 10 & \AR{غ} & \AR{خميرة} & \AR{جافة} & \AR{للخبز}& . \\   
 case1 & - & - & - & - & \makecell{${\color{orange}T_{r-1}="\textbf{\AR{نعت}}"}$ \\ ${\color{black}\Gamma_{r-1}=2}$} & ${\color{blue}\Gamma_{r}=1}$ & - \\
    case2 & - & - & - & - & - & \makecell{${\color{red}T_{r}="\textbf{\AR{إسم مجرور}}"}$ \\ ${\color{blue}\Gamma_{r}=1}$} & - \\
    case3 & - & - & - & - & - & \makecell{${\color{red}T_{r-1}="\textbf{\AR{إسم مجرور}}"}$ \\ ${\color{blue}\Gamma_{r-1}=1}$} & ${\color{green}\Gamma_{r}=0}$
\end{tabular}
\begin{tikzpicture}[overlay]
 \draw[green, line width=1.5pt] (1,2.2) ellipse (3cm and 0.5cm);
 \node (A) at (4, 2.2) {};
\node (B) at (5, 2.2) {};
\node[] at (6.2,2.2) {\makecell{parasite  \\ probability \\ eliminated}};
\draw[->,green, to path={-| (\tikztotarget)}]
  (A) edge (B) ; 
\end{tikzpicture} 
\caption{Situations table for $\delta_{5}({\color{red}12},{\color{blue}2}) + log(a_{{\color{red}12}{\color{blue}2}{\color{green}1}})$ for a simple double agent Viterbi with external knowledge transfer without unknown words consideration, the parasite probability probability $P({\color{blue}\Gamma_{r}=1}/\Gamma_{r-1}=3,{\color{red}T_{r-1}="\textbf{\AR{إسم مجرور}}"})$ is eliminated and replaced by the probability $P({\color{blue}\Gamma_{r}=1}/\Gamma_{r-1}=2,{\color{orange}T_{r-1}="\textbf{\AR{نعت}}"})$}
\label{tab:tab40}
\end{center}
\end{table}
{\color{blue}Estimating an ingredient state using a simple double agent Viterbi with unknown words consideration without external knowledge transfer:}

sentence=.,10,\AR{غ},\AR{خميرة},\AR{جافة},$\underbrace{\textbf{\AR{للخبز}}}_{unknown}$,.

ingredient state=[0,0,0,1,2,0,0]

POS tags=[.,B,I,E,F,H,.] coded as [4,10,11,7,8,9,4]

ingredient to be extracted=\AR{خميرة جافة}

\begin{gather}\label{mat8}
	\psi_{6}(i,j)=\begin{bmatrix}
	1 & 0 & 1 & 1 \\
	3 & 0 & 3 & 3 \\
	3 & 3 & 3 & 3 \\	
	0 &0 & 1 & 1 \\
          3 & 0 & 1 & 3 \\
         1 & 0 & 1 & 3 \\
          1 & 3 & 1 & 3 \\
	0 & 0 & 0 & 0 \\
	1 & 3 & 3 & 3 \\	
	0 &0 & 1 & 0 \\
           3 & 0& 3 & 3 \\
	3 & 3 & 3 & 3 \\
	0 & 3 & 3 & 3 \\	
	3 &3 & 3 & 3 
	\end{bmatrix}
	\end{gather}

\begin{equation*} 
\begin{split}
\psi_{6}(9,1) & = argmax_{1 \leq i \leq 4}(\delta_{5}(9,i) + log(a_{9i1}))
\end{split}
\end{equation*}

\begin{equation} \label{eq33}
\begin{split}
\delta_{5}({\color{red}9},{\color{blue}1}) + log(a_{{\color{red}9}{\color{blue}1}{\color{green}1}}) & =max_{1 \leq i \leq 4}(\delta_{4}({\color{red}9},i)+log(a_{{\color{red}9}i{\color{blue}1}}))+\lambda_{max} log(b_{{\color{red}9}{\color{blue}1}}("\textbf{\AR{للخبز}}"))+ log(a_{{\color{red}9}{\color{blue}1}{\color{green}1}}) \\
&  = \underbrace{\delta_{4}({\color{red}9},2)}_{-137.679}+\underbrace{log(P({\color{blue}\Gamma_{r}=0}/\Gamma_{r-1}=1,{\color{red}T_{r-1}="\textbf{\AR{إسم مجرور}}"}))}_{-3.155} \\
& +\underbrace{log(P(V_{r}="\textbf{\AR{لل}}"/{\color{blue}\Gamma_{r}=0},{\color{red}T_{r}="\textbf{\AR{إسم مجرور}}"})}_{-6.948} \\
& + \underbrace{log(P({\color{green}\Gamma_{r}=0}/{\color{blue}\Gamma_{r-1}=0},{\color{red}T_{r-1}="\textbf{\AR{إسم مجرور}}"}))}_{-4.1916}\\
& = -151.97
\end{split}
\end{equation}

\begin{table}[h]
\begin{center}
\begin{tabular}{ c c c c c c c c c }
&  &  &  &  &  &  &  \\
  & . & 10 & \AR{غ} & \AR{خميرة} & \AR{جافة} & \AR{للخبز}& . \\   
 case1 & - & - & - & - & \makecell{${\color{red}T_{r-1}=\cancel{"\textbf{\AR{إسم مجرور}}"}}$ \\ ${\color{black}\cancel{\Gamma_{r-1}=1}}$} & ${\color{blue}\Gamma_{r}=0}$ & - \\
    case2 & - & - & - & - & - & \makecell{${\color{red}T_{r}="\textbf{\AR{إسم مجرور}}"}$ \\ ${\color{blue}\Gamma_{r}=0}$} & - \\
    case3 & - & - & - & - & - & \makecell{${\color{red}T_{r-1}="\textbf{\AR{إسم مجرور}}"}$ \\ ${\color{blue}\Gamma_{r-1}=0}$} & ${\color{green}\Gamma_{r}=0}$
\end{tabular}
\begin{tikzpicture}[overlay]
 \draw[red, line width=1.5pt] (1.7,2.1) ellipse (3.2cm and 0.6cm);

 \node (A) at (4.5, 2) {};
\node (B) at (8, 1) {};
\node (C) at (0.8, -1.2) {};
\node (D) at (1.8, -0.6) {};
\draw[->,red, to path={-| (\tikztotarget)}]
  (A) edge (B) ;
  \node[] at (7,1) {\makecell{parasite probability}};
\end{tikzpicture} 
\caption{Situations table for $\delta_{5}({\color{red}9},{\color{blue}1}) + log(a_{{\color{red}9}{\color{blue}1}{\color{green}1}})$ for a simple double agent Viterbi with unknown words consideration without external knowledge transfer, $P({\color{blue}\Gamma_{r}=0}/\Gamma_{r-1}=1,{\color{red}T_{r-1}="\textbf{\AR{إسم مجرور}}"})$ is a parasite probability because the real ingredient state of the word \AR{جافة} is $\Gamma_{r-1}=2$ and the real POS tag of the word \AR{جافة} is \AR{نعت}}
\label{tab:tab41}
\end{center}
\end{table}

\begin{equation} \label{eq34}
\begin{split}
\delta_{5}({\color{red}9},{\color{blue}2}) + log(a_{{\color{red}9}{\color{blue}2}{\color{green}1}}) & =max_{1 \leq i \leq 4}(\delta_{4}({\color{red}9},i)+log(a_{{\color{red}9}i{\color{blue}2}}))+\lambda_{max} log(b_{{\color{red}9}{\color{blue}2}}("\textbf{\AR{لل}}"))+ log(a_{{\color{red}9}{\color{blue}2}{\color{green}1}}) \\
&  = \underbrace{\delta_{4}({\color{red}9},4)}_{-137.062}+\underbrace{log(P({\color{blue}\Gamma_{r}=1}/\Gamma_{r-1}=3,{\color{red}T_{r-1}="\textbf{\AR{إسم مجرور}}"}))}_{-4.0604} \\
& +\underbrace{ log(P(V_{r}="\textbf{\AR{لل}}"/{\color{blue}\Gamma_{r}=1},{\color{red}T_{r}="\textbf{\AR{إسم مجرور}}"})}_{-6.7473} \\
& + \underbrace{log(P({\color{green}\Gamma_{r}=0}/{\color{blue}\Gamma_{r-1}=1},{\color{red}T_{r-1}="\textbf{\AR{إسم مجرور}}"}))}_{-3.155}\\
& = -151.025
\end{split}
\end{equation}

\begin{table}[h]
\begin{center}
\begin{tabular}{ c c c c c c c c c }
  & . & 10 & \AR{غ} & \AR{خميرة} & \AR{جافة} & \AR{للخبز}& . \\   
 case1 & - & - & - & - & \makecell{${\color{red}\cancel{T_{r-1}="\textbf{\AR{إسم مجرور}}"}}$ \\ ${\color{black}\cancel{\Gamma_{r-1}=3}}$} & ${\color{blue}\Gamma_{r}=1}$ & - \\
    case2 & - & - & - & - & - & \makecell{${\color{red}T_{r}="\textbf{\AR{إسم مجرور}}"}$ \\ ${\color{blue}\Gamma_{r}=1}$} & - \\
    case3 & - & - & - & - & - & \makecell{${\color{red}T_{r-1}="\textbf{\AR{إسم مجرور}}"}$ \\ ${\color{blue}\Gamma_{r-1}=1}$} & ${\color{green}\Gamma_{r}=0}$
\end{tabular}
\begin{tikzpicture}[overlay]
 \draw[red, line width=1.5pt] (2,2.2) ellipse (3cm and 0.5cm);

 \node (A) at (5, 2.2) {};
\node (B) at (8.5, 1) {};
\node (C) at (0.8, -1.2) {};
\node (D) at (1.8, -0.6) {};
\draw[->,red, to path={-| (\tikztotarget)}]
  (A) edge (B) ;
  \node[] at (8.5,1) {\makecell{parasite probability}};

\end{tikzpicture} 
\caption{Situations table for $\delta_{5}({\color{red}9},{\color{blue}2}) + log(a_{{\color{red}9}{\color{blue}2}{\color{green}1}})$ for a simple double agent Viterbi with unknown words consideration, $P({\color{blue}\Gamma_{r}=1}/\Gamma_{r-1}=3,{\color{red}T_{r-1}="\textbf{\AR{إسم مجرور}}"})$ is a parasite probability because the real ingredient state of the word \AR{جافة} is $\Gamma_{r-1}=2$ and the real POS tag of the word \AR{جافة} is \AR{نعت}} 
\label{tab:tab42}
\end{center}
\end{table}
{\color{blue}Estimating an ingredient state using a simple double agent Viterbi algorithm with external knowledge transfer :}

sentence=.,10,\AR{غ},\AR{خميرة},\AR{جافة},$\underbrace{\textbf{\AR{للخبز}}}_{unknown}$,.

ingredient state=[0,0,0,1,2,0,0]

POS tags=[.,B,I,E,F,H,.] coded as [4,10,11,7,8,9,4]

ingredient to be extracted=\AR{خميرة جافة}

\begin{equation} \label{eq35}
\begin{split}
\delta_{5}({\color{red}9},{\color{blue}1}) + log(a_{{\color{red}9}{\color{blue}1}{\color{green}1}}) & =max_{1 \leq i \leq 4}(\delta_{4}({\color{orange}POS[5]},i)+log(a_{{\color{orange}POS[5]}i{\color{blue}1}}))+ log(b_{{\color{red}9}{\color{blue}1}}("\textbf{\AR{للخبز}}"))+ log(a_{{\color{red}9}{\color{blue}1}{\color{green}1}}) \\
&  = \underbrace{\delta_{4}({\color{orange}8},{\color{blue}3})}_{-96.916}+\underbrace{log(P({\color{blue}\Gamma_{r}=0}/\Gamma_{r-1}=2,{\color{orange}T_{r-1}="\textbf{\AR{نعت}}"}))}_{-0.8461} \\
& +\underbrace{log(P(V_{r}="\textbf{\AR{لل}}"/{\color{blue}\Gamma_{r}=0},{\color{red}T_{r}="\textbf{\AR{إسم مجرور}}"})}_{-6.9488} \\
& + \underbrace{log(P({\color{green}\Gamma_{r}=0}/{\color{blue}\Gamma_{r-1}=0},{\color{red}T_{r-1}="\textbf{\AR{إسم مجرور}}"}))}_{-4.1916}\\
& = -108.902
\end{split}
\end{equation}

\begin{table}[h]

\begin{center}
\begin{tabular}{ c c c c c c c c c }
  & . & 10 & \AR{غ} & \AR{خميرة} & \AR{جافة} & \AR{للخبز}& . \\   
 case1 & - & - & - & - & \makecell{${\color{orange}T_{r-1}="\textbf{\AR{نعت}}"}$ \\ ${\color{black}\Gamma_{r-1}=2}$}  & ${\color{blue}\Gamma_{r}=0}$ & - \\
    case2 & - & - & - & - & - & \makecell{${\color{red}T_{r}="\textbf{\AR{إسم مجرور}}"}$ \\ ${\color{blue}\Gamma_{r}=0}$} & - \\
    case3 & - & - & - & - & - & \makecell{${\color{red}T_{r-1}="\textbf{\AR{إسم مجرور}}"}$ \\ ${\color{blue}\Gamma_{r-1}=0}$} & ${\color{green}\Gamma_{r}=0}$
\end{tabular}
\begin{tikzpicture}[overlay]
 \draw[green, line width=1.5pt] (1,2.2) ellipse (3cm and 0.5cm);
 \node (A) at (4, 2) {};
\node (B) at (7, 1) {};
\node (C) at (0.8, -1.2) {};
\node (D) at (1.8, -0.8) {};
\draw[->,green, to path={-| (\tikztotarget)}]
  (A) edge (B) ;

  \node[] at (7,0.7) {\makecell{parasite probability \\ eleminated}};
\end{tikzpicture} 
\caption{Situations table for $\delta_{5}({\color{red}9},{\color{blue}1}) + log(a_{{\color{red}9}{\color{blue}1}{\color{green}1}})$ for a simple double agent Viterbi with external knowledge transfer with unknown words consideration, the parasite probability probability $P({\color{blue}\Gamma_{r}=0}/\Gamma_{r-1}=1,{\color{red}T_{r-1}="\textbf{\AR{إسم مجرور}}"})$ is eliminated and replaced by the probability $P({\color{blue}\Gamma_{r}=0}/\Gamma_{r-1}=2,{\color{orange}T_{r-1}="\textbf{\AR{نعت}}"})$}
\label{tab:tab43}
\end{center}

\end{table}

\begin{equation} \label{eq36}
\begin{split}
\delta_{5}({\color{red}9},{\color{blue}2}) + log(a_{{\color{red}9}{\color{blue}2}{\color{green}1}}) & =max_{1 \leq i \leq 4}(\delta_{4}({\color{orange}POS[5]},i)+log(a_{{\color{orange}POS[5]}i{\color{blue}2}}))+ log(b_{{\color{red}9}{\color{blue}2}}("\textbf{\AR{للخبز}}"))+ log(a_{{\color{red}9}{\color{blue}2}{\color{green}1}}) \\
&  = \underbrace{\delta_{4}({\color{orange}8},3)}_{-96.91}+\underbrace{log(P({\color{blue}\Gamma_{r}=1}/\Gamma_{r-1}=2,{\color{orange}T_{r-1}="\textbf{\AR{نعت}}"}))}_{-6.335} \\
& +\underbrace{ log(P(V_{r}="\textbf{\AR{لل}}"/{\color{blue}\Gamma_{r}=1},{\color{red}T_{r}="\textbf{\AR{إسم مجرور}}"})}_{-6.747} \\
& + \underbrace{log(P({\color{green}\Gamma_{r}=0}/{\color{blue}\Gamma_{r-1}=1},{\color{red}T_{r-1}="\textbf{\AR{إسم مجرور}}"}))}_{-3.155}\\
& = -113.153
\end{split}
\end{equation}

\begin{table}[h]
\begin{center}
\begin{tabular}{ c c c c c c c c c }
 & . & 10 & \AR{غ} & \AR{خميرة} & \AR{جافة} & \AR{للخبز}& . \\   
 case1 & - & - & - & - & \makecell{${\color{orange}T_{r-1}="\textbf{\AR{نعت}}"}$ \\ ${\color{black}\Gamma_{r-1}=2}$} & ${\color{blue}\Gamma_{r}=1}$ & - \\
    case2 & - & - & - & - & - & \makecell{${\color{red}T_{r}="\textbf{\AR{إسم مجرور}}"}$ \\ ${\color{blue}\Gamma_{r}=1}$} & - \\
    case3 & - & - & - & - & - & \makecell{${\color{red}T_{r-1}="\textbf{\AR{إسم مجرور}}"}$ \\ ${\color{blue}\Gamma_{r-1}=1}$} & ${\color{green}\Gamma_{r}=0}$
\end{tabular}
\begin{tikzpicture}[overlay]
 \draw[green, line width=1.5pt] (1,2.2) ellipse (3cm and 0.5cm);
 \node (A) at (4, 2.2) {};
\node (B) at (5, 2.2) {};
\node[] at (6.2,2.2) {\makecell{parasite  \\ probability \\ eliminated}};
\draw[->,green, to path={-| (\tikztotarget)}]
  (A) edge (B) ; 
\end{tikzpicture} 
\caption{Situations table for $\delta_{5}({\color{red}9},{\color{blue}2}) + log(a_{{\color{red}9}{\color{blue}2}{\color{green}1}})$ for a primitive double agent Viterbi with external knowledge transfer with unknown words consideration, the parasite probability probability $P({\color{blue}\Gamma_{r}=1}/\Gamma_{r-1}=3,{\color{red}T_{r-1}="\textbf{\AR{إسم مجرور}}"})$ is eliminated and replaced by the probability $P({\color{blue}\Gamma_{r}=1}/\Gamma_{r-1}=2,{\color{orange}T_{r-1}="\textbf{\AR{نعت}}"})$}
\label{tab:tab44}
\end{center}
\end{table}

\begin{figure}
\begin{tabular}{lll}
  &  \makecell{simple double \\ agent Viterbi \\ without \\ external knowledge \\ transfer } &\makecell{simple double \\ agent Viterbi \\ with \\ external knowledge \\ transfer} \\
\makecell{\AR{للخبز} is \\ a known \\ word} & \makecell{
 \begin{tikzpicture}[ycomb]
\draw[] (0.5,-0.2) node {X};
\draw[] (1,-0.2) node {Y};
\draw[] (2.5,-0.2) node {p};
\draw[] (-0.4,1) node {-183};
\draw[] (-0.4,2) node {-180};
\draw[] (-0.4,3) node {-177};
\draw[] (3.2,1) node {1};
\draw[] (3.2,2) node {2};
\draw[] (3.2,3) node {3};
\draw [help lines] grid (3,4);
\draw[color=green,line width=10pt]
plot coordinates{(0.5,0.92) };
\draw[color=blue,line width=10pt]
plot coordinates{ (1,0.73)};
\draw[color=red!66,line width=10pt]
plot coordinates{(2.5,2)};
\end{tikzpicture}} &
\makecell{
 \begin{tikzpicture}[ycomb]
\draw[] (0.5,-0.2) node {X};
\draw[] (1,-0.2) node {Y};
\draw[] (2.5,-0.2) node {p};
\draw[] (-0.4,1) node {-145};
\draw[] (-0.4,2) node {-140};
\draw[] (-0.4,3) node {-135};
\draw[] (3.2,1) node {1};
\draw[] (3.2,2) node {2};
\draw[] (3.2,3) node {3};
\draw [help lines] grid (3,4);
\draw[color=green,line width=10pt]
plot coordinates{(0.5,2.98) };
\draw[color=blue,line width=10pt]
plot coordinates{ (1,1.79)};
\draw[color=red!100,line width=10pt]
plot coordinates{(2.5,0.02)};
\end{tikzpicture}} \\
\makecell{\AR{للخبز} is \\ an unknown \\ word} & \makecell{
\begin{tikzpicture}[ycomb]
\draw[] (0.5,-0.2) node {X};
\draw[] (1,-0.2) node {Y};
\draw[] (2.5,-0.2) node {p};
\draw[] (-0.4,1) node {-150};
\draw[] (-0.4,2) node {-145};
\draw[] (-0.4,3) node {-140};
\draw[] (3.2,1) node {1};
\draw[] (3.2,2) node {2};
\draw[] (3.2,3) node {3};
\draw [help lines] grid (3,4);
\draw[color=green,line width=10pt]
plot coordinates{(0.5,0.6) };
\draw[color=blue,line width=10pt]
plot coordinates{ (1,0.79)};
\draw[color=red!66,line width=10pt]
plot coordinates{(2.5,2)};
\end{tikzpicture}} & \makecell{\begin{tikzpicture}[ycomb]
\draw[] (0.5,-0.2) node {X};
\draw[] (1,-0.2) node {Y};
\draw[] (2.5,-0.2) node {p};
\draw[] (-0.4,1) node {-115};
\draw[] (-0.4,2) node {-110};
\draw[] (-0.4,3) node {-105};
\draw[] (3.2,1) node {1};
\draw[] (3.2,2) node {2};
\draw[] (3.2,3) node {3};
\draw [help lines] grid (3,4);
\draw[color=green,line width=10pt]
plot coordinates{(0.5,2.22) };
\draw[color=blue,line width=10pt]
plot coordinates{ (1,1.37)};
\draw[color=red!100,line width=10pt]
plot coordinates{(2.5,0.02)};
\end{tikzpicture}}  
\end{tabular}
\caption{data sensitivity bar plot for the word \AR{للخبز} inside the sentence \AR{ 10 غ خميرة جافة للخبز} under graduated conditions. X is the first most probable case inside a state matrix (\AR{للخبز} is not an ingredient), Y is the second most probable case inside a state matrix (\AR{للخبز} is an ingredient constituted with a single word), P is the number of parasite probabilities observed when calculating X and Y.}
\label{fig:fig3}
\end{figure}
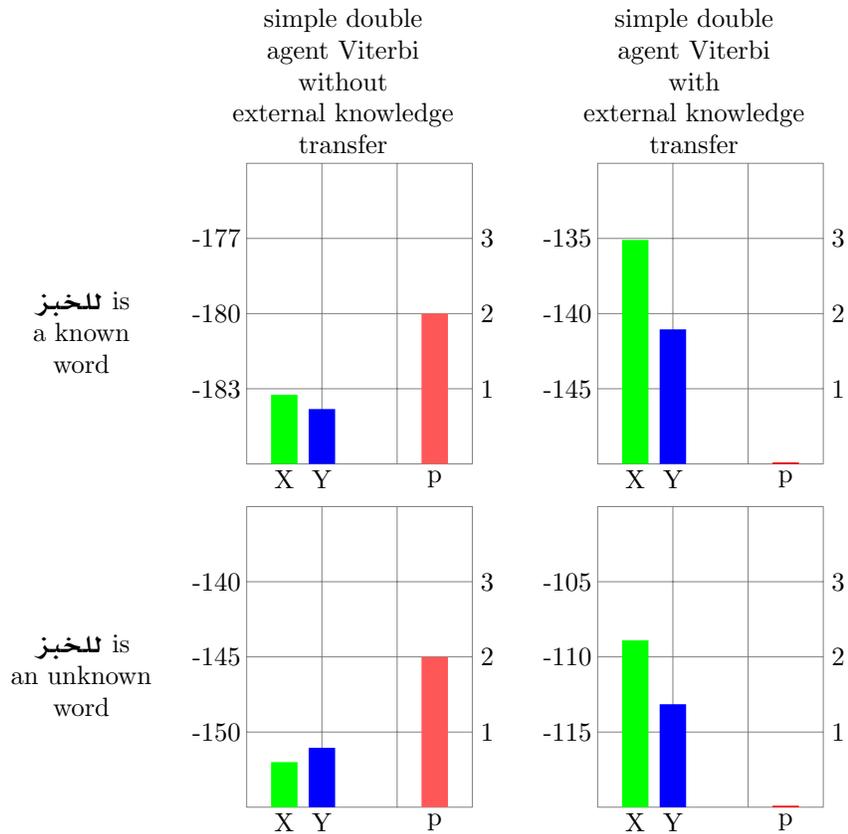

\section{Results and Discussion}

When comparing the results obtained in table {\color{blue}Table \ref{tab:tab31}}, we found that the double layered Viterbi algorithm with external knowledge transfer is overcoming largely the 3 other methods in terms of precision with a f-score value of 96,32 \% versus 70,06 \% for a first order mono-layered HMM model , 68,18 \% for a second order mono-layered HMM model and 81,32 \% for a simple double agent Viterbi algorithm. This can be explained by the external knowledge transfer characterizing our developed algorithm that minimized the amount of parasite probabilities that may appear during the calculation of the most probable elements of a state matrix. For example for the sentence "\AR{غ خميرة جافة  للخبز}10" where all the words are known both a simple double agent Viterbi and a simple double agent Viterbi with external knowlage transfer estimates well the token \AR{للخبز} which is not an ingredient, but for the first algorithm we have $|X-Y|=|\delta_{5}({\color{red}12},{\color{blue}1}) + log(a_{{\color{red}12}{\color{blue}1}{\color{green}1}})-\delta_{5}({\color{red}12},{\color{blue}2}) - log(a_{{\color{red}12}{\color{blue}2}{\color{green}1}})|=0,58$ and we have for the second one $|X-Y|=|\delta_{5}({\color{red}12},{\color{blue}1}) + log(a_{{\color{red}12}{\color{blue}1}{\color{green}1}})-\delta_{5}({\color{red}12},{\color{blue}2}) - log(a_{{\color{red}12}{\color{blue}2}{\color{green}1}})|=5,97$ ({\color{blue}figure \ref{fig:fig3}}) which means that the state matrix for a double agent Viterbi algorithm with external knowledge transfer differentiate well between the most probable cases inside a state matrix and it is more confident in estimating an ingredient state. We can explain the efficiency of our new algorithm by the fact that when the probabilities constituting the equation of the most probable elements of the state matrix are describing well the values hold by a token and not wrong parameters like we see in a parasite probability , the elements of the state matrix tend to enlarge the differences between a state describing the real estimation and a state describing a wrong estimation.

One important characteristic found in a double agent Viterbi algorithm with external knowledge transfer is its capacity to well estimate ingredients state against a mono-layered Viterbi algorithm and a simple double agent Viterbi algorithm without external knowledge transfer in a context of unknown words. When an unknown word is estimated, the lexical probability $P(V_{r}="\textbf{\AR{للخبز}}"/{\color{blue}\Gamma_{r}=0},{\color{red}T_{r}="\textbf{\AR{إسم مجرور}}"})$ becomes $P(V_{r}="\textbf{\AR{لل}}"/{\color{blue}\Gamma_{r}=0},{\color{red}T_{r}="\textbf{\AR{إسم مجرور}}"})$ and its importance for estimating hidden states on the second layer of our model becomes lower . External knowledge transfer allowed us to clean our double layered Viterbi algorithm from parasite probabilities and maintain a good f-score of 72.19 \% ({\color{blue}table \ref{tab:tab36}}) because we are confirming that knowledge transfer is increasing the precision comparing to what we got for our ingredient extractor {\color{blue}\cite{baklouti2019hidden}}({\color{blue}table \ref{tab:tab36}}). Best accuracy for unknown words is detained by our new algorithm ({\color{blue}table \ref{tab:tab36}}) . These promising performances can be explained by observing the equations determining the most probable cases of a state matrix. For example , in the sentence  "\AR{غ خميرة جافة  للخبز}10" when the token \AR{للخبز} is unknown , a primitive Viterbi algorithm does not estimated the exact ingredient state for that token which is $\Gamma=0$ given that the word \AR{للخبز} is not an ingredient , on the other side , a double layered Viterbi algorithm with internal knowledge transfer estimated well the ingredient state , this can be explained by the absence of parasite probabilities in the second algorithm given that the lexical probabilities $P(V_{r}="\textbf{\AR{لل}}"/{\color{blue}\Gamma_{r}=0},{\color{red}T_{r}="\textbf{\AR{إسم مجرور}}"})$ and $P(V_{r}="\textbf{\AR{لل}}"/{\color{blue}\Gamma_{r}=1},{\color{red}T_{r}="\textbf{\AR{إسم مجرور}}"})$ keep the same values when they are used in the a double agent Viterbi algorithm with and without external knowledge transfer.

Another concepts characterizing a simple double agent Viterbi algorithm with external knowledge transfer is its capacity to react effectively toward dataset tuning and also its capacity to resist data monotonicity logic. For example , our dataset is describing the ingredients used in a recipe and to achieve a better understanding , a typical sentence structure is frequently used to present an ingredient like using the preposition "of" and then announcing the wanted information but sometimes this monotonous logic is not always true when the preposition "of" is used to describe the ingredient like in the sentence "olive oil of good quality". We created an adapted graph to study the comportment of our new algorithm toward monotonous logic called a data sensitivity bar plot. A data sensitivity bar plot is presenting two informations : the value of the most probable cases inside a state matrix and the number of parasite probabilities observed. Those informations are experimented under graduated conditions. A graduate condition is used to explore the dynamical comportment of the algorithm , for example , we have in our dataset the sentence "olive oil of good quality" where the ingredient is situated on the beginning of the sentence and it is hard for any algorithm to give a correct prediction for this sentence when it is tagged only one time on the tag-set especially for a mono-agent Viterbi algorithm where the only parameters of this model are the tokens and the ingredients states. Data tuning is one of the graduated condition we can use by adding first a similar sentence that keep the same tags for part of speech and for ingredients state like the sentence "one tomatoes of big size" , and second by adding the same sentence to obtain a tag-set with the sentence "olive oil of good quality" repeated twice.We can associate this graduated condition with the algorithm used to predict ingredients states and deployed in a data sensitivity plot gradually from the method with the lowest f-score to the algorithm with the highest f-score ({\color{blue}figure \ref{fig:fig2}}). Data sensitivity bar plot with the previous described combination shows a certain static behaviour for a second order mono-agent Viterbi and a simple double agent Viterbi without external knowledge transfer. A dynamic behaviour is observed only for a simple double agent Viterbi without external knowledge transfer where the value of the correct state increase and the number of parasite probabilities decrease when we move from a condition with low grade to a condition with high grade. Moving from known to unknown word could be also a graduated condition used to analyse dynamic behaviour of an algorithm in a data sensitivity bar plot ({\color{blue}figure \ref{fig:fig3}}).

\section{Conclusion}

We have proposed an algorithm mixing Hidden Markov Models and external knowledge transfer to solve the problem of information extraction in natural language processing, our algorithm can be used in other problems where Hidden Markov Models are deployed.This algorithm has proven a spectacular improvement in f1 score without unknown words consideration and performances for unknown words are improved comparing to old results for a simple double agents Viterbi without external knowledge transfer {\color{blue}\cite{baklouti2019hidden}} but still can be improved by creating a full double agent Viterbi with external knowledge transfer where the number of parasite probabilities generated is less than we obtained on this paper.

\newpage

\appendix
\section{Detailed Iterations of a simple double agent Viterbi Algorithm with external knowledge transfer} \label{sec:num1}
\begin{enumerate}
	\item Predicting POS Tags for the first Layer:

	We can use for predicting  POS tags any method that provide a high accuracy and f-score. Some features used for natural language programming problems are directly extracted from the tokens like extracting the first letters from a token. 
	
	\item Predicting Ingredients state for the second Layer:
	
	 \textbf{The variables:}
	 \begin{itemize}
		\item $\delta_{l}(i,j)=max_{\tau_{1}...\tau_{l-1}}P(\tau_{1}...\tau_{l}=[t_{i},\gamma_{j}]/v_{1}...v_{l},t_{1}...t_{l},y_{1}...y_{l}), l={1...L}$
		\item $\psi_{l}(i,j)=argmax_{\tau_{1}...\tau_{l-1}}P(\tau_{1}...\tau_{l}=[t_{i},\gamma_{j}]/v_{1}...v_{l},t_{1}...t_{l},y_{1}...y_{l}), l={1...L}$
	\end{itemize}
	\textbf{The procedure:}
	\begin{enumerate}
		\item Initialization step:
		\begin{gather*} 
		\delta_{1}(i,j) = \begin{cases}
	\pi_{j}b_{ij}(v_{1}) & v_{1} \text{ is known} \text{ i={1..14} j={1..4} } \\
	\,  \\
	\pi_{j}c_{ij}(v_{1}) & v_{1} \text{ is unknown} \text{ i={1..14} j={1..4} } \\
	\end{cases}\\
		 \end{gather*}
		 \begin{gather*}
	\psi_{1}(i,j)=0  \text{ i={1..14} j={1..4}}
	\end{gather*}
		\item Iteration step:
		\begin{gather*} 
		\delta_{l}(i,j) = \begin{cases}
	max_{j}[\delta_{l-1}(y(l),j)a_{y(l)jk}]b_{ij}(v_{l}) & v_{l} \text{ is known , l={2...L} } \\ & \text{ j={1...4}, i={1..14}} \\ & \text{k={1...4}},\text{y(l) is the tag} \\ & \text{calculated at previous layer} \\ & \text{in position l} \\
	\,  \\
	max_{j}[\delta_{l-1}(y(l),j)a_{y(l)jk}]c_{ij}(v_{l}) & v_{l} \text{ is unknown , l={2...L} } \\ & \text{ j={1...4}, i={1..14}} \\ & \text{k={1...4}} ,\text{y(l) is the tag} \\ & \text{calculated at previous layer} \\ & \text{in position l}
	\end{cases}\\
		 \end{gather*}
		  \begin{gather*}
	\psi_{l}(i,j)=argmax_{j}[\delta_{l-1}(y(l),j)a_{y(l)jk}] \\  \text{j={1..4},i={1..14},k={1..4},y(l) is the tag calculated at previous layer in position l}
	\end{gather*}
	\item Termination:
		 \begin{gather*} 
\tau_{L}^{*}=argmax_{i=y(L),j=1..4}\delta_{L}(i,j)   
\\
\text{ where y(L) is the tag calculated at previous layer in position L}
		 \end{gather*}
		 \item Backtracking:
		\begin{gather*} 
		\tau_{l}^{*}=\psi_{l}(y(l),\tau_{l+1}^{*})   \text{ l=L-1..2,1} 
		\\
\text{ where y(l) is the tag calculated at previous layer in position l}
		 \end{gather*}
		 
	\end{enumerate}
	\item Predicting Ingredients state for the second Layer in Log-space and introduction of the hyper-parameter $\lambda$
	
	\textbf{The variables:}
	 \begin{itemize}
		\item $\delta_{l}(i,j)=max_{\tau_{1}...\tau_{l-1}}P(\tau_{1}...\tau_{l}=[t_{i},\gamma_{j}]/v_{1}...v_{l},t_{1}...t_{l},y_{1}...y_{l}), l={1...L}$
		\item $\psi_{l}(i,j)=argmax_{\tau_{1}...\tau_{l-1}}P(\tau_{1}...\tau_{l}=[t_{i},\gamma_{j}]/v_{1}...v_{l},t_{1}...t_{l},y_{1}...y_{l}), l={1...L}$
	\end{itemize}
	\textbf{The procedure:}
	\begin{enumerate}
		\item Initialization step:
		\begin{gather*} 
		\delta_{1}(i,j) = \begin{cases}
	\log(\pi_{j}) + \log(b_{ij}(v_{1})) & v_{1} \text{ is known} \text{ i={1..14} j={1..4} } \\
	\,  \\
	\log(\pi_{j})+\log(c_{ij}(v_{1})) & v_{1} \text{ is unknown} \text{ i={1..14} j={1..4} } \\
	\end{cases}\\
		 \end{gather*}
		 \begin{gather*}
	\psi_{1}(i,j)=0  \text{ i={1..14} j={1..4}}
	\end{gather*}
		\item Iteration step:
		\begin{gather*} 
		\hspace*{-2cm}
		\delta_{l}(i,j) = \begin{cases}
	max_{j}[\delta_{l-1}(y(l),j)+\log(a_{y(l)jk})]+\lambda_{max}\log(b_{ij}(v_{l})) & v_{l} \text{ is known  l={2..L} } \\ & \text{j={1...4}, i={1..14}} \\ & \text{k={1...4}},\text{y(l) is the tag} \\ & \text{calculated at previous layer} \\ & \text{in position l} \\
	\,  \\
	max_{j}[\delta_{l-1}(y(l),j)+\log(a_{y(l)jk})]+\lambda_{max}\log(c_{ij}(v_{l})) & v_{l} \text{ is unknown} \\ & \text{l={2...L} j={1..4}} \\ & \text{i={1..14} k={1..4}},\text{y(l) is the tag} \\ & \text{calculated at previous layer} \\ & \text{in position l} \\
	\end{cases}\\
		 \end{gather*}
		  \begin{gather*}
	\psi_{l}(i,j)=argmax_{j}[\delta_{l-1}(y(l),j)+\log(a_{y(l)jk})]  \text{ j={1..4},i={1..14},k={1..4},y(l) is the tag} \\	
	\text{ calculated at previous layer in position l}
	\end{gather*}
	\item Termination:
		 \begin{gather*} 
\tau_{L}^{*}=argmax_{i=y(L),j=1..4}\delta_{L}(i,j) 
\\
\text{ where y(L) is the tag calculated at previous layer in position L}
		 \end{gather*}
		 \item Backtracking:
		\begin{gather*} 
		\tau_{l}^{*}=\psi_{l}(y(l),\tau_{l+1}^{*})   \text{ l=L-1..2,1} 
		\\
		\text{ where y(l) is the tag calculated at previous layer in position l}
		 \end{gather*}
		 
	\end{enumerate}
	\end{enumerate}

\begin{itemize}
\item Initialization step:
	\begin{gather*}
	\hspace*{-1.5cm}
	\begin{bmatrix}
	\color{red}  d_{1,1,1}  & d_{1,1,2} & d_{1,1,3} & d_{1,1,4} \\
	d_{1,2,1} & d_{1,2,2} & d_{1,2,3} & d_{1,2,4} \\
	d_{1,3,1} & d_{1,3,2} & d_{1,3,3} & d_{1,3,4} \\	
	\vdots & \vdots & \vdots & \vdots \\
	d_{1,14,1} & d_{1,14,2} & d_{1,14,3} & d_{1,14,4}
	\end{bmatrix}
	\begin{bmatrix}
	\color{red}pi_{1}  & pi_{2} & pi_{3} & pi_{4}
	\end{bmatrix}
		\begin{bmatrix}
	\color{red}b_{1,1,x[1]}  & b_{1,2,x[1]} & b_{1,3,x[1]} & b_{1,4,x[1]} \\
	b_{2,1,x[1]} & b_{2,2,x[1]} & b_{2,3,x[1]} & b_{2,4,x[1]} \\
	b_{3,1,x[1]} & b_{3,2,x[1]} & b_{3,3,x[1]} & b_{3,4,x[1]} \\
	\vdots & \vdots & \vdots & \vdots \\	
	b_{14,1,x[1]} & b_{14,2,x[1]} & b_{14,3,x[1]} & b_{14,4,x[1]}
	\end{bmatrix}
	\end{gather*}
	\begin{gather*}
	\color{red}d_{1,1,1}=pi_{1} \mtimes b_{1,1,x[1]}
	\end{gather*}
	\begin{gather*}
	\hspace*{-1.5cm}
	\begin{bmatrix}
	d_{1,1,1}  & \color{red} d_{1,1,2} & d_{1,1,3} & d_{1,1,4} \\
	d_{1,2,1} & d_{1,2,2} & d_{1,2,3} & d_{1,2,4} \\
	d_{1,3,1} & d_{1,3,2} & d_{1,3,3} & d_{1,3,4} \\	
	\vdots & \vdots & \vdots & \vdots \\
	d_{1,14,1} & d_{1,14,2} & d_{1,14,3} & d_{1,14,4}
	\end{bmatrix}
	\begin{bmatrix}
	pi_{1}  &\color{red} pi_{2} & pi_{3} & pi_{4}
	\end{bmatrix}
		\begin{bmatrix}
	b_{1,1,x[1]} & \color{red} b_{1,2,x[1]} & b_{1,3,x[1]} & b_{1,4,x[1]} \\
	b_{2,1,x[1]} & b_{2,2,x[1]} & b_{2,3,x[1]} & b_{2,4,x[1]} \\
	b_{3,1,x[1]} & b_{3,2,x[1]} & b_{3,3,x[1]} & b_{3,4,x[1]} \\	
	\vdots & \vdots & \vdots & \vdots \\
	b_{14,1,x[1]} & b_{14,2,x[1]} & b_{14,3,x[1]} & b_{14,4,x[1]}
	\end{bmatrix}
	\end{gather*}
	\begin{gather*}
	\color{red}d_{1,1,2}=pi_{2} \mtimes b_{1,2,x[1]}
	\end{gather*}
	\begin{gather*}
	\hspace*{-1.5cm}
	\begin{bmatrix}
	d_{1,1,1}  & d_{1,1,2}  & \color{red} d_{1,1,3} & d_{1,1,4} \\
	d_{1,2,1} & d_{1,2,2} & d_{1,2,3} & d_{1,2,4} \\
	d_{1,3,1} & d_{1,3,2} & d_{1,3,3} & d_{1,3,4} \\	
	\vdots & \vdots & \vdots & \vdots \\
	d_{1,14,1} & d_{1,14,2} & d_{1,14,3} & d_{1,14,4}
	\end{bmatrix}
	\begin{bmatrix}
	pi_{1}  & pi_{2} &\color{red} pi_{3} & pi_{4}
	\end{bmatrix}
		\begin{bmatrix}
	b_{1,1,x[1]} & b_{1,2,x[1]} & \color{red} b_{1,3,x[1]} & b_{1,4,x[1]} \\
	b_{2,1,x[1]} & b_{2,2,x[1]} & b_{2,3,x[1]} & b_{2,4,x[1]} \\
	b_{3,1,x[1]} & b_{3,2,x[1]} & b_{3,3,x[1]} & b_{3,4,x[1]} \\
	\vdots & \vdots & \vdots & \vdots \\	
	b_{14,1,x[1]} & b_{14,2,x[1]} & b_{14,3,x[1]} & b_{14,4,x[1]}
	\end{bmatrix}
	\end{gather*}
	\begin{gather*}
	\color{red}d_{1,1,3}=pi_{3} \mtimes b_{1,3,x[1]}
	\end{gather*}
	\begin{gather*}
	\hspace*{-1.5cm}
	\begin{bmatrix}
	d_{1,1,1}  & d_{1,1,2}  &  d_{1,1,3}&  \color{red} d_{1,1,4}  \\
	d_{1,2,1} & d_{1,2,2} & d_{1,2,3} & d_{1,2,4} \\
	d_{1,3,1} & d_{1,3,2} & d_{1,3,3} & d_{1,3,4} \\	
	\vdots & \vdots & \vdots & \vdots \\
	d_{1,14,1} & d_{1,14,2} & d_{1,14,3} & d_{1,14,4}
	\end{bmatrix}
	\begin{bmatrix}
	pi_{1}  & pi_{2} & pi_{3} &\color{red} pi_{4} 
	\end{bmatrix}
		\begin{bmatrix}
	b_{1,1,x[1]} & b_{1,2,x[1]} & b_{1,3,x[1]} &\color{red} b_{1,4,x[1]} \\
	b_{2,1,x[1]} & b_{2,2,x[1]} & b_{2,3,x[1]} & b_{2,4,x[1]} \\
	b_{3,1,x[1]} & b_{3,2,x[1]} & b_{3,3,x[1]} & b_{3,4,x[1]} \\
	\vdots & \vdots & \vdots & \vdots \\	
	b_{14,1,x[1]} & b_{14,2,x[1]} & b_{14,3,x[1]} & b_{14,4,x[1]}
	\end{bmatrix}
	\end{gather*}
	\begin{gather*}
	\color{red}d_{1,1,4}=pi_{4} \mtimes b_{1,4,x[1]}
	\end{gather*}
	\begin{gather*}
	\hspace*{-1.5cm}
	\begin{bmatrix}
	d_{1,1,1}  & d_{1,1,2}  &  d_{1,1,3}& d_{1,1,4} \\
	 \color{red} d_{1,2,1} & d_{1,2,2} & d_{1,2,3} & d_{1,2,4} \\
	d_{1,3,1} & d_{1,3,2} & d_{1,3,3} & d_{1,3,4} \\	
	\vdots & \vdots & \vdots & \vdots \\
	d_{1,14,1} & d_{1,14,2} & d_{1,14,3} & d_{1,14,4}
	\end{bmatrix}
	\begin{bmatrix}
	\color{red} pi_{1}  & pi_{2} & pi_{3} & pi_{4}
	\end{bmatrix}
		\begin{bmatrix}
	b_{1,1,x[1]} & b_{1,2,x[1]} & b_{1,3,x[1]} & b_{1,4,x[1]} \\
	\color{red} b_{2,1,x[1]}  & b_{2,2,x[1]} & b_{2,3,x[1]} & b_{2,4,x[1]} \\
	b_{3,1,x[1]} & b_{3,2,x[1]} & b_{3,3,x[1]} & b_{3,4,x[1]} \\
	\vdots & \vdots & \vdots & \vdots \\	
	b_{14,1,x[1]} & b_{14,2,x[1]} & b_{14,3,x[1]} & b_{14,4,x[1]}
	\end{bmatrix}
	\end{gather*}
	\begin{gather*}
	\color{red}d_{1,2,1}=pi_{1} \mtimes b_{2,1,x[1]}
	\end{gather*}
	\begin{gather*}
	\vdots
	\end{gather*}
	\begin{gather*}
	\hspace*{-1.5cm}
	\begin{bmatrix}
	d_{1,1,1}  & d_{1,1,2}  &  d_{1,1,3}& d_{1,1,4} \\
	  d_{1,2,1} & d_{1,2,2} & d_{1,2,3} & d_{1,2,4} \\
	d_{1,3,1} & d_{1,3,2} & d_{1,3,3} & d_{1,3,4} \\	
	\vdots & \vdots & \vdots & \vdots \\
	d_{1,14,1} & d_{1,14,2} & d_{1,14,3} & \color{red} d_{1,14,4} 
	\end{bmatrix}
	\begin{bmatrix}
	 pi_{1}  & pi_{2} & pi_{3} & \color{red} pi_{4} 
	\end{bmatrix}
		\begin{bmatrix}
	b_{1,1,x[1]} & b_{1,2,x[1]} & b_{1,3,x[1]} & b_{1,4,x[1]} \\
	 b_{2,1,x[1]}  & b_{2,2,x[1]} & b_{2,3,x[1]} & b_{2,4,x[1]} \\
	b_{3,1,x[1]} & b_{3,2,x[1]} & b_{3,3,x[1]} & b_{3,4,x[1]} \\
	\vdots & \vdots & \vdots & \vdots \\	
	b_{14,1,x[1]} & b_{14,2,x[1]} & b_{14,3,x[1]} & \color{red} b_{14,4,x[1]} 
	\end{bmatrix}
	\end{gather*}
	\begin{gather*}
	\color{red}d_{1,14,4}=pi_{4} \mtimes b_{14,4,x[1]}
	\end{gather*}
\item Iteration step:
\begin{gather*}
\hspace*{-5cm}
\setlength\arraycolsep{-0.5pt}
\scalefont{1}{
\begin{bmatrix}
	\color{red} d_{2,1,1} \tikz[overlay,remember picture] \node (d_{2,1,1}){}; & d_{2,1,2}  & d_{2,1,3} & d_{2,1,4} \\
	d_{2,2,1}  & d_{2,2,2} & d_{2,2,3} & d_{2,2,4} \\
	d_{2,3,1} & d_{2,3,2} & d_{2,3,3} & d_{2,3,4} \\
	\vdots & \vdots & \vdots & \vdots \\	
	d_{2,14,1} & d_{2,14,2} & d_{2,14,3} &  d_{2,14,4} 
	\end{bmatrix}
=max_{i}
\Bigg(
\begin{bmatrix}
	\color{red}d_{1,y[2],1}   \tikz[overlay,remember picture] \node (d_{1,y[2],1}){};& \color{red} d_{1,y[2],2}  &\color{red} d_{1,y[2],3} &\color{red} d_{1,y[2],4}  
	\end{bmatrix}
	\cdot
	\begin{bmatrix}
	\color{red}a_{y[2],1,1}  \tikz[overlay,remember picture] \node (a_{y[2],1,1}){}; &\color{red} a_{y[2],2,1}  & \color{red}a_{y[2],3,1} & \color{red}a_{y[2],4,1} 
	\end{bmatrix}
	\Bigg)
	\cdot
		\begin{bmatrix}
	\color{red}b_{1,1,x[2]} \tikz[overlay,remember picture] \node (b_{1,1,x[2]}){}; & b_{1,2,x[2]} & b_{1,3,x[2]} &  b_{1,4,x[2]} \\
	b_{2,1,x[2]}  & b_{2,2,x[2]} & b_{2,3,x[2]} & b_{2,4,x[2]} \\
	b_{3,1,x[2]} & b_{3,2,x[2]} & b_{3,3,x[2]} & b_{3,4,x[2]} \\
	\vdots & \vdots & \vdots & \vdots \\	
	b_{14,1,x[2]} & b_{14,2,x[2]} & b_{14,3,x[2]} &  b_{14,4,x[2]} 
	\end{bmatrix}
	}
\end{gather*}				
\begin{gather*}
	\color{red} d_{2,1,1}=max(d_{1,y[2],1} \mtimes a_{y[2],1,1},d_{1,y[2],2} \mtimes a_{y[2],2,1},\cdots,d_{1,y[2],4} \mtimes a_{y[2],4,1})\mtimes b_{1,1,x[2]}
	\end{gather*}
\begin{gather*}
\hspace*{-2cm}
\setlength\arraycolsep{0pt}
\scalefont{1}{
\begin{bmatrix}
	\color{red} p_{2,1,1}  & p_{2,1,2}  & p_{2,1,3} & p_{2,1,4} \\
	p_{2,2,1}  & p_{2,2,2} & p_{2,2,3} & p_{2,2,4} \\
	p_{2,3,1} & p_{2,3,2} & p_{2,3,3} & p_{2,3,4} \\
	\vdots & \vdots & \vdots & \vdots \\	
	p_{2,14,1} & p_{2,14,2} & p_{2,14,3} &  p_{2,14,4} 
	\end{bmatrix}
=argmax_{i}
\Bigg(
\begin{bmatrix}
	\color{red}d_{1,y[2],1}   \tikz[overlay,remember picture] \node (d_{1,y[2],1}){};& \color{red} d_{1,y[2],2}  &\color{red} d_{1,y[2],3} &\color{red} d_{1,y[2],4}  
	\end{bmatrix}
	\cdot
	\begin{bmatrix}
	\color{red}a_{y[2],1,1}  \tikz[overlay,remember picture] \node (a_{y[2],1,1}){}; &\color{red} a_{y[2],2,1}  & \color{red}a_{y[2],3,1} & \color{red}a_{y[2],4,1} 
	\end{bmatrix}
	\Bigg)
	}
\end{gather*}				
\begin{gather*}
	\color{red} p_{2,1,1}=argmax(d_{1,y[2],1} \mtimes a_{y[2],1,1},d_{1,y[2],2} \mtimes a_{y[2],2,1},\cdots,d_{1,y[2],4} \mtimes a_{y[2],4,1})
	\end{gather*}		
	
\begin{gather*}
\hspace*{-5cm}
\setlength\arraycolsep{-0.5pt}
\scalefont{1}{
\begin{bmatrix}
	 d_{2,1,1} \tikz[overlay,remember picture] \node (d_{2,1,1}){}; & d_{2,1,2}  & d_{2,1,3} & d_{2,1,4} \\
	\color{red}d_{2,2,1}  & d_{2,2,2} & d_{2,2,3} & d_{2,2,4} \\
	d_{2,3,1} & d_{2,3,2} & d_{2,3,3} & d_{2,3,4} \\
	\vdots & \vdots & \vdots & \vdots \\	
	d_{2,14,1} & d_{2,14,2} & d_{2,14,3} &  d_{2,14,4} 
	\end{bmatrix}
=max_{i}
\Bigg(
\begin{bmatrix}
	\color{red}d_{1,y[2],1}   \tikz[overlay,remember picture] \node (d_{1,y[2],1}){};& \color{red} d_{1,y[2],2}  &\color{red} d_{1,y[2],3} &\color{red} d_{1,y[2],4}  
	\end{bmatrix}
	\cdot
	\begin{bmatrix}
	\color{red}a_{y[2],1,1}  \tikz[overlay,remember picture] \node (a_{y[2],1,1}){}; &\color{red} a_{y[2],2,1}  & \color{red}a_{y[2],3,1} & \color{red}a_{y[2],4,1} 
	\end{bmatrix}
	\Bigg)
	\cdot
		\begin{bmatrix}
	b_{1,1,x[2]} \tikz[overlay,remember picture] \node (b_{1,1,x[2]}){}; & b_{1,2,x[2]} & b_{1,3,x[2]} &  b_{1,4,x[2]} \\
	\color{red}b_{2,1,x[2]}  & b_{2,2,x[2]} & b_{2,3,x[2]} & b_{2,4,x[2]} \\
	b_{3,1,x[2]} & b_{3,2,x[2]} & b_{3,3,x[2]} & b_{3,4,x[2]} \\
	\vdots & \vdots & \vdots & \vdots \\	
	b_{14,1,x[2]} & b_{14,2,x[2]} & b_{14,3,x[2]} &  b_{14,4,x[2]} 
	\end{bmatrix}
	}
\end{gather*}				
\begin{gather*}
	\color{red} d_{2,2,1}=max(d_{1,y[2],1} \mtimes a_{y[2],1,1},d_{1,y[2],2} \mtimes a_{y[2],2,1},\cdots,d_{1,y[2],4} \mtimes a_{y[2],4,1})\mtimes b_{2,1,x[2]}
	\end{gather*}
\begin{gather*}
\hspace*{-2cm}
\setlength\arraycolsep{0pt}
\scalefont{1}{
\begin{bmatrix}
	 p_{2,1,1}  & p_{2,1,2}  & p_{2,1,3} & p_{2,1,4} \\
	\color{red}p_{2,2,1}  & p_{2,2,2} & p_{2,2,3} & p_{2,2,4} \\
	p_{2,3,1} & p_{2,3,2} & p_{2,3,3} & p_{2,3,4} \\
	\vdots & \vdots & \vdots & \vdots \\	
	p_{2,14,1} & p_{2,14,2} & p_{2,14,3} &  p_{2,14,4} 
	\end{bmatrix}
=argmax_{i}
\Bigg(
\begin{bmatrix}
	\color{red}d_{1,y[2],1}   \tikz[overlay,remember picture] \node (d_{1,y[2],1}){};& \color{red} d_{1,y[2],2}  &\color{red} d_{1,y[2],3} &\color{red} d_{1,y[2],4}  
	\end{bmatrix}
	\cdot
	\begin{bmatrix}
	\color{red}a_{y[2],1,1}  \tikz[overlay,remember picture] \node (a_{y[2],1,1}){}; &\color{red} a_{y[2],2,1}  & \color{red}a_{y[2],3,1} & \color{red}a_{y[2],4,1} 
	\end{bmatrix}
	\Bigg)
	}
\end{gather*}				
\begin{gather*}
	\color{red} p_{2,2,1}=argmax(d_{1,y[2],1} \mtimes a_{y[2],1,1},d_{1,y[2],2} \mtimes a_{y[2],2,1},\cdots,d_{1,y[2],4} \mtimes a_{y[2],4,1})
	\end{gather*}	
				
\begin{gather*}
\hspace*{-5cm}
\setlength\arraycolsep{-0.5pt}
\scalefont{1}{
\begin{bmatrix}
	 d_{2,1,1} \tikz[overlay,remember picture] \node (d_{2,1,1}){}; & d_{2,1,2}  & d_{2,1,3} & d_{2,1,4} \\
	d_{2,2,1}  & d_{2,2,2} & d_{2,2,3} & d_{2,2,4} \\
	\color{red}d_{2,3,1} & d_{2,3,2} & d_{2,3,3} & d_{2,3,4} \\
	\vdots & \vdots & \vdots & \vdots \\	
	d_{2,14,1} & d_{2,14,2} & d_{2,14,3} &  d_{2,14,4} 
	\end{bmatrix}
=max_{i}
\Bigg(
\begin{bmatrix}
	\color{red}d_{1,y[2],1}   \tikz[overlay,remember picture] \node (d_{1,y[2],1}){};& \color{red} d_{1,y[2],2}  &\color{red} d_{1,y[2],3} &\color{red} d_{1,y[2],4}  
	\end{bmatrix}
	\cdot
	\begin{bmatrix}
	\color{red}a_{y[2],1,1}  \tikz[overlay,remember picture] \node (a_{y[2],1,1}){}; &\color{red} a_{y[2],2,1}  & \color{red}a_{y[2],3,1} & \color{red}a_{y[2],4,1} 
	\end{bmatrix}
	\Bigg)
	\cdot
		\begin{bmatrix}
	b_{1,1,x[2]} \tikz[overlay,remember picture] \node (b_{1,1,x[2]}){}; & b_{1,2,x[2]} & b_{1,3,x[2]} &  b_{1,4,x[2]} \\
	b_{2,1,x[2]}  & b_{2,2,x[2]} & b_{2,3,x[2]} & b_{2,4,x[2]} \\
	\color{red}b_{3,1,x[2]} & b_{3,2,x[2]} & b_{3,3,x[2]} & b_{3,4,x[2]} \\
	\vdots & \vdots & \vdots & \vdots \\	
	b_{14,1,x[2]} & b_{14,2,x[2]} & b_{14,3,x[2]} &  b_{14,4,x[2]} 
	\end{bmatrix}
	}
\end{gather*}				
\begin{gather*}
	\color{red} d_{2,3,1}=max(d_{1,y[2],1} \mtimes a_{y[2],1,1},d_{1,y[2],2} \mtimes a_{y[2],2,1},\cdots,d_{1,y[2],4} \mtimes a_{y[2],4,1})\mtimes b_{3,1,x[2]}
	\end{gather*}
\begin{gather*}
\hspace*{-2cm}
\setlength\arraycolsep{0pt}
\scalefont{1}{
\begin{bmatrix}
	 p_{2,1,1}  & p_{2,1,2}  & p_{2,1,3} & p_{2,1,4} \\
	p_{2,2,1}  & p_{2,2,2} & p_{2,2,3} & p_{2,2,4} \\
	\color{red}p_{2,3,1} & p_{2,3,2} & p_{2,3,3} & p_{2,3,4} \\
	\vdots & \vdots & \vdots & \vdots \\	
	p_{2,14,1} & p_{2,14,2} & p_{2,14,3} &  p_{2,14,4} 
	\end{bmatrix}
=argmax_{i}
\Bigg(
\begin{bmatrix}
	\color{red}d_{1,y[2],1}   \tikz[overlay,remember picture] \node (d_{1,y[2],1}){};& \color{red} d_{1,y[2],2}  &\color{red} d_{1,y[2],3} &\color{red} d_{1,y[2],4}  
	\end{bmatrix}
	\cdot
	\begin{bmatrix}
	\color{red}a_{y[2],1,1}  \tikz[overlay,remember picture] \node (a_{y[2],1,1}){}; &\color{red} a_{y[2],2,1}  & \color{red}a_{y[2],3,1} & \color{red}a_{y[2],4,1} 
	\end{bmatrix}
	\Bigg)
	}
\end{gather*}				
\begin{gather*}
	\color{red} p_{2,3,1}=argmax(d_{1,y[2],1} \mtimes a_{y[2],1,1},d_{1,y[2],2} \mtimes a_{y[2],2,1},\cdots,d_{1,y[2],4} \mtimes a_{y[2],4,1})
	\end{gather*}	
	
		\begin{gather*}
	\vdots
	\end{gather*}
\begin{gather*}
\hspace*{-5cm}
\setlength\arraycolsep{-0.5pt}
\scalefont{1}{
\begin{bmatrix}
	 d_{2,1,1} \tikz[overlay,remember picture] \node (d_{2,1,1}){}; & d_{2,1,2}  & d_{2,1,3} & d_{2,1,4} \\
	d_{2,2,1}  & d_{2,2,2} & d_{2,2,3} & d_{2,2,4} \\
	d_{2,3,1} & d_{2,3,2} & d_{2,3,3} & d_{2,3,4} \\
	\vdots & \vdots & \vdots & \vdots \\	
	\color{red}d_{2,14,1} & d_{2,14,2} & d_{2,14,3} &  d_{2,14,4} 
	\end{bmatrix}
=max_{i}
\Bigg(
\begin{bmatrix}
	\color{red}d_{1,y[2],1}   \tikz[overlay,remember picture] \node (d_{1,y[2],1}){};& \color{red} d_{1,y[2],2}  &\color{red} d_{1,y[2],3} &\color{red} d_{1,y[2],4}  
	\end{bmatrix}
	\cdot
	\begin{bmatrix}
	\color{red}a_{y[2],1,1}  \tikz[overlay,remember picture] \node (a_{y[2],1,1}){}; &\color{red} a_{y[2],2,1}  & \color{red}a_{y[2],3,1} & \color{red}a_{y[2],4,1} 
	\end{bmatrix}
	\Bigg)
	\cdot
		\begin{bmatrix}
	b_{1,1,x[2]} \tikz[overlay,remember picture] \node (b_{1,1,x[2]}){}; & b_{1,2,x[2]} & b_{1,3,x[2]} &  b_{1,4,x[2]} \\
	b_{2,1,x[2]}  & b_{2,2,x[2]} & b_{2,3,x[2]} & b_{2,4,x[2]} \\
	b_{3,1,x[2]} & b_{3,2,x[2]} & b_{3,3,x[2]} & b_{3,4,x[2]} \\
	\vdots & \vdots & \vdots & \vdots \\	
	\color{red}b_{14,1,x[2]} & b_{14,2,x[2]} & b_{14,3,x[2]} &  b_{14,4,x[2]} 
	\end{bmatrix}
	}
\end{gather*}				
\begin{gather*}
	\color{red} d_{2,14,1}=max(d_{1,y[2],1} \mtimes a_{y[2],1,1},d_{1,y[2],2} \mtimes a_{y[2],2,1},\cdots,d_{1,y[2],4} \mtimes a_{y[2],4,1})\mtimes b_{14,1,x[2]}
	\end{gather*}
\begin{gather*}
\hspace*{-2cm}
\setlength\arraycolsep{0pt}
\scalefont{1}{
\begin{bmatrix}
	 p_{2,1,1}  & p_{2,1,2}  & p_{2,1,3} & p_{2,1,4} \\
	p_{2,2,1}  & p_{2,2,2} & p_{2,2,3} & p_{2,2,4} \\
	p_{2,3,1} & p_{2,3,2} & p_{2,3,3} & p_{2,3,4} \\
	\vdots & \vdots & \vdots & \vdots \\	
	\color{red}p_{2,14,1} & p_{2,14,2} & p_{2,14,3} &  p_{2,14,4} 
	\end{bmatrix}
=argmax_{i}
\Bigg(
\begin{bmatrix}
	\color{red}d_{1,y[2],1}   \tikz[overlay,remember picture] \node (d_{1,y[2],1}){};& \color{red} d_{1,y[2],2}  &\color{red} d_{1,y[2],3} &\color{red} d_{1,y[2],4}  
	\end{bmatrix}
	\cdot
	\begin{bmatrix}
	\color{red}a_{y[2],1,1}  \tikz[overlay,remember picture] \node (a_{y[2],1,1}){}; &\color{red} a_{y[2],2,1}  & \color{red}a_{y[2],3,1} & \color{red}a_{y[2],4,1} 
	\end{bmatrix}
	\Bigg)
	}
\end{gather*}				
\begin{gather*}
	\color{red} p_{2,14,1}=argmax(d_{1,y[2],1} \mtimes a_{y[2],1,1},d_{1,y[2],2} \mtimes a_{y[2],2,1},\cdots,d_{1,y[2],4} \mtimes a_{y[2],4,1})
	\end{gather*}		
		
\begin{gather*}
\hspace*{-5cm}
\setlength\arraycolsep{-0.5pt}
\scalefont{1}{
\begin{bmatrix}
	 d_{2,1,1}  & \color{red}d_{2,1,2}  & d_{2,1,3} & d_{2,1,4} \\
	d_{2,2,1}  & d_{2,2,2} & d_{2,2,3} & d_{2,2,4} \\
	d_{2,3,1} & d_{2,3,2} & d_{2,3,3} & d_{2,3,4} \\
	\vdots & \vdots & \vdots & \vdots \\	
	d_{2,14,1} & d_{2,14,2} & d_{2,14,3} &  d_{2,14,4} 
	\end{bmatrix}
=max_{i}
\Bigg(
\begin{bmatrix}
	\color{red}d_{1,y[2],1}   \tikz[overlay,remember picture] \node (d_{1,y[2],1}){};& \color{red} d_{1,y[2],2}  &\color{red} d_{1,y[2],3} &\color{red} d_{1,y[2],4}  
	\end{bmatrix}
	\cdot
	\begin{bmatrix}
	\color{red}a_{y[2],1,2}   &\color{red} a_{y[2],2,2}  & \color{red}a_{y[2],3,2} & \color{red}a_{y[2],4,2} 
	\end{bmatrix}
	\Bigg)
	\cdot
		\begin{bmatrix}
	b_{1,1,x[2]}  & \color{red}b_{1,2,x[2]} & b_{1,3,x[2]} &  b_{1,4,x[2]} \\
	b_{2,1,x[2]}  & b_{2,2,x[2]} & b_{2,3,x[2]} & b_{2,4,x[2]} \\
	b_{3,1,x[2]} & b_{3,2,x[2]} & b_{3,3,x[2]} & b_{3,4,x[2]} \\
	\vdots & \vdots & \vdots & \vdots \\	
	b_{14,1,x[2]} & b_{14,2,x[2]} & b_{14,3,x[2]} &  b_{14,4,x[2]} 
	\end{bmatrix}
	}
\end{gather*}				
\begin{gather*}
	\color{red} d_{2,1,2}=max(d_{1,y[2],1} \mtimes a_{y[2],1,2},d_{1,y[2],2} \mtimes a_{y[2],2,2},\cdots,d_{1,y[2],4} \mtimes a_{y[2],4,2})\mtimes b_{1,2,x[2]}
	\end{gather*}
\begin{gather*}
\hspace*{-2cm}
\setlength\arraycolsep{0pt}
\scalefont{1}{
\begin{bmatrix}
	 p_{2,1,1}  & \color{red}p_{2,1,2}  & p_{2,1,3} & p_{2,1,4} \\
	p_{2,2,1}  & p_{2,2,2} & p_{2,2,3} & p_{2,2,4} \\
	p_{2,3,1} & p_{2,3,2} & p_{2,3,3} & p_{2,3,4} \\
	\vdots & \vdots & \vdots & \vdots \\	
	p_{2,14,1} & p_{2,14,2} & p_{2,14,3} &  p_{2,14,4} 
	\end{bmatrix}
=argmax_{i}
\Bigg(
\begin{bmatrix}
	\color{red}d_{1,y[2],1}   \tikz[overlay,remember picture] \node (d_{1,y[2],1}){};& \color{red} d_{1,y[2],2}  &\color{red} d_{1,y[2],3} &\color{red} d_{1,y[2],4}  
	\end{bmatrix}
	\cdot
	\begin{bmatrix}
	\color{red}a_{y[2],1,2}  &\color{red} a_{y[2],2,2}  & \color{red}a_{y[2],3,2} & \color{red}a_{y[2],4,2} 
	\end{bmatrix}
	\Bigg)
	}
\end{gather*}				
\begin{gather*}
	\color{red} p_{2,1,2}=argmax(d_{1,y[2],1} \mtimes a_{y[2],1,2},d_{1,y[2],2} \mtimes a_{y[2],2,2},\cdots,d_{1,y[2],4} \mtimes a_{y[2],4,2})
	\end{gather*}	
	
\begin{gather*}
\hspace*{-5cm}
\setlength\arraycolsep{-0.5pt}
\scalefont{1}{
\begin{bmatrix}
	 d_{2,1,1}  & d_{2,1,2}  & d_{2,1,3} & d_{2,1,4} \\
	d_{2,2,1}  & \color{red}d_{2,2,2} & d_{2,2,3} & d_{2,2,4} \\
	d_{2,3,1} & d_{2,3,2} & d_{2,3,3} & d_{2,3,4} \\
	\vdots & \vdots & \vdots & \vdots \\	
	d_{2,14,1} & d_{2,14,2} & d_{2,14,3} &  d_{2,14,4} 
	\end{bmatrix}
=max_{i}
\Bigg(
\begin{bmatrix}
	\color{red}d_{1,y[2],1}   \tikz[overlay,remember picture] \node (d_{1,y[2],1}){};& \color{red} d_{1,y[2],2}  &\color{red} d_{1,y[2],3} &\color{red} d_{1,y[2],4}  
	\end{bmatrix}
	\cdot
	\begin{bmatrix}
	\color{red}a_{y[2],1,2}   &\color{red} a_{y[2],2,2}  & \color{red}a_{y[2],3,2} & \color{red}a_{y[2],4,2} 
	\end{bmatrix}
	\Bigg)
	\cdot
		\begin{bmatrix}
	b_{1,1,x[2]}  & b_{1,2,x[2]} & b_{1,3,x[2]} &  b_{1,4,x[2]} \\
	b_{2,1,x[2]}  & \color{red}b_{2,2,x[2]} & b_{2,3,x[2]} & b_{2,4,x[2]} \\
	b_{3,1,x[2]} & b_{3,2,x[2]} & b_{3,3,x[2]} & b_{3,4,x[2]} \\
	\vdots & \vdots & \vdots & \vdots \\	
	b_{14,1,x[2]} & b_{14,2,x[2]} & b_{14,3,x[2]} &  b_{14,4,x[2]} 
	\end{bmatrix}
	}
\end{gather*}				
\begin{gather*}
	\color{red} d_{2,2,2}=max(d_{1,y[2],1} \mtimes a_{y[2],1,2},d_{1,y[2],2} \mtimes a_{y[2],2,2},\cdots,d_{1,y[2],4} \mtimes a_{y[2],4,2})\mtimes b_{2,2,x[2]}
	\end{gather*}
\begin{gather*}
\hspace*{-2cm}
\setlength\arraycolsep{0pt}
\scalefont{1}{
\begin{bmatrix}
	 p_{2,1,1}  & p_{2,1,2}  & p_{2,1,3} & p_{2,1,4} \\
	p_{2,2,1}  & \color{red}p_{2,2,2} & p_{2,2,3} & p_{2,2,4} \\
	p_{2,3,1} & p_{2,3,2} & p_{2,3,3} & p_{2,3,4} \\
	\vdots & \vdots & \vdots & \vdots \\	
	p_{2,14,1} & p_{2,14,2} & p_{2,14,3} &  p_{2,14,4} 
	\end{bmatrix}
=argmax_{i}
\Bigg(
\begin{bmatrix}
	\color{red}d_{1,y[2],1}   \tikz[overlay,remember picture] \node (d_{1,y[2],1}){};& \color{red} d_{1,y[2],2}  &\color{red} d_{1,y[2],3} &\color{red} d_{1,y[2],4}  
	\end{bmatrix}
	\cdot
	\begin{bmatrix}
	\color{red}a_{y[2],1,2}  &\color{red} a_{y[2],2,2}  & \color{red}a_{y[2],3,2} & \color{red}a_{y[2],4,2} 
	\end{bmatrix}
	\Bigg)
	}
\end{gather*}				
\begin{gather*}
	\color{red} p_{2,2,2}=argmax(d_{1,y[2],1} \mtimes a_{y[2],1,2},d_{1,y[2],2} \mtimes a_{y[2],2,2},\cdots,d_{1,y[2],4} \mtimes a_{y[2],4,2})
	\end{gather*}	
	\begin{gather*}
	\vdots
	\end{gather*}
\begin{gather*}
\hspace*{-5cm}
\setlength\arraycolsep{-0.5pt}
\scalefont{1}{
\begin{bmatrix}
	 d_{2,1,1}  & d_{2,1,2}  & d_{2,1,3} & d_{2,1,4} \\
	d_{2,2,1}  & d_{2,2,2} & d_{2,2,3} & d_{2,2,4} \\
	d_{2,3,1} & d_{2,3,2} & d_{2,3,3} & d_{2,3,4} \\
	\vdots & \vdots & \vdots & \vdots \\	
	d_{2,14,1} & d_{2,14,2} & d_{2,14,3} &  \color{red}d_{2,14,4} 
	\end{bmatrix}
=max_{i}
\Bigg(
\begin{bmatrix}
	\color{red}d_{1,y[2],1}   \tikz[overlay,remember picture] \node (d_{1,y[2],1}){};& \color{red} d_{1,y[2],2}  &\color{red} d_{1,y[2],3} &\color{red} d_{1,y[2],4}  
	\end{bmatrix}
	\cdot
	\begin{bmatrix}
	\color{red}a_{y[2],1,4}   &\color{red} a_{y[2],2,4}  & \color{red}a_{y[2],3,4} & \color{red}a_{y[2],4,4} 
	\end{bmatrix}
	\Bigg)
	\cdot
		\begin{bmatrix}
	b_{1,1,x[2]}  & b_{1,2,x[2]} & b_{1,3,x[2]} &  b_{1,4,x[2]} \\
	b_{2,1,x[2]}  & b_{2,2,x[2]} & b_{2,3,x[2]} & b_{2,4,x[2]} \\
	b_{3,1,x[2]} & b_{3,2,x[2]} & b_{3,3,x[2]} & b_{3,4,x[2]} \\
	\vdots & \vdots & \vdots & \vdots \\	
	b_{14,1,x[2]} & b_{14,2,x[2]} & b_{14,3,x[2]} &  \color{red}b_{14,4,x[2]} 
	\end{bmatrix}
	}
\end{gather*}				
\begin{gather*}
	\color{red} d_{2,14,4}=max(d_{1,y[2],1} \mtimes a_{y[2],1,4},d_{1,y[2],2} \mtimes a_{y[2],2,4},\cdots,d_{1,y[2],4} \mtimes a_{y[2],4,4})\mtimes b_{14,4,x[2]}
	\end{gather*}
\begin{gather*}
\hspace*{-2cm}
\setlength\arraycolsep{0pt}
\scalefont{1}{
\begin{bmatrix}
	 p_{2,1,1}  & p_{2,1,2}  & p_{2,1,3} & p_{2,1,4} \\
	p_{2,2,1}  & p_{2,2,2} & p_{2,2,3} & p_{2,2,4} \\
	p_{2,3,1} & p_{2,3,2} & p_{2,3,3} & p_{2,3,4} \\
	\vdots & \vdots & \vdots & \vdots \\	
	p_{2,14,1} & p_{2,14,2} & p_{2,14,3} &  \color{red}p_{2,14,4} 
	\end{bmatrix}
=argmax_{i}
\Bigg(
\begin{bmatrix}
	\color{red}d_{1,y[2],1}   \tikz[overlay,remember picture] \node (d_{1,y[2],1}){};& \color{red} d_{1,y[2],2}  &\color{red} d_{1,y[2],3} &\color{red} d_{1,y[2],4}  
	\end{bmatrix}
	\cdot
	\begin{bmatrix}
	\color{red}a_{y[2],1,4}  &\color{red} a_{y[2],2,4}  & \color{red}a_{y[2],3,4} & \color{red}a_{y[2],4,4} 
	\end{bmatrix}
	\Bigg)
	}
\end{gather*}				
\begin{gather*}
	\color{red} p_{2,14,4}=argmax(d_{1,y[2],1} \mtimes a_{y[2],1,4},d_{1,y[2],2} \mtimes a_{y[2],2,4},\cdots,d_{1,y[2],4} \mtimes a_{y[2],4,4})
	\end{gather*}		

\begin{gather*}
\hspace*{-5cm}
\setlength\arraycolsep{-0.5pt}
\scalefont{1}{
\begin{bmatrix}
	\color{red} d_{3,1,1}  & d_{3,1,2}  & d_{3,1,3} & d_{3,1,4} \\
	d_{3,2,1}  & d_{3,2,2} & d_{3,2,3} & d_{3,2,4} \\
	d_{3,3,1} & d_{3,3,2} & d_{3,3,3} & d_{3,3,4} \\
	\vdots & \vdots & \vdots & \vdots \\	
	d_{3,14,1} & d_{3,14,2} & d_{3,14,3} &  d_{3,14,4} 
	\end{bmatrix}
=max_{i}
\Bigg(
\begin{bmatrix}
	\color{red}d_{2,y[3],1}   & \color{red} d_{2,y[3],2}  &\color{red} d_{2,y[3],3} &\color{red} d_{2,y[3],4}  
	\end{bmatrix}
	\cdot
	\begin{bmatrix}
	\color{red}a_{y[3],1,1}   &\color{red} a_{y[3],2,1}  & \color{red}a_{y[3],3,1} & \color{red}a_{y[3],4,1} 
	\end{bmatrix}
	\Bigg)
	\cdot
		\begin{bmatrix}
	\color{red}b_{1,1,x[3]}  & b_{1,2,x[3]} & b_{1,3,x[3]} &  b_{1,4,x[3]} \\
	b_{2,1,x[3]}  & b_{2,2,x[3]} & b_{2,3,x[3]} & b_{2,4,x[3]} \\
	b_{3,1,x[3]} & b_{3,2,x[3]} & b_{3,3,x[3]} & b_{3,4,x[3]} \\
	\vdots & \vdots & \vdots & \vdots \\	
	b_{14,1,x[3]} & b_{14,2,x[3]} & b_{14,3,x[3]} &  b_{14,4,x[3]} 
	\end{bmatrix}
	}
\end{gather*}				
\begin{gather*}
\color{red} d_{3,1,1}=max(d_{2,y[3],1} \mtimes a_{y[3],1,1},d_{2,y[3],2} \mtimes a_{y[3],2,1},\cdots,d_{2,y[3],4} \mtimes a_{y[3],4,1})\mtimes b_{1,1,x[3]}
\end{gather*}
\begin{gather*}
\hspace*{-2cm}
\setlength\arraycolsep{0pt}
\scalefont{1}{
\begin{bmatrix}
	 \color{red}p_{3,1,1}  & p_{3,1,2}  & p_{3,1,3} & p_{3,1,4} \\
	p_{3,2,1}  & p_{3,2,2} & p_{3,2,3} & p_{3,2,4} \\
	p_{3,3,1} & p_{3,3,2} & p_{3,3,3} & p_{3,3,4} \\
	\vdots & \vdots & \vdots & \vdots \\	
	p_{3,14,1} & p_{3,14,2} & p_{3,14,3} &  p_{3,14,4} 
	\end{bmatrix}
=argmax_{i}
\Bigg(
\begin{bmatrix}
	\color{red}d_{2,y[3],1}   & \color{red} d_{2,y[3],2}  &\color{red} d_{2,y[3],3} &\color{red} d_{2,y[3],4}  
	\end{bmatrix}
	\cdot
	\begin{bmatrix}
	\color{red}a_{y[3],1,1}   &\color{red} a_{y[3],2,1}  & \color{red}a_{y[3],3,1} & \color{red}a_{y[3],4,1} 
	\end{bmatrix}
	\Bigg)
	}
\end{gather*}				
\begin{gather*}
	\color{red} p_{3,1,1}=argmax(d_{2,y[3],1} \mtimes a_{y[3],1,1},d_{2,y[3],2} \mtimes a_{y[3],2,1},\cdots,d_{2,y[3],4} \mtimes a_{y[3],4,1})
	\end{gather*}		
	\begin{gather*}
	\vdots
	\end{gather*}
\begin{gather*}
\scalefont{0.7}{
\hspace*{-5cm}
\setlength\arraycolsep{-0.5pt}
\scalefont{1}{
\begin{bmatrix}
	\color{red} d_{L,1,1}  & d_{L,1,2}  & d_{L,1,3} & d_{L,1,4} \\
	d_{L,2,1}  & d_{L,2,2} & d_{L,2,3} & d_{L,2,4} \\
	d_{L,3,1} & d_{L,3,2} & d_{L,3,3} & d_{L,3,4} \\
	\vdots & \vdots & \vdots & \vdots \\	
	d_{L,14,1} & d_{L,14,2} & d_{L,14,3} &  d_{L,14,4} 
	\end{bmatrix}
=max_{i}
\Bigg(
\begin{bmatrix}
	\color{red}d_{L-1,y[L],1}   & \color{red} d_{L-1,y[L],2}  &\color{red} d_{L-1,y[L],3} &\color{red} d_{L-1,y[L],4}  
	\end{bmatrix}
	\cdot
	\begin{bmatrix}
	\color{red}a_{y[L],1,1}   &\color{red} a_{y[L],2,1}  & \color{red}a_{y[L],3,1} & \color{red}a_{y[L],4,1} 
	\end{bmatrix}
	\Bigg)
	\cdot
		\begin{bmatrix}
	\color{red}b_{1,1,x[L]}  & b_{1,2,x[L]} & b_{1,3,x[L]} &  b_{1,4,x[L]} \\
	b_{2,1,x[L]}  & b_{2,2,x[L]} & b_{2,3,x[L]} & b_{2,4,x[L]} \\
	b_{3,1,x[L]} & b_{3,2,x[L]} & b_{3,3,x[L]} & b_{3,4,x[L]} \\
	\vdots & \vdots & \vdots & \vdots \\	
	b_{14,1,x[L]} & b_{14,2,x[L]} & b_{14,3,x[L]} &  b_{14,4,x[L]} 
	\end{bmatrix}
	}
	}
\end{gather*}				
\begin{gather*}
\color{red} d_{L,1,1}=max(d_{L-1,y[L],1} \mtimes a_{y[L],1,1},d_{L-1,y[L],2} \mtimes a_{y[L],2,1},\cdots,d_{L-1,y[L],4} \mtimes a_{y[L],4,1})\mtimes b_{1,1,x[L]}
\end{gather*}
\begin{gather*}
\hspace*{-3cm}
\setlength\arraycolsep{0pt}
\scalefont{1}{
\begin{bmatrix}
	 \color{red}p_{L,1,1}  & p_{L,1,2}  & p_{L,1,3} & p_{L,1,4} \\
	p_{L,2,1}  & p_{L,2,2} & p_{L,2,3} & p_{L,2,4} \\
	p_{L,3,1} & p_{L,3,2} & p_{L,3,3} & p_{L,3,4} \\
	\vdots & \vdots & \vdots & \vdots \\	
	p_{L,14,1} & p_{L,14,2} & p_{L,14,3} &  p_{L,14,4} 
	\end{bmatrix}
=argmax_{i}
\Bigg(
\begin{bmatrix}
	\color{red}d_{L-1,y[L],1}   & \color{red} d_{L-1,y[L],2}  &\color{red} d_{L-1,y[L],3} &\color{red} d_{L-1,y[L],4}  
	\end{bmatrix}
	\cdot
	\begin{bmatrix}
	\color{red}a_{y[L],1,1}   &\color{red} a_{y[L],2,1}  & \color{red}a_{y[L],3,1} & \color{red}a_{y[L],4,1} 
	\end{bmatrix}
	\Bigg)
	}
\end{gather*}				
\begin{gather*}
	\color{red} p_{L,1,1}=argmax(d_{L-1,y[L],1} \mtimes a_{y[L],1,1},d_{L-1,y[L],2} \mtimes a_{y[L],2,1},\cdots,d_{L-1,y[L],4} \mtimes a_{y[L],4,1})
	\end{gather*}	
	
	\begin{gather*}
	\vdots
	\end{gather*}	
	
\begin{gather*}
\scalefont{0.7}{
\hspace*{-5cm}
\setlength\arraycolsep{-0.5pt}
\scalefont{1}{
\begin{bmatrix}
	 d_{L,1,1}  & d_{L,1,2}  & d_{L,1,3} & d_{L,1,4} \\
	d_{L,2,1}  & d_{L,2,2} & d_{L,2,3} & d_{L,2,4} \\
	d_{L,3,1} & d_{L,3,2} & d_{L,3,3} & d_{L,3,4} \\
	\vdots & \vdots & \vdots & \vdots \\	
	d_{L,14,1} & d_{L,14,2} & d_{L,14,3} & \color{red} d_{L,14,4} 
	\end{bmatrix}
=max_{i}
\Bigg(
\begin{bmatrix}
	\color{red}d_{L-1,y[L],1}   & \color{red} d_{L-1,y[L],2}  &\color{red} d_{L-1,y[L],3} &\color{red} d_{L-1,y[L],4}  
	\end{bmatrix}
	\cdot
	\begin{bmatrix}
	\color{red}a_{y[L],1,4}   &\color{red} a_{y[L],2,4}  & \color{red}a_{y[L],3,4} & \color{red}a_{y[L],4,4} 
	\end{bmatrix}
	\Bigg)
	\cdot
		\begin{bmatrix}
	b_{1,1,x[L]}  & b_{1,2,x[L]} & b_{1,3,x[L]} &  b_{1,4,x[L]} \\
	b_{2,1,x[L]}  & b_{2,2,x[L]} & b_{2,3,x[L]} & b_{2,4,x[L]} \\
	b_{3,1,x[L]} & b_{3,2,x[L]} & b_{3,3,x[L]} & b_{3,4,x[L]} \\
	\vdots & \vdots & \vdots & \vdots \\	
	b_{14,1,x[L]} & b_{14,2,x[L]} & b_{14,3,x[L]} &  \color{red}b_{14,4,x[L]} 
	\end{bmatrix}
	}
	}
\end{gather*}				
\begin{gather*}
\color{red} d_{L,14,4}=max(d_{L-1,y[L],1} \mtimes a_{y[L],1,4},d_{L-1,y[L],2} \mtimes a_{y[L],2,4},\cdots,d_{L-1,y[L],4} \mtimes a_{y[L],4,4})\mtimes b_{14,4,x[L]}
\end{gather*}
\begin{gather*}
\hspace*{-3cm}
\setlength\arraycolsep{0pt}
\scalefont{1}{
\begin{bmatrix}
	 p_{L,1,1}  & p_{L,1,2}  & p_{L,1,3} & p_{L,1,4} \\
	p_{L,2,1}  & p_{L,2,2} & p_{L,2,3} & p_{L,2,4} \\
	p_{L,3,1} & p_{L,3,2} & p_{L,3,3} & p_{L,3,4} \\
	\vdots & \vdots & \vdots & \vdots \\	
	p_{L,14,1} & p_{L,14,2} & p_{L,14,3} &  \color{red}p_{L,14,4} 
	\end{bmatrix}
=argmax_{i}
\Bigg(
\begin{bmatrix}
	\color{red}d_{L-1,y[L],1}   & \color{red} d_{L-1,y[L],2}  &\color{red} d_{L-1,y[L],3} &\color{red} d_{L-1,y[L],4}  
	\end{bmatrix}
	\cdot
	\begin{bmatrix}
	\color{red}a_{y[L],1,1}   &\color{red} a_{y[L],2,1}  & \color{red}a_{y[L],3,1} & \color{red}a_{y[L],4,1} 
	\end{bmatrix}
	\Bigg)
	}
\end{gather*}				
\begin{gather*}
	\color{red} p_{L,14,4}=argmax(d_{L-1,y[L],1} \mtimes a_{y[L],1,4},d_{L-1,y[L],2} \mtimes a_{y[L],2,4},\cdots,d_{L-1,y[L],4} \mtimes a_{y[L],4,4})
	\end{gather*}		
		\item Termination:
	\begin{gather*}
	\hspace*{-2.5cm}
	\begin{bmatrix}
	\tau_{1}  & \tau_{2} & \cdots& \tau_{L-2} & \tau_{L-1} & \color{red}\tau_{L}
	\end{bmatrix}
	=arg_{j} max_{1 \leq i \leq 14,1 \leq j \leq 4}
	\Bigg(
	\begin{bmatrix}
	  d_{L,1,1}& d_{L,1,2}   & d_{L,1,3} & d_{L,1,4} \\
	 d_{L,2,1}  & d_{L,2,2} & d_{L,2,3} & d_{L,2,4} \\
	d_{L,3,1}  & d_{L,3,2} & d_{L,3,3} & d_{L,3,4} \\
	\vdots & \vdots & \vdots & \vdots \\	
	d_{L,14,1}  & d_{L,14,2} & d_{L,14,3} &   d_{L,14,4} 
	\end{bmatrix}
	\Bigg)		
	\end{gather*}
	\item Backtracking:
	\begin{gather*}
	\hspace*{-2.5cm}
	\begin{bmatrix}
	\tau_{1}  & \tau_{2} & \cdots & \tau_{L-2} & \color{red} \tau_{L-1} & \tau_{L}
	\end{bmatrix}
	=\psi(L,Y_{L},\tau_{L})
	\end{gather*}
	\begin{gather*}
	\vdots
	\end{gather*}
	\begin{gather*}
	\hspace*{-2.5cm}
	\begin{bmatrix}
	\color{red}\tau_{1}  & \tau_{2} & \cdots & \tau_{L-2} &  \tau_{L-1} & \tau_{L}
	\end{bmatrix}
	=\psi(2,Y_{2},\tau_{2})
	\end{gather*}
\end{itemize}

\newpage
\listoffigures
\listoftables
\bibliographystyle{apacite}
\bibliography{mybibfile}

\end{document}